\documentclass[final]{cvpr}

\usepackage{times}
\usepackage{epsfig}
\usepackage{graphicx}
\usepackage{amsmath}
\usepackage{amssymb}
\usepackage{kotex}
\usepackage{xcolor} %

\usepackage[pagebackref=true,breaklinks=true,colorlinks,bookmarks=false]{hyperref}

\def\cvprPaperID{11462} %
\def\confYear{CVPR 2021}

\usepackage{pifont} %
\usepackage{graphicx}

\newcommand{\stylemap}{stylemap\xspace}
\newcommand{\stylemaps}{stylemaps\xspace}
\newcommand{\stylemapgan}{StyleMapGAN\xspace}
\newcommand{\celebahq}{CelebA-HQ\xspace}
\newcommand{\FIDlerp}{FID\textsubscript{lerp}\xspace}

\newcommand{\m}{\mathbf{m}}
\newcommand{\w}{\mathbf{w}}

\newcommand{\x}{\mathbf{x}}
\newcommand{\xt}{\widetilde{\mathbf{x}}}
\newcommand{\wt}{\widetilde{\mathbf{w}}}

\newcommand{\R}{\mathbb{R}}

\newcommand{\figdemo}{
\begin{figure*}[t]
\centering
\includegraphics[width=0.918\linewidth]{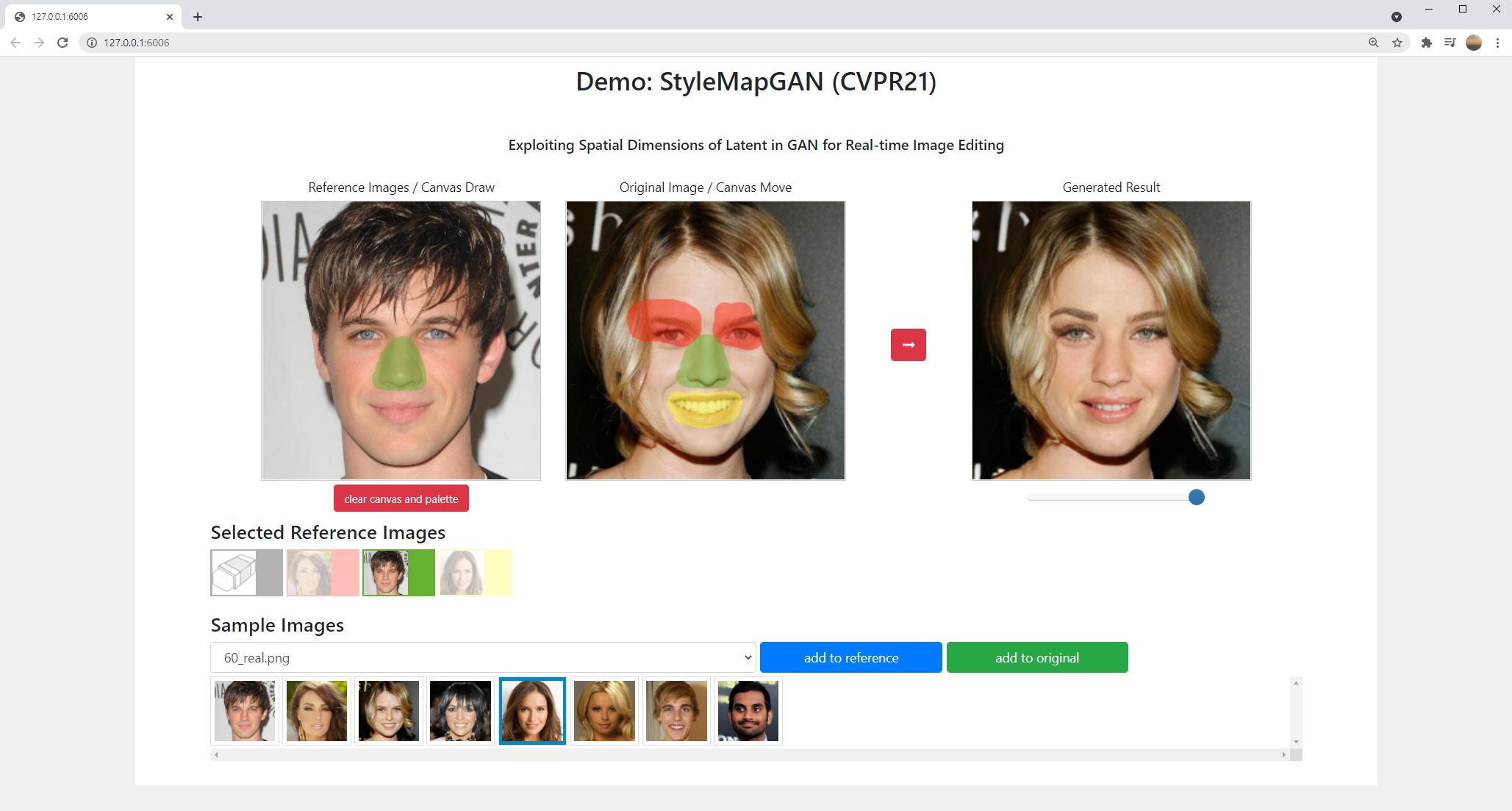}\\
\caption{\textbf{StyleMapGAN Interactive Demo.} Please see our code for further details.} 
\label{fig:demo}
\vspace{-3mm}
\end{figure*}
}

\newcommand{\fignetarch}{
\begin{figure}[t]
\centering
\includegraphics[width=1.0\linewidth]{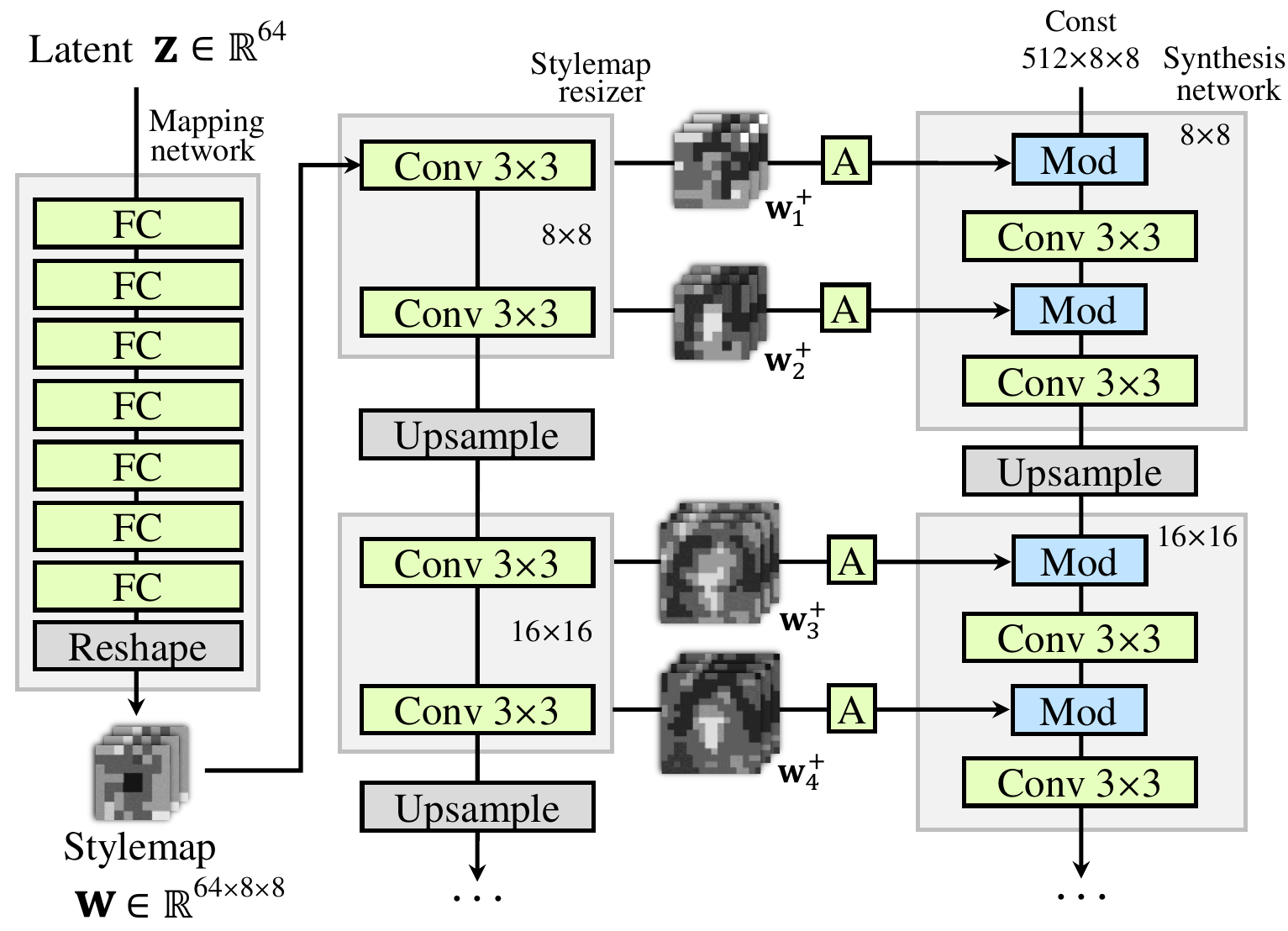}\\
\caption{\textbf{StyleMapGAN Generator.} The stylemap $\w$ is resized to $\w^{+}$ through convolutional layers to match the spatial resolution of each feature in the synthesis network. Here ``A" stands for a learned affine transform, which produces spatial modulation parameters ($\gamma$ and $\beta$ in Equation~\ref{eqn::modulation}). ``Mod" indicates modulation consisting of element-wise multiplication and addition. Note that the synthesis network starts from a learned constant input, and the output image's \textit{style} is adjusted by resized \stylemaps.} 
\label{fig:network_architecture}
\end{figure}
}

\newcommand{\figtraininference}{
\begin{figure}[t]
\centering
\includegraphics[width=1.0\columnwidth]{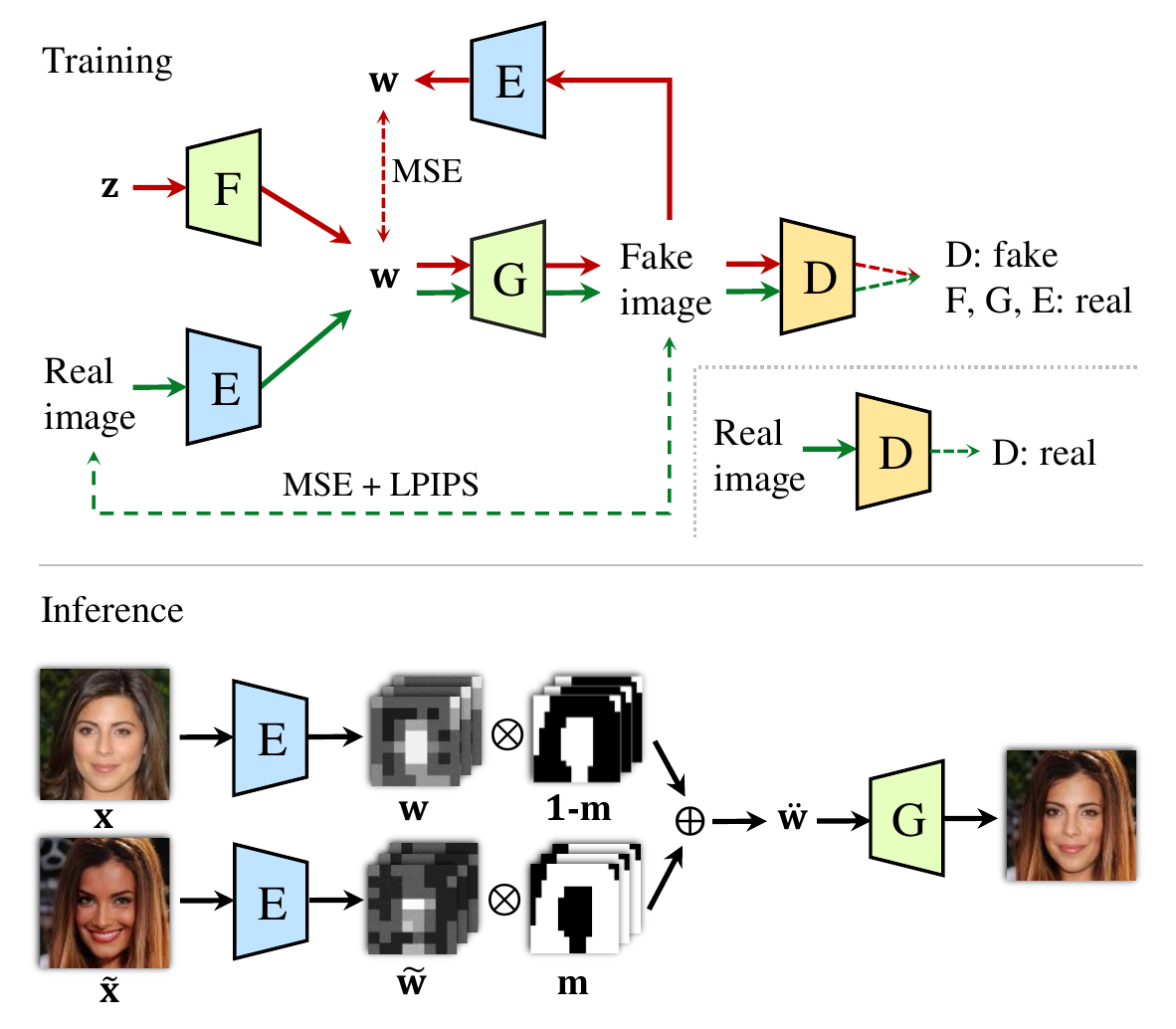}
   \caption{The upper figure contains an overall training scheme. Green and red arrows refer to flows associated with the real image and the generated image from Gaussian distribution, respectively. Dashed lines indicate loss functions. The lower figure shows our local editing method on the \stylemap.
   }
\label{fig:figtraininference}
\end{figure}
}

\newcommand{\figAblationEditing}{
\begin{figure*}[t]
    \centering
    \makebox[\himg\linewidth][c]{~~~Original}\hfill
    \makebox[\himg\linewidth][c]{~Reference}\hfill
    \makebox[\himg\linewidth][c]{$4\times4$}\hfill
    \makebox[\himg\linewidth][c]{$8\times8$}\hfill
    \makebox[\himg\linewidth][c]{$16\times16$~}\hfill
    \makebox[\himg\linewidth][c]{$32\times32$~~~}\hfill 
    \includegraphics[width=\himg\linewidth]{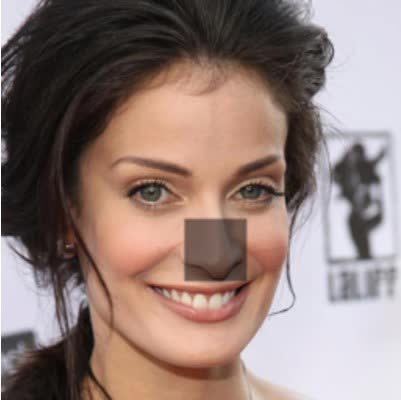}
    \includegraphics[width=\himg\linewidth]{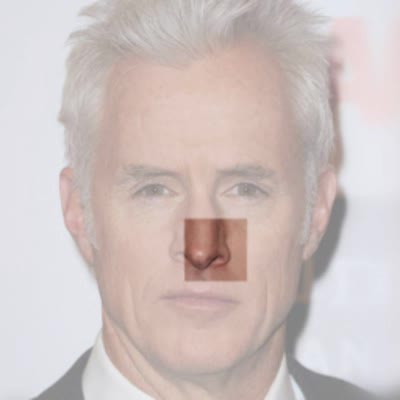}
    \includegraphics[width=\himg\linewidth]{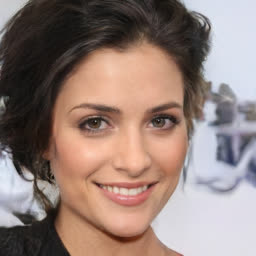}
    \includegraphics[width=\himg\linewidth]{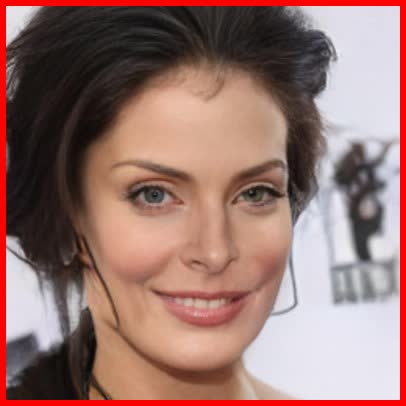}
    \includegraphics[width=\himg\linewidth]{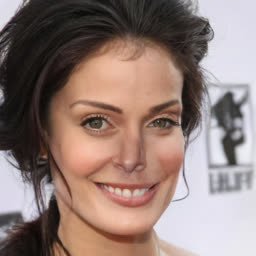}
    \includegraphics[width=\himg\linewidth]{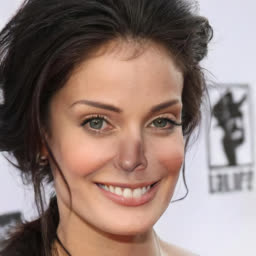} \\

    \includegraphics[width=\himg\linewidth]{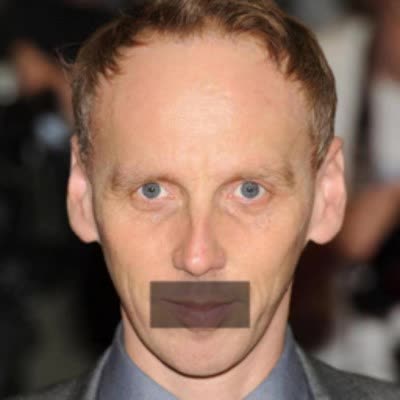}
    \includegraphics[width=\himg\linewidth]{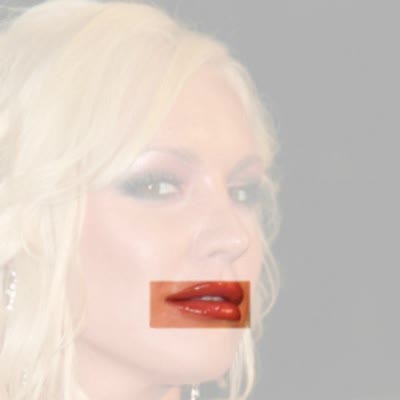}
    \includegraphics[width=\himg\linewidth]{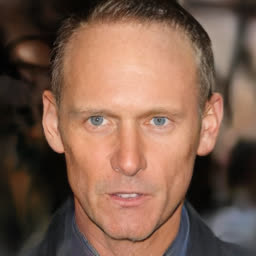}
    \includegraphics[width=\himg\linewidth]{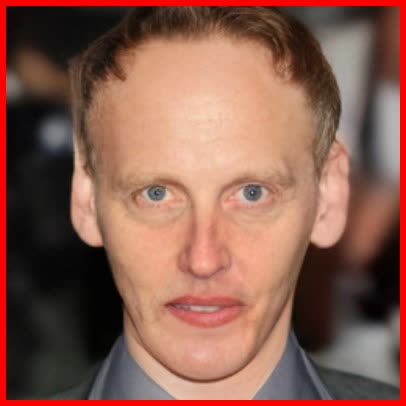}
    \includegraphics[width=\himg\linewidth]{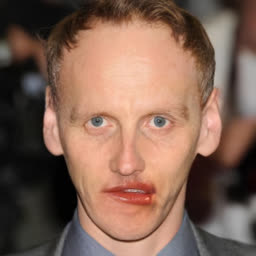}
    \includegraphics[width=\himg\linewidth]{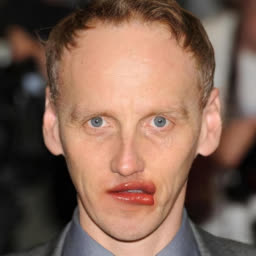}    
       
\caption{
Local editing comparison across different resolutions of the \stylemap. Regions to be discarded are faded on the original and the reference images. $4\times4$ suffers from the poor reconstruction. Resolutions greater than or equal to $16\times16$ result in too heterogeneous images. $8\times8$ resolution shows acceptable reconstruction and natural integration. Note that our method works well even in the case that the mask locates improperly as shown in the reference image of the first row.} 
\label{fig:ablation_editing}
\end{figure*}
}

\newcommand{\figComparisonCeleb}{
\newcommand{\hzero}{0.15}
\newcommand{\hone}{0.135}
\newcommand{\htwo}{0.135}
\begin{figure*}[t]
    \begin{minipage}[]{\linewidth}
    \footnotesize{
    \newcolumntype{h}{>{\centering\arraybackslash}p{19.8mm}}
    \setlength\tabcolsep{7pt} %
    \begin{tabular}{hhhhhhh}
    ~~~~Original & ~~~Reference & ~Structured Noise & ~Editing in Style & ~In-DomainGAN & SEAN & Ours
    \end{tabular}
    }
    \end{minipage}
    \centering
    \\
    \begin{subfigure}[h]{\hone\linewidth}
        \includegraphics[width=\linewidth]{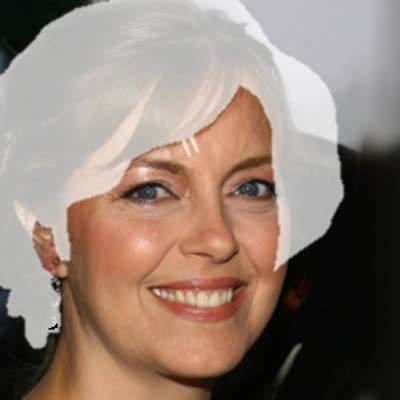}
    \end{subfigure}
    \begin{subfigure}[h]{\htwo\linewidth}
        \includegraphics[width=\linewidth]{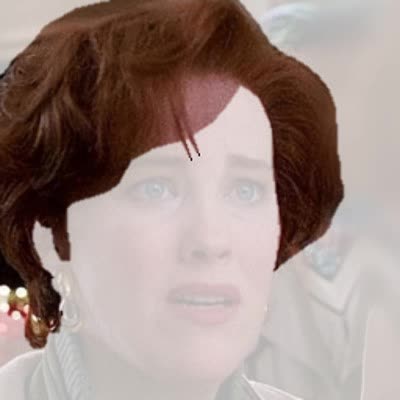}
    \end{subfigure}
    \begin{subfigure}[h]{\htwo\linewidth}
        \includegraphics[width=\linewidth]{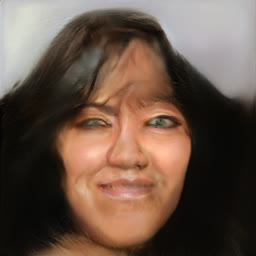}
    \end{subfigure}
    \begin{subfigure}[h]{\htwo\linewidth}
        \includegraphics[width=\linewidth]{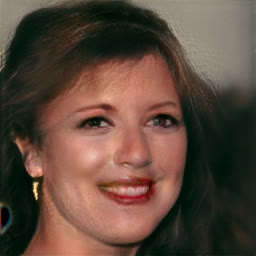}
    \end{subfigure}
    \begin{subfigure}[h]{\htwo\linewidth}
        \includegraphics[width=\linewidth]{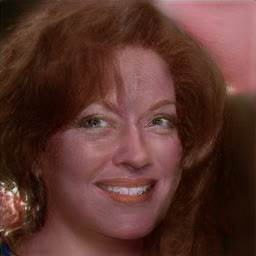}
    \end{subfigure}    
    \begin{subfigure}[h]{\htwo\linewidth}
        \includegraphics[width=\linewidth]{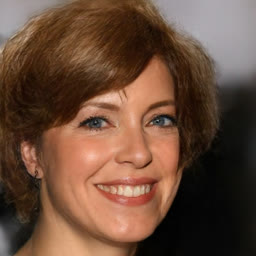}
    \end{subfigure}
    \begin{subfigure}[h]{\htwo\linewidth}
        \centering
        \includegraphics[width=\linewidth]{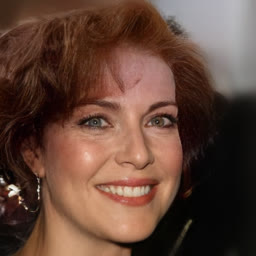}
    \end{subfigure}\\

    \begin{subfigure}[h]{\hone\linewidth}
        \includegraphics[width=\linewidth]{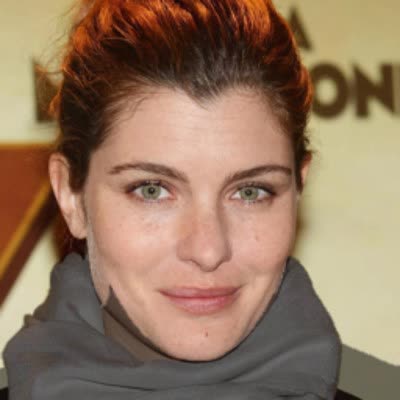}
    \end{subfigure}
    \begin{subfigure}[h]{\htwo\linewidth}
        \includegraphics[width=\linewidth]{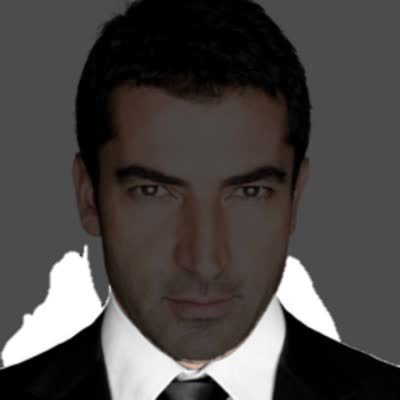}
    \end{subfigure}
    \begin{subfigure}[h]{\htwo\linewidth}
        \includegraphics[width=\linewidth]{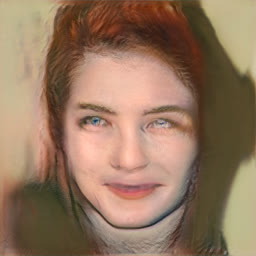}
    \end{subfigure}
    \begin{subfigure}[h]{\htwo\linewidth}
        \includegraphics[width=\linewidth]{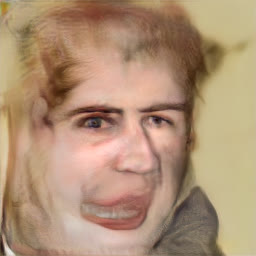}
    \end{subfigure}
    \begin{subfigure}[h]{\htwo\linewidth}
        \includegraphics[width=\linewidth]{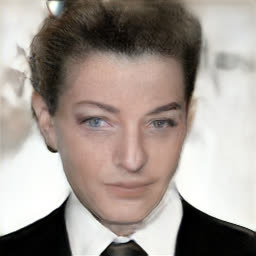}
    \end{subfigure}    
    \begin{subfigure}[h]{\htwo\linewidth}
        \includegraphics[width=\linewidth]{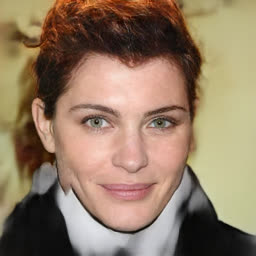}
    \end{subfigure}
    \begin{subfigure}[h]{\htwo\linewidth}
        \centering
        \includegraphics[width=\linewidth]{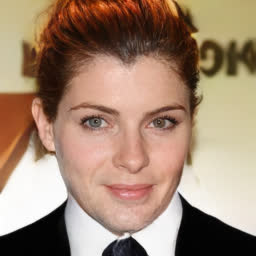}
    \end{subfigure}\\
\caption{
Local editing comparison on CelebA-HQ. The first two baselines~\cite{alharbi2020structurednoise,collins2020editinginstyle} even fail to preserve the untouched region. In-DomainGAN loses a lot of the original image's identity and poorly blends the two images, leaking colors to faces, hair, or background, respectively. 
SEAN locally transfers coarse structure and color but significantly loses details. Ours seamlessly transplants the target region from the reference to the original.}
\label{fig:local_editing_comparison_celeb}
\end{figure*}
}

\newcommand{\figComparisonAFHQ}{
\newcommand{\hthree}{0.16}
\begin{figure*}[t]
    \centering
    \makebox[\hthree\linewidth][c]{\small{~~~Original}}\hfill
    \makebox[\hthree\linewidth][c]{\small{~Reference}}\hfill
    \makebox[\hthree\linewidth][c]{\small{Structured Noise}}\hfill
    \makebox[\hthree\linewidth][c]{\small{Editing in Style~}}\hfill
    \makebox[\hthree\linewidth][c]{\small{In-DomainGAN~}}\hfill
    \makebox[\hthree\linewidth][c]{\small{Ours~~~~}}\hfill 
    \\
    \begin{subfigure}[h]{\hthree\linewidth}
        \includegraphics[width=\linewidth]{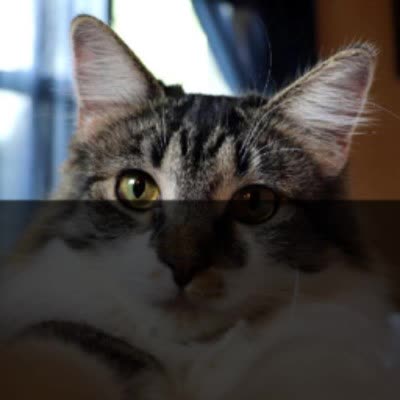}
    \end{subfigure}
    \begin{subfigure}[h]{\hthree\linewidth}
        \includegraphics[width=\linewidth]{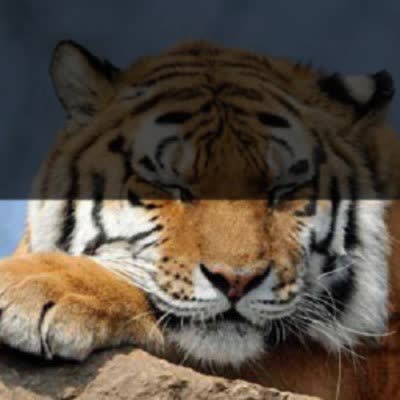}
    \end{subfigure}
    \begin{subfigure}[h]{\hthree\linewidth}
        \includegraphics[width=\linewidth]{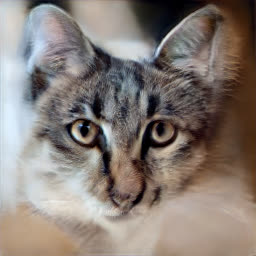}
    \end{subfigure}
    \begin{subfigure}[h]{\hthree\linewidth}
        \includegraphics[width=\linewidth]{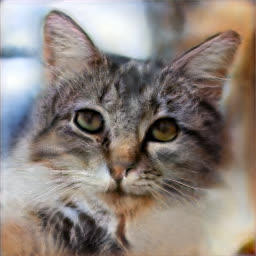}
    \end{subfigure}
    \begin{subfigure}[h]{\hthree\linewidth}
        \includegraphics[width=\linewidth]{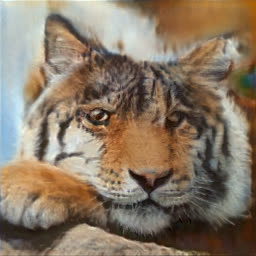}
    \end{subfigure}    
    \begin{subfigure}[h]{\hthree\linewidth}
        \centering
        \includegraphics[width=\linewidth]{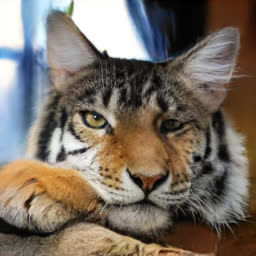}
    \end{subfigure}\\

    \begin{subfigure}[h]{\hthree\linewidth}
        \includegraphics[width=\linewidth]{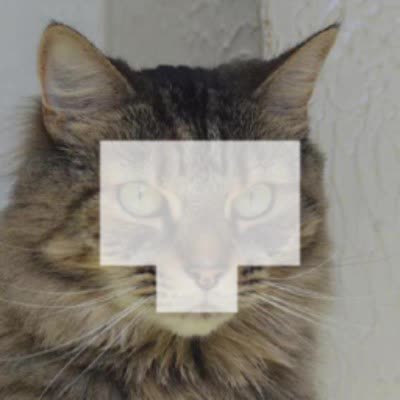}
    \end{subfigure}
    \begin{subfigure}[h]{\hthree\linewidth}
        \includegraphics[width=\linewidth]{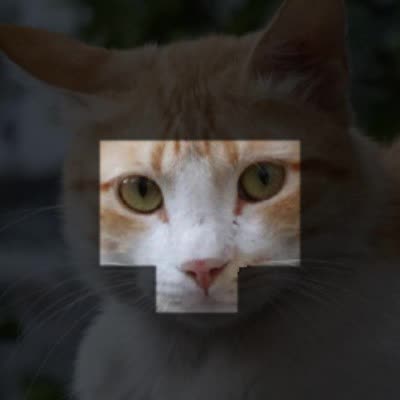}
    \end{subfigure}
    \begin{subfigure}[h]{\hthree\linewidth}
        \includegraphics[width=\linewidth]{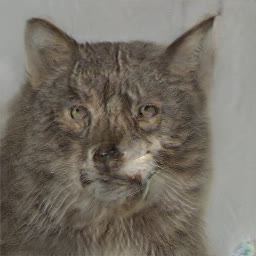}
    \end{subfigure}
    \begin{subfigure}[h]{\hthree\linewidth}
        \includegraphics[width=\linewidth]{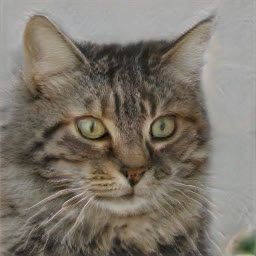}
    \end{subfigure}
    \begin{subfigure}[h]{\hthree\linewidth}
        \includegraphics[width=\linewidth]{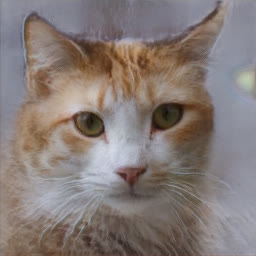}
    \end{subfigure}    
    \begin{subfigure}[h]{\hthree\linewidth}
        \centering
        \includegraphics[width=\linewidth]{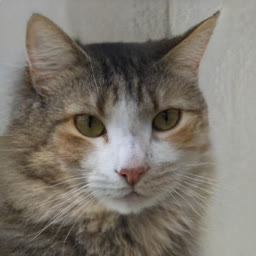}
    \end{subfigure}\\
\caption{
Local editing comparison on AFHQ. Each row blends the two images with horizontal and custom masks, respectively.
Our method seamlessly composes two species with well-preserved details resulting in novel creatures, while others tend to lean towards one species.
}

\label{fig:local_editing_comparison_afhq}
\end{figure*}
}

\newcommand{\figunaligned}{
\newcommand{\h}{26.5mm}
\newcommand{\hh}{2.5mm}
\begin{figure*}[t]
\centering
\begin{minipage}[t]{\linewidth}
\makebox[\hh][c]{}\hspace{0mm}%
\makebox[\h][c]{\textbf{\footnotesize{\ Reference}}}\hspace{3mm}
\rotatebox[origin=l]{90}{\makebox[0mm][l]{\hspace*{0.045\linewidth}\textbf{\footnotesize{\raisebox{0.5mm}[0mm][0mm]{Original}}}}}\hspace{0.8mm}%
\includegraphics[height=\h]{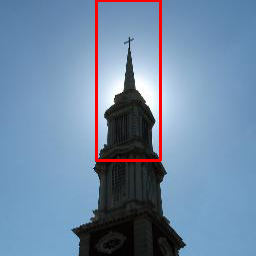}
\includegraphics[height=\h]{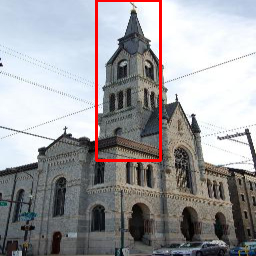}
\includegraphics[height=\h]{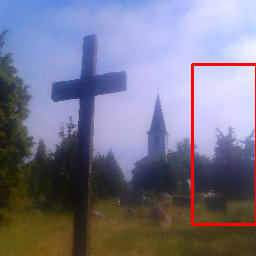}
\includegraphics[height=\h]{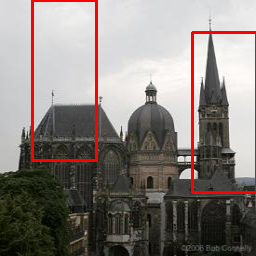}
\includegraphics[height=\h]{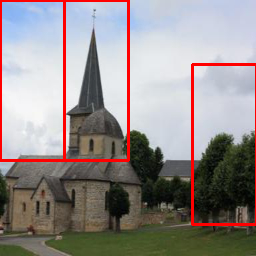}\vspace{1.mm}\\
\makebox[\hh][c]{}\hspace{0.3mm}%
\includegraphics[height=\h]{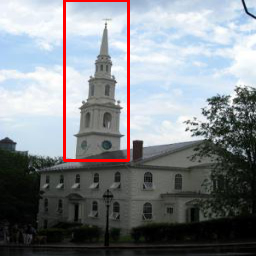}\hspace{3.5mm}
\includegraphics[height=\h]{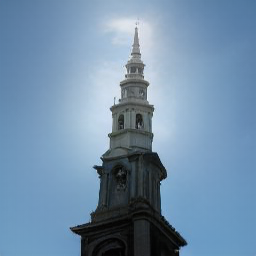}
\includegraphics[height=\h]{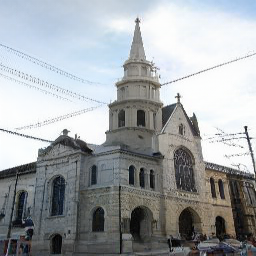}
\includegraphics[height=\h]{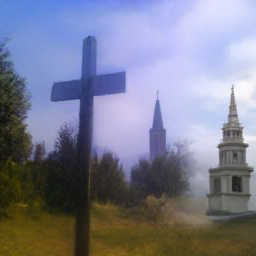}
\includegraphics[height=\h]{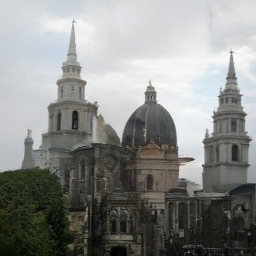}
\includegraphics[height=\h]{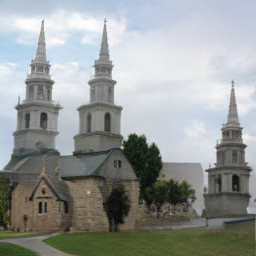}\vspace{1.5mm}\\
\makebox[\hh][c]{}\hspace{0mm}%
\makebox[\h][c]{\textbf{\footnotesize{\ Reference}}}\hspace{3mm}
\rotatebox[origin=l]{90}{\makebox[0mm][l]{\hspace*{0.045\linewidth}\textbf{\footnotesize{\raisebox{.5mm}[0mm][0mm]{Original}}}}}\hspace{0.8mm}%
\includegraphics[height=\h]{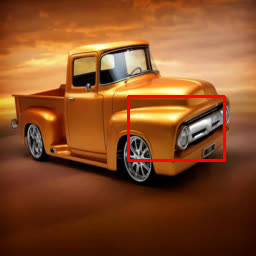}
\includegraphics[height=\h]{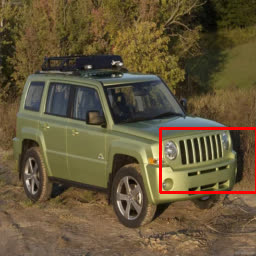}
\includegraphics[height=\h]{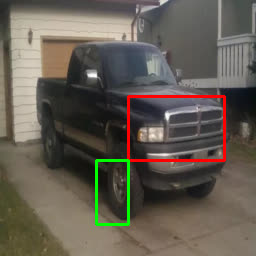}
\includegraphics[height=\h]{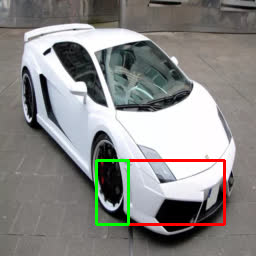}
\includegraphics[height=\h]{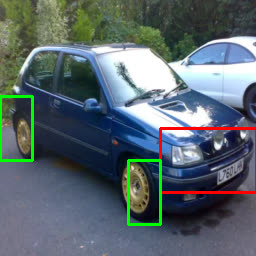}\vspace{1.mm}\\
\makebox[\hh][c]{}\hspace{0.5mm}%
\includegraphics[height=\h]{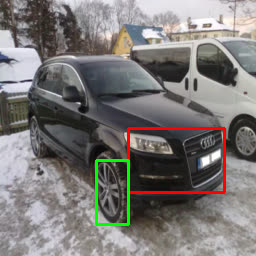}\hspace{3.3mm}
\includegraphics[height=\h]{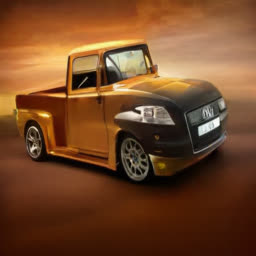}
\includegraphics[height=\h]{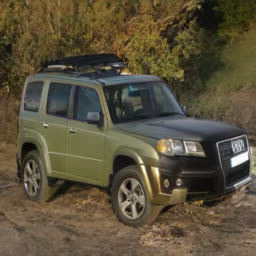}
\includegraphics[height=\h]{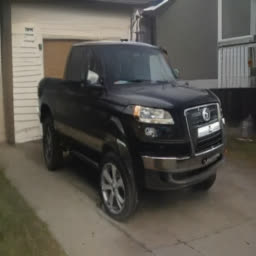}
\includegraphics[height=\h]{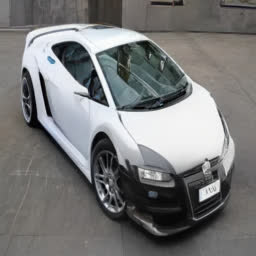}
\includegraphics[height=\h]{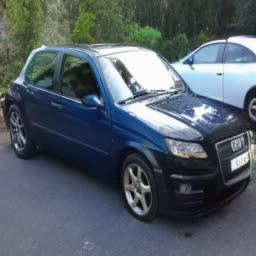}\\
\end{minipage}
\caption{\label{fig:unaligned}%
Examples of unaligned transplantation. \stylemapgan allows composing arbitrary number of any regions. The size and pose of the tower, bumper and wheels are automatically adjusted regarding the surroundings. The masks are specified on $8\times8$ grid and the \stylemaps are blended on $\w$ space. The first row shows an example of copying one area of the reference image into multiple areas of the original images. The second row shows another example of copying two areas of the reference image. Our method can transplant the arbitrary number and size of areas of reference images.
}
\end{figure*}
}

\newcommand{\tabAblationReconst}{
\newcommand{\himg}{0.16}
\begin{table*}[t]
\centering

\vspace{-2mm}
\begin{minipage}{\linewidth}
    \centering
    \makebox[\himg\linewidth][c]{~~~Input}\hfill
    \makebox[\himg\linewidth][c]{~$1\times1$}\hfill
    \makebox[\himg\linewidth][c]{$4\times4$}\hfill
    \makebox[\himg\linewidth][c]{$8\times8$}\hfill
    \makebox[\himg\linewidth][c]{$16\times16$~}\hfill
    \makebox[\himg\linewidth][c]{$32\times32$~~~}\hfill 
    \includegraphics[width=\himg\linewidth]{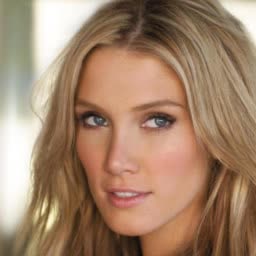}
    \includegraphics[width=\himg\linewidth]{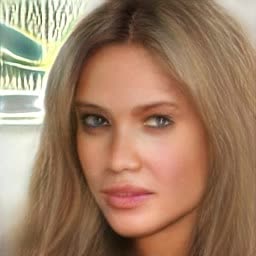}
    \includegraphics[width=\himg\linewidth]{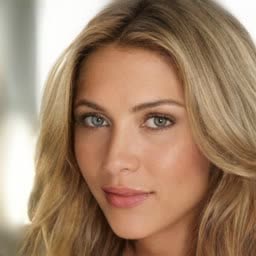}
    \includegraphics[width=\himg\linewidth]{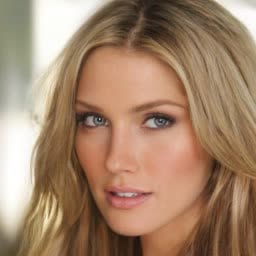}
    \includegraphics[width=\himg\linewidth]{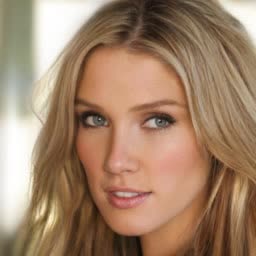}
    \includegraphics[width=\himg\linewidth]{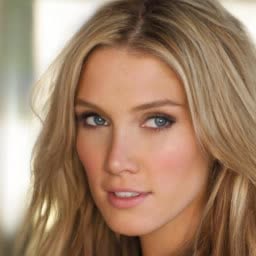}
    
    \hspace{-0.3mm}
    \includegraphics[width=\himg\linewidth]{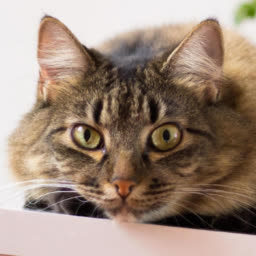}
    \includegraphics[width=\himg\linewidth]{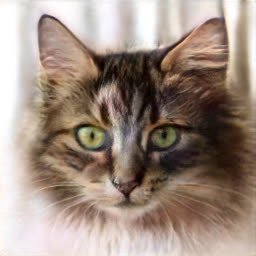}
    \includegraphics[width=\himg\linewidth]{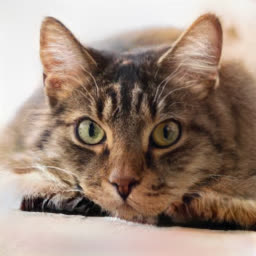}
    \includegraphics[width=\himg\linewidth]{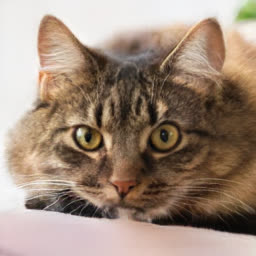}
    \includegraphics[width=\himg\linewidth]{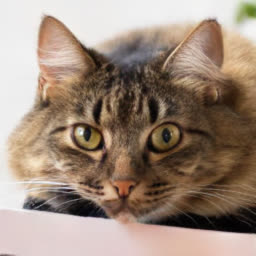}
    \includegraphics[width=\himg\linewidth]{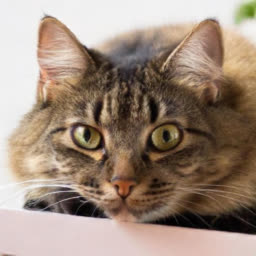}    
    \vspace{2mm}
\end{minipage}
\newcolumntype{x}{>{\centering\arraybackslash\hspace{0pt}}p{12mm}}
\begin{tabular}{lccxxxxxx}
\toprule
\multirow{2}{*}{\textbf{Method}}
    & \multicolumn{1}{c}{\textbf{Style}}
    & \multirow{2}{*}{\textbf{Runtime (s)}}
    & \multicolumn{3}{c}{\textbf{CelebA-HQ}}
    & \multicolumn{3}{c}{\textbf{AFHQ}}
\\
 
& \textbf{resolution}
& 
    & MSE & LPIPS & FID
    & MSE & LPIPS & FID
\\ 
\cmidrule(lr){0-2}
\cmidrule(lr){4-6}
\cmidrule(lr){7-9}
\\[-0.8em]
StyleGAN2 \ \                   
    & 1$ \times $1
    & 0.030
          & 0.089    & 0.428 & 4.97
          & 0.139    & 0.539 & 8.59
\\[0.3mm]
\cmidrule(lr){0-2}
\cmidrule(lr){4-6}
\cmidrule(lr){7-9}
StyleMapGAN \ \                             
    & 4$ \times $4 
    & 0.085
      & 0.062    & 0.351    & \textbf{4.03}
      & 0.070    & 0.394    & 14.82
\\[0.3mm] 
StyleMapGAN \ \
    & 8$ \times $8  
    & 0.082
      & 0.023    & 0.237    & 4.72
      & 0.037    & 0.304    & 11.10
\\[0.3mm]
StyleMapGAN \ \
    & 16$ \times $16
    & 0.078
      & 0.010   & 0.146    & 4.71
     & 0.016    &  0.183    & \textbf{6.71}
\\[0.3mm]
StyleMapGAN  \ \
    & 32$ \times $32 
    &0.074
      & \textbf{0.004}    & \textbf{0.076}    & 7.18
      & \textbf{0.006}    & \textbf{0.090}        & 7.87  %
\\
\bottomrule
\end{tabular}\vspace{0mm}
\caption{
Comparison of reconstruction and generation quality across different resolutions of the \stylemap. 
The higher resolution helps accurate reconstruction, validating the effectiveness of \stylemap. We observe that $8\times8$ \stylemap already provides accurate enough reconstruction and accuracy gain, and afterward, improvements get visually negligible. Although FID varies differently across datasets, possibly due to the different contextual relationships between locations for generation, the \stylemap does not seriously harm the images' quality; rather, it is even better in some configurations. Using our encoder and generator, total inference time is less than 0.1s with almost perfectly reconstructed images. Although StyleGAN2 with our encoder is faster than StyleMapGAN, but it suffers from poorly reconstructed images (second column).
}
\label{tab:ablation_reconst}
\end{table*}
}

\newcommand{\tabBaselineReconst}{
\renewcommand{\himg}{0.131}
\newcommand{\himgtwo}{0.12}
\begin{table*}[t]
\centering
\begin{minipage}{\linewidth}
    \centering
    \hspace{6mm}
    \makebox[\himg\linewidth][c]{Input A}
    \makebox[\himg\linewidth][c]{Inversion A}
    \makebox[\himg\linewidth][c]{$\xleftarrow[]{~~~~~~~~~~~~~~~~~~~}$}\hfill
    \makebox[\himg\linewidth][c]{Interpolation}\hfill
    \makebox[\himg\linewidth][c]{$\xrightarrow[]{~~~~~~~~~~~~~~~~~~~}$~~}\hfill
    \makebox[\himg\linewidth][c]{Inversion B~~~~}\hfill 
    \makebox[\himg\linewidth][c]{Input B~~~~~~}\hfill 
    
    \rotatebox{90}{\makebox[20mm][c]{\small{~~~~Im2StyleGAN}}}\vspace{-0.4mm}\hspace{0.5mm}
    \includegraphics[width=\himg\linewidth]{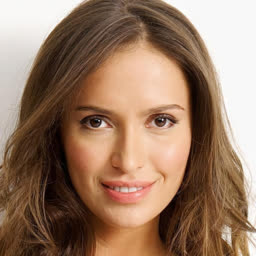}
    \includegraphics[width=\himg\linewidth]{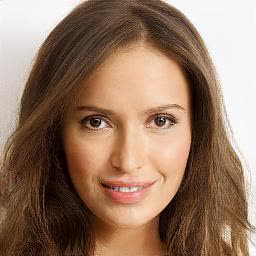}
    \includegraphics[width=\himg\linewidth]{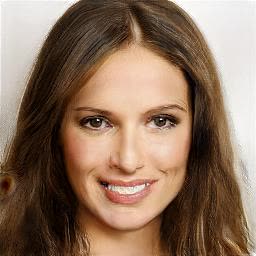}
    \includegraphics[width=\himg\linewidth]{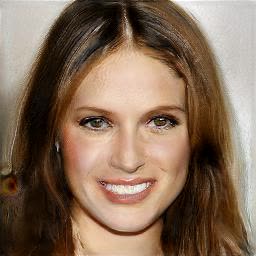}
    \includegraphics[width=\himg\linewidth]{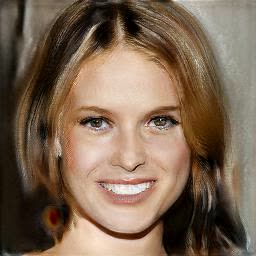}
    \includegraphics[width=\himg\linewidth]{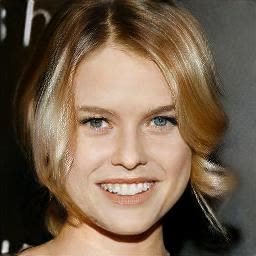}
    \includegraphics[width=\himg\linewidth]{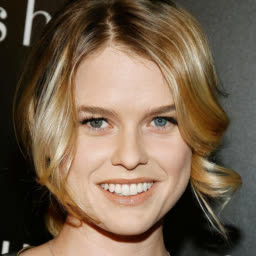}
    
    \hspace{0.5mm}
     \rotatebox{90}{\makebox[20mm][c]{\small{Ours}}}\vspace{0mm}\hspace{0.65mm}
    \includegraphics[width=\himg\linewidth]{figures/3.3.reconstruction/Ours/src.jpg}
    \includegraphics[width=\himg\linewidth]{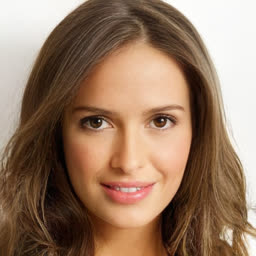}
    \includegraphics[width=\himg\linewidth]{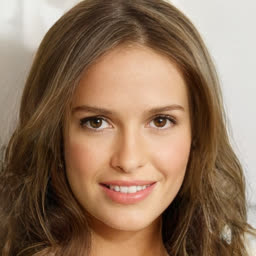}
    \includegraphics[width=\himg\linewidth]{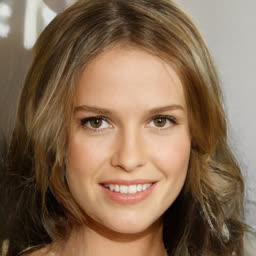}
    \includegraphics[width=\himg\linewidth]{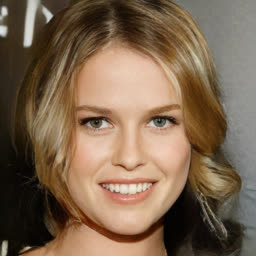}
    \includegraphics[width=\himg\linewidth]{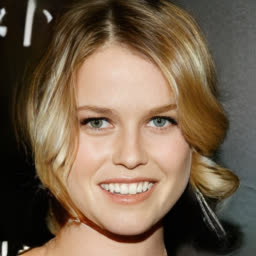}
    \includegraphics[width=\himg\linewidth]{figures/3.3.reconstruction/Ours/ref.jpg}    
    \vspace{2mm}
\end{minipage}
\newcolumntype{x}{>{\centering\arraybackslash\hspace{0pt}}p{10.5mm}}
\begin{tabular}{lcxxxxxx}
\toprule
\multirow{2}{*}{\textbf{Method}}
& \multirow{2}{*}{\textbf{Runtime (s)}}
    & \multicolumn{3}{c}{\textbf{CelebA-HQ}}
    & \multicolumn{3}{c}{\textbf{AFHQ}}
\\
&
   & MSE & LPIPS & \FIDlerp
   & MSE & LPIPS & \FIDlerp
\\ 
\cmidrule(lr){0-1}
\cmidrule(lr){1-2}
\cmidrule(lr){3-5}
\cmidrule(lr){6-8}
\\[-0.8em]
StyleGAN2~\cite{karras2020stylegan2} \ \                  
    & 80.4
    & 0.079    & 0.247    & 30.30 %
    & 0.091        & 0.288    & \textbf{13.87} %
\\[0.3mm]
Image2StyleGAN~\cite{abdal2019image2stylegan} \ \
    & 192.5
      & \textbf{0.009}     & \textbf{0.203}    & 23.68 %
         & \textbf{0.018}         & \textbf{0.282}     & 40.80 %
\\[0.3mm] 
Structured Noise~\cite{alharbi2020structurednoise} \ \
    & 64.4
       & 0.097    & 0.256    & 27.96 %
       & 0.144        & 0.332     & 34.99 %
\\[0.3mm]
In-DomainGAN~\cite{zhu2020indomaingan} \ \
    & 6.8
      & 0.052   & 0.340    & \textbf{22.05} %
         & 0.077       & 0.414    & 17.54 %
\\[0.3mm]
\cmidrule(lr){0-1}
\cmidrule(lr){1-2}
\cmidrule(lr){3-5}
\cmidrule(lr){6-8}
SEAN~\cite{zhu2020sean}  \ \ 
    & 0.146
       & 0.064     & 0.334    & 30.29 %
    & N/A       & N/A       & N/A         %
\\[0.3mm]
StyleMapGAN (Ours, $8 \times 8$) \ \
    & \textbf{0.082}
      & \textbf{0.024}   & \textbf{0.242}    & \textbf{9.97} %
      & \textbf{0.037}    & \textbf{0.304}    & \textbf{12.42} %
\\[0.3mm]

\bottomrule
\end{tabular}\vspace{0mm}
\caption{
Comparison with the baselines for real image projection.
Runtime covers the end-to-end interval of projection and generation in seconds. \FIDlerp measures the quality of the images interpolated on the style space as a proxy for the potential quality of the manipulated images.
Our method allows real-time manipulation of real images while achieving the best reconstruction accuracy and the best quality of the interpolated images. Although Image2StyleGAN produces the smallest reconstruction error, it suffers from minutes of runtime and poor interpolation quality, which are not suitable for practical editing. Its flaws can be found in the figure: rugged details in overall images, especially in teeth. SEAN is not applicable to AFHQ because it requires segmentation masks for training which are not available.
The horizontal line between methods separates optimization-based methods and encoder-based methods.
}\vspace{4mm}
\label{tab:baseline_reconst}
\end{table*}
}

\newcommand{\tabBaselineEdit}{
\begin{table*}[t!]
\centering
\newcolumntype{x}{>{\centering\arraybackslash\hspace{0pt}}p{11mm}}
\begin{tabular}{lcxxxxxx}
\toprule
\multirow{2}{*}{\textbf{Method}}
& \multirow{2}{*}{\textbf{Runtime (s)}}
    & \multicolumn{3}{c}{\textbf{CelebA-HQ}}
    & \multicolumn{3}{c}{\textbf{AFHQ}}
\\
 
&
    & AP & MSE\textsubscript{src} & MSE\textsubscript{ref}
    & AP & MSE\textsubscript{src} & MSE\textsubscript{ref}
\\ 
\cmidrule(lr){0-1}
\cmidrule(lr){1-2}
\cmidrule(lr){3-5}
\cmidrule(lr){6-8}
\\[-0.8em]

Structured Noise~\cite{alharbi2020structurednoise} \ \                 
    & 64.4
    & 99.16      & 0.105     & 0.395
    & 99.88      & 0.137   & 0.444
\\[0.3mm]
Editing in Style~\cite{collins2020editinginstyle} \ \
    & 55.6
    & 98.34      & 0.094     & 0.321
    & 99.52      & 0.130   & 0.417
\\[0.3mm] 
In-DomainGAN~\cite{zhu2020indomaingan} \ \
    & 6.8
    & 98.72      & 0.164   & \textbf{0.015}
    & 99.59      & 0.172   & \textbf{0.028}
\\[0.3mm] 
\cmidrule(lr){0-1}
\cmidrule(lr){1-2}
\cmidrule(lr){3-5}
\cmidrule(lr){6-8}
SEAN~\cite{zhu2020sean} \ \
    & 0.155
    & 90.41       & 0.067    & 0.141
    & N/A        & N/A      & N/A 
\\[0.3mm]
StyleMapGAN (Ours, $8 \times 8$) \ \
    & \textbf{0.099}
    & \textbf{83.60}      & \textbf{0.039}   & \textbf{0.105}
    & \textbf{98.66}      & \textbf{0.050}    & \textbf{0.050}
\\[0.3mm]
\bottomrule
\end{tabular}\vspace{0mm}
\caption{
Comparison with the baselines for local image editing. Average precision (AP) is measured with the binary classifier trained on real and fake images~\cite{wang2020cnndetector}. Low AP shows our edited images are more indistinguishable from real images than other baselines.
Low MSE\textsubscript{src} and MSE\textsubscript{ref} imply that our model preserves the identity of the original image and brings the characteristics of the reference image well, respectively.
Our method outperforms in all metrics except MSE\textsubscript{ref} in In-DomainGAN. 
In-DomainGAN uses masked optimization, which only optimizes the target mask so that the identity of the original image has a great loss as shown in Figures \ref{fig:local_editing_comparison_celeb} and \ref{fig:local_editing_comparison_afhq}.
}
\label{tab:baseline_edit}
\end{table*}
}

\newcommand{\tabLosses}{
\begin{table}[t]
\centering
\newcolumntype{x}{>{\centering\arraybackslash\hspace{0pt}}p{20mm}}
\bgroup
\begin{tabular}{ccccc}
\toprule
Loss & G & D & E
\\ 
\hline
Adversarial loss~\cite{goodfellow2014gan} & \cmark & \cmark &\\
R\textsubscript{1} regularization~\cite{mescheder2018r1reg} &  & \cmark &\\
Latent reconstruction & & & \cmark\\
Image reconstruction & \cmark & & \cmark\\
Perceptual loss~\cite{zhang2018lpips} & \cmark & & \cmark \\
Domain-guided loss~\cite{zhu2020indomaingan} & \cmark & \cmark & \cmark \\
\bottomrule
\end{tabular}
\egroup
\caption{
Losses for training each network. Non-saturating loss \cite{goodfellow2014gan} is used as the adversarial loss. R1-regularization is applied every 16 steps~\cite{karras2020stylegan2} for D to stabilize training. Latent reconstruction loss is mean squared error (MSE) in the $\w$ space. Image reconstruction is MSE in image pixel-level space. We use learned perceptual image patch similarity (LPIPS) as perceptual loss for calculating perceptual differences between original and reconstructed images. Domain-guided loss is related to adversarial training that reconstructed images from the encoder tries to be classified as real by the discriminator.
}
\label{tab:losses}
\end{table}
}

\usepackage{array,multirow,makecell,tabularx}
\newcommand{\arch}[1]{\textsc{#1}}
\usepackage{booktabs}

\usepackage{caption}
\usepackage{subcaption}
\usepackage{xspace}

\newcommand{\Fref}[1]{Figure \ref{#1}}

\newcommand{\sref}[1]{\S\ref{#1}}
\newcommand{\Tref}[1]{Table \ref{#1}}

\definecolor{mygreen}{rgb}{0.032, 0.6392, 0.2039}
\newcommand{\cmark}{\textcolor{mygreen}{\ding{51}}}
\newcommand{\xmark}{\textcolor{red}{\ding{55}}}

\begin{document}

\title{Exploiting Spatial Dimensions of Latent in GAN for Real-time Image Editing}

\author{Hyunsu Kim\textsuperscript{1}\textsuperscript{2} \thinspace\quad Yunjey Choi\textsuperscript{1} \thinspace\quad Junho Kim\textsuperscript{1} \thinspace\quad Sungjoo Yoo\textsuperscript{2} \thinspace\quad Youngjung Uh\textsuperscript{3} \\
\\
\textsuperscript{1}NAVER AI Lab \quad  \textsuperscript{2}Seoul National University \quad  \textsuperscript{3}Yonsei University 
}

\twocolumn[{
\maketitle
\begin{center}
    \vspace{-5mm}
    
    \includegraphics[width=1\textwidth]{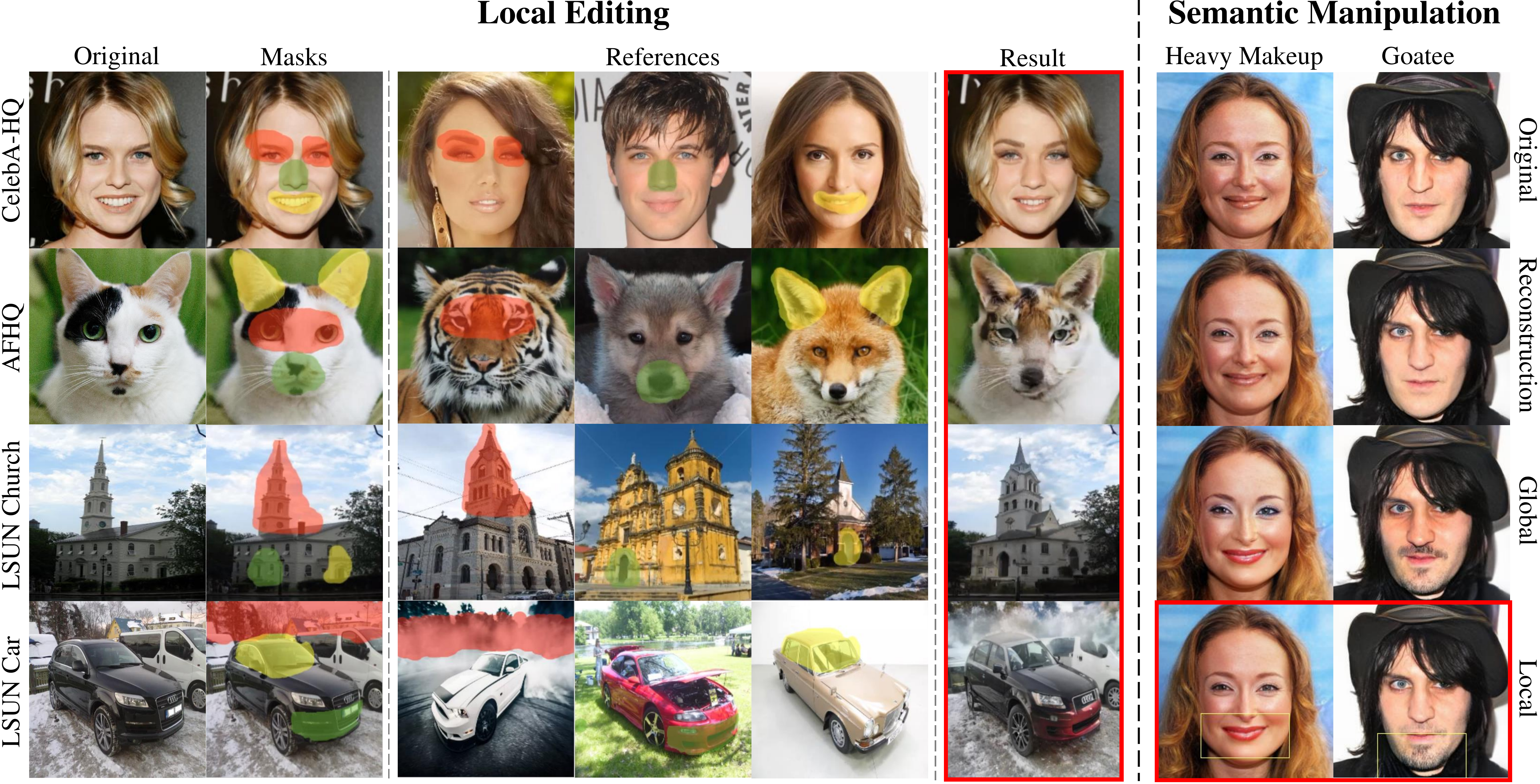}
    \captionof{figure}{Various image editing results on multiple datasets. Local editing mixes multiple parts of reference images with the original image. Unlike other methods (\textit{Global} case), ours can do semantic manipulation locally (Yellow box, \textit{Local} case).} 
    \label{fig:teaser}
\end{center}
}]

\begin{abstract}

Generative adversarial networks (GANs) synthesize realistic images from random latent vectors. Although manipulating the latent vectors controls the synthesized outputs, editing real images with GANs suffers from i) time-consuming optimization for projecting real images to the latent vectors, ii) or inaccurate embedding through an encoder. We propose StyleMapGAN: the intermediate latent space has spatial dimensions, and a spatially variant modulation replaces AdaIN. It makes the embedding through an encoder more accurate than existing optimization-based methods while maintaining the properties of GANs. Experimental results demonstrate that our method significantly outperforms state-of-the-art models in various image manipulation tasks such as local editing and image interpolation. Last but not least, conventional editing methods on GANs are still valid on our StyleMapGAN. Source code is available at \url{https://github.com/naver-ai/StyleMapGAN}.
\end{abstract}

\figdemo
\section{Introduction}

Generative adversarial networks (GANs)~\cite{goodfellow2014gan} have evolved dramatically in recent years, enabling high-fidelity image synthesis with models that are learned directly from data~\cite{brock2018biggan,karras2019stylegan,karras2020stylegan2}. Recent studies have shown that GANs naturally learn to encode rich semantics within the latent space, thus changing the latent code leads to manipulating the corresponding attributes of the output images~\cite{jahanian2019steerability,shen2020interfacegan,harkonen2020ganspace,goetschalckx2019gananalyze, shen2020closedformgan,alharbi2020structurednoise, voynov2020unsupervised,bau2018gan}. However, it is still challenging to apply these manipulations to real images since the GAN lacks an inverse mapping from an image back to its corresponding latent code. 

One promising approach for manipulating real images is image-to-image translation~\cite{isola2017pix2pix,zhu2017cyclegan,choi2018stargan, kim2019tag2pix, Kim2020U-GAT-IT:}, where the model learns to synthesize an output image given a user's input directly. However, these methods require pre-defined tasks and heavy supervision (\eg, input-output pairs, class labels) for training and limit the user controllability at inference time. Another approach is to utilize pretrained GAN models by directly optimizing the latent code for an individual image \cite{abdal2019image2stylegan,abdal2020image2stylegan,zhu2016igan,ma2018invertibility,noguchi2019smalldatagan}. However, even on high-end GPUs, it requires minutes of computation for each target image, and it does not guarantee that the optimized code would be placed in the original latent space of GAN.

A more practical approach is to train an extra encoder, which learns to project an image into its corresponding latent code~\cite{larsen2016vaegan,zhu2020indomaingan,perarnau2016icgan,luo2017autoencodergan,pidhorskyi2020alae}. Although this approach enables real-time projection in a single feed-forward manner, it suffers from the low fidelity of the projected image (\ie, losing details of the target image). We attribute this limitation to the absence of spatial dimensions in the latent space. Without the spatial dimensions, an encoder compresses an image's local semantics into a vector in an entangled manner, making it difficult to reconstruct the image (\eg, vector-based or low-resolution bottleneck layer is not capable of producing high-frequency details~\cite{lample2017fadernetwork,chang2018pairedcyclegan}).

As a solution to such problems, we propose \stylemapgan which exploits \textit{\stylemap}, a novel representation of the latent space. Our key idea is simple. Instead of learning a vector-based latent representation, we utilize a tensor with explicit spatial dimensions. Our proposed representation benefits from its spatial dimensions, enabling GANs to easily encode the local semantics of images into the latent space. This property allows an encoder to effectively project an image into the latent space, thus providing high-fidelity and real-time projection. Our method also offers a new capability to edit specific regions of an image by manipulating the matching positions of the \stylemap. \Fref{fig:teaser} shows our local editing and local semantic manipulation results. Note that all editing is done in real-time. As shown in \Fref{fig:demo}, you can test our web demo to do interactive editing.

On multiple datasets, our \stylemap indeed substantially enhances the projection quality compared to the traditional vector-based latent representation~(\sref{sec:ablation}). Furthermore, we show the advantage of our method over state-of-the-art methods on image projection, interpolation, and local editing~(\sref{sec:real_image_projection} \& \sref{sec:local_editing}). Finally, we show that our method can transplant regions even when the regions are not aligned between one image and another~(\sref{sec:unaligned}).%

\section{Related work}

\noindent\textbf{Optimization-based editing methods} iteratively update the latent vector of pre-trained GANs to project a real image into the latent space~\cite{zhu2016igan, brock2016neuralphotoediting,abdal2019image2stylegan,zhu2020indomaingan,huh2020pix2latent,Bau_2019}. For example, Image2StyleGAN~\cite{abdal2019image2stylegan} reconstructs the image by optimizing intermediate representation for each layer of StyleGAN~\cite{karras2019stylegan}. In-DomainGAN~\cite{zhu2020indomaingan} focuses not only on reconstructing the image in pixel space, but also on landing the inverted code in the semantic domain of the original latent space. Neural Collage~\cite{suzuki2018neuralcollage} and pix2latent~\cite{huh2020pix2latent} present a hybrid optimization strategy for projecting an  image into the latent space of class-conditional GANs~(\eg, BigGAN~\cite{brock2018biggan}). On the other hand, we exploit an encoder, which makes editing two to three orders of magnitude faster than optimization methods.

\medskip

\noindent\textbf{Learning-based editing methods} train an extra encoder to directly infer the latent code given a target image~\cite{larsen2016vaegan,donahue2019bigbigan,donahue2016bigan, dumoulin2016ali,pidhorskyi2020alae}. For example, ALI~\cite{dumoulin2016ali} and BiGAN~\cite{donahue2016bigan} introduce a fully adversarial framework to jointly learn the generator and the inverse mapping. Several work~\cite{larsen2016vaegan,srivastava2017veegan,ulyanov2017generatorencoder} has been made towards combining the variational autoencoder~\cite{kingma2013vae} with GANs for latent projection. ALAE~\cite{pidhorskyi2020alae} builds an encoder to predict the intermediate latent space of StyleGAN. However, all the above methods provide limited reconstruction quality due to the lack of spatial dimensions of latent space. Swap Autoencoder~\cite{park2020swapping} learns to encode an image into two components, structure code and texture code, and generate a realistic image given any swapped combination. Although it can reconstruct images fast and precisely thanks to such representation, texture code is still a vector, which makes structured texture transfer challenging. Our editing method successfully reflects not only the color and texture but also the shape of a reference image. %

\medskip

\noindent\textbf{Local editing methods} tackle editing specific parts~\cite{collins2020editinginstyle, alharbi2020structurednoise, zhu2020sean, zhan2019spatial, shocher2020semantic} (\eg, nose, background) as opposed to the most GAN-based image editing methods modifying global appearance~\cite{shen2020interfacegan, voynov2020unsupervised, park2020swapping}. For example, Editing in Style~\cite{collins2020editinginstyle} tries to identify each channel's contribution of the per-layer style vectors to specific parts. Structured Noise~\cite{alharbi2020structurednoise} replaces the learned constant from StyleGAN with an input tensor, which is a combination of local and global codes. However, these methods~\cite{collins2020editinginstyle, alharbi2020structurednoise, bau2018gan} do not target real image, which performances are degraded significantly in the real image.
SEAN~\cite{zhu2020sean} facilitates editing real images by encoding images into the per-region style codes and manipulating them, but it requires pairs of images and segmentation masks for training. Besides, the style code is still a vector, so it has the same problem as Swap Autoencoder~\cite{park2020swapping}.

\section{\stylemapgan}

Our goal is to project images to a latent space accurately with an encoder in real-time and locally manipulate images on the latent space. We propose \stylemapgan which adopts \textit{\stylemap}, an intermediate latent space with spatial dimensions, and a spatially variant modulation based on the \stylemap (\sref{sec:stylemap}). Note that the \textit{style} denotes not only textures (fine style) but also shapes (coarse style) following \cite{karras2019stylegan}. Now an encoder can embed an image to the \stylemap which reconstructs the image more accurately than optimization-based methods, and partial change in the \stylemap leads to local editing on the image (\sref{sec:editing}).

\subsection{Stylemap-based generator}
\label{sec:stylemap}
\fignetarch
\Fref{fig:network_architecture} describes the proposed stylemap-based generator. While a traditional mapping network produces style vectors to control feature maps, we create a stylemap with spatial dimensions, which not only makes the projection of a real image much more effective at inference but also enables local editing. The mapping network has a reshape layer at the end to produce the \stylemap which forms the input to the spatially varying affine parameters. Since the feature maps in the synthesis network grow larger as getting closer to the output image, we introduce a \stylemap resizer, which consists of convolutions and upsampling, to match the resolutions of stylemaps with the feature maps. The \stylemap resizer resizes and transforms the \stylemap with learned convolutions to convey more detailed and structured styles.

Then, the affine transform produces parameters for the modulation regarding the resized \stylemaps. The modulation operation of the $i$-th layer in the synthesis network is as follows:
\begin{equation}
h_{i+1}=\left(\gamma_i \otimes \frac{h_i-\mu_i}{\sigma_i}\right) \oplus \beta_i
\label{eqn::modulation}
\end{equation}
where $\mu_i,\sigma_i \in \R$ are the mean and standard deviation of activations $h_i \in \R^{C_i \times H_i \times W_i}$ of the $i$-th layer, respectively. $\gamma_i,\beta_i \in \R^{C_i \times H_i \times W_i}$ are modulation parameters. $\otimes$ and $\oplus$ are element-wise multiplication and addition, respectively.

We remove per-pixel noise which was an extra source of spatially varying inputs in StyleGAN, because our \stylemap already provides spatially varying inputs and the single input makes the projection and editing simpler. Please see the supplementary material (\sref{sec:implementation_details}) for other details about the network and relationship with the autoencoder approach~\cite{hinton2006reducing}.

\figtraininference

\subsection{Training procedure and losses}
In \Fref{fig:figtraininference}, we use F, G, E, and D to indicate the mapping network, synthesis network with stylemap resizer, encoder, and discriminator, respectively, for brevity. D is the same as StyleGAN2, and the architecture of E is similar to D except without minibatch discrimination~\cite{salimans2016improved}. All networks are jointly trained using multiple losses as shown in \Tref{tab:losses}. G and E are trained to reconstruct real images in terms of both pixel-level and perceptual-level~\cite{zhang2018lpips}. Not only the image but E tries to reconstruct the \stylemap with mean squared error (MSE) when G(F) synthesizes an image from $z$. D attempts to classify the real images and the fake images generated from Gaussian distribution. Lastly, we exploit domain-guided loss for the in-domain property~\cite{zhu2020indomaingan}. E tries to reconstruct more realistic images by competing with D, making projected stylemap more suitable for image editing. If we remove any of the loss functions, generation and editing performance are degraded. Refer to the supplementary material for the effect of each loss function (\sref{sec:contribution_loss}) and joint learning (\sref{sec:1K_comparison}). Further training details (\sref{sec:implementation_details}) are also involved.

\tabLosses

\subsection{Local editing}
\label{sec:editing}

As shown at the bottom of \Fref{fig:figtraininference}, the goal of local editing is to transplant some parts of a reference image to an original image with respect to a mask, which indicates the region to be modified. Note that the mask can be in any shape allowing interactive editing or label-based editing with semantic segmentation methods.

We project the original image and the reference image through the encoder to obtain \stylemaps $\w$ and $\wt$, respectively. The edited \stylemap \textbf{\"w} is an alpha blending of $\w$ and $\wt$:
\begin{equation}
\text{\textbf{\"w}}=\m \otimes {\wt} \oplus (1-\m) \otimes {\w}
\label{eqn::local_editing}
\end{equation}
where the mask $\m$ is shrunk by max pooling, and $\otimes$ and $\oplus$ are the same as Equation~\ref{eqn::modulation}. In general, the mask is finer than $8\times8$, so we blend the \stylemaps on the $\w^+$ space to achieve detailed manipulation. But for simplicity, we explain blending on the $\w$ space; the $\w^+$ space blending method is in the supplementary material (\sref{sec:w_plus_blending}). Unless otherwise stated, local editing figures are blends on the $\w^+$ space.

Contrarily to SPADE~\cite{park2019spade} or SEAN~\cite{zhu2020sean}, even rough masks as coarse as $8\times8$ produces plausible images so that the burden for user to provide detailed masks is lifted. This operation can be further revised for unidentical masks of the two images (\sref{sec:unaligned}).
\section{Experiments}
Our proposed method efficiently projects images into the style space in real-time and effectively manipulates specific regions of real images. We first describe our experimental setup~(\sref{sec:eval_setup}) and evaluation metrics (\sref{sec:eval_metrics}) and show how the proposed spatial dimensions of \stylemap affect the image projection and generation quality~(\sref{sec:ablation}). We then compare our method with the state-of-the-art methods on real image projection~(\sref{sec:real_image_projection}) and local editing~(\sref{sec:local_editing}). We finally show a more flexible editing scenario and the usefulness of our proposed method~(\sref{sec:unaligned}). Please see the supplementary material for high-resolution experiments (\sref{sec:1K_comparison}) and additional results (\sref{sec:additional_results}) such as random generation, style mixing, semantic manipulation, and failure cases.
  
\tabAblationReconst
\figAblationEditing

\subsection{Experimental setup}
\label{sec:eval_setup}
\noindent\textbf{Baselines.} We compare our model with recent generative models, including StyleGAN2~\cite{karras2020stylegan2}, Image2StyleGAN~\cite{abdal2019image2stylegan},  In-DomainGAN~\cite{zhu2020indomaingan}, Structured Noise~\cite{alharbi2020structurednoise}, Editing in Style~\cite{collins2020editinginstyle}, and SEAN~\cite{zhu2020sean}. We train all the baselines from scratch until they converge using the official implementations provided by the authors. For optimization-based methods~\cite{karras2020stylegan2,abdal2019image2stylegan,zhu2020indomaingan,alharbi2020structurednoise,collins2020editinginstyle}, we use all the hyperparameters specified in their papers. We also compare our method with ALAE~\cite{pidhorskyi2020alae} qualitatively in the supplementary material (\sref{sec:supp_reconstruction}). Note that we do not compare our method against Image2StyleGAN++~\cite{abdal2020image2stylegan} and Swap Autoencoder~\cite{park2020swapping}, since the authors have not published their code yet.

\medskip

\noindent\textbf{Datasets.}  We evaluate our model on \celebahq~\cite{karras2017progressivegan}, AFHQ~\cite{choi2020starganv2}, and LSUN Car \& Church~\cite{yu2015lsun}. We adopt CelebA-HQ instead of FFHQ~\cite{karras2019stylegan}, since \celebahq includes segmentation masks so that we can train the SEAN baseline and exploit the masks to evaluate local editing accurately in a semantic level. The AFHQ dataset includes wider variation than the human face dataset, which is suitable for showing the generality of our model. The optimization methods take an extremely long time, we limited the test and validation set to 500 images the same as In-DomainGAN~\cite{zhu2020indomaingan}. The numbers of training images for \celebahq, AFHQ, and LSUN Car \& Church are 29K, 15K, 5.5M, and 126K, respectively. We trained all models at $256\times256$ resolution for comparison in a reasonable time, but we also provide $1024\times1024$ FFHQ results in the supplementary material (\sref{sec:1K_comparison}).

\subsection{Evaluation metrics}
\label{sec:eval_metrics}
\noindent\textbf{Fr\'echet inception distance (FID).} To evaluate the performance of image generation, we calculate FID~\cite{heusel2017fid} between images generated from Gaussian distribution and training set. We set the number of generated samples equal to that of training samples. We use the ImageNet-pretrained Inception-V3~\cite{szegedy2016inceptionv3} for feature extraction.

\smallskip

\noindent\textbf{\FIDlerp.} To evaluate the global manipulation performance, we calculate FID between interpolated samples and training samples (\FIDlerp). To generate interpolated samples, we first project 500 test images into the latent space and randomly choose pairs of latent vectors. We then generate an image using a linearly interpolated latent vector whose interpolation coefficient is randomly chosen between 0 and 1.  We set the number of interpolated samples equal to that of training samples. Low \FIDlerp indicates that the model provides high-fidelity and diverse interpolated samples.

\smallskip

\noindent\textbf{MSE \& LPIPS.} To evaluate the projection quality, we estimate pixel-level and perceptual-level differences between target images and reconstructed images, which are mean square error (MSE) and learned perceptual image patch similarity (LPIPS)~\cite{zhang2018lpips}, respectively.

\smallskip

\noindent\textbf{Average precision (AP).} To evaluate the quality of locally edited images, we measure the average precision with the binary classifier trained on real and fake images~\cite{wang2020cnndetector}, following the convention of the previous work~\cite{park2020swapping}. We use the Blur+JPEG(0.5) model and full images for evaluation. The lower AP indicates that manipulated images are more indistinguishable from real images.

\smallskip

\noindent\textbf{MSE\textsubscript{src} \& MSE\textsubscript{ref}.} 
In order to mix specific semantic, we make merged masks by combining target semantic masks of original and reference images. MSE\textsubscript{src} and MSE\textsubscript{ref} measure mean square error from the original image outside the mask and from the reference image inside the mask, respectively.
To naturally combine them, images are paired by target semantic mask similarity. 
For local editing comparison on \celebahq, 250 sets of test images are paired in each semantic (\eg, background, hair)~\cite{CelebAMask-HQ}, which produces a total of 2500 images. For local editing on AFHQ, 250 sets of test images are paired randomly, and masks are chosen between the horizontal and vertical half-and-half mask, which produces 250 images.

\tabBaselineReconst

\subsection{Effects of stylemap resolution}
\label{sec:ablation}
To manipulate an image using a generative model, we first need to project the image into its latent space accurately. In \Tref{tab:ablation_reconst}, we vary the spatial resolution of \stylemap and compare the performance of reconstruction and generation. For a fair comparison, we train our encoder model after training the StyleGAN2 generator.
As the spatial resolution increases, the reconstruction accuracy improves significantly. 
It demonstrates that our \stylemap with spatial dimensions is highly effective for image projection. FID varies differently across datasets, possibly due to different contextual relationships between locations for a generation.
Note that our method with spatial resolution accurately preserves small details, \eg, the eyes are not blurred.

Next, we evaluate the effect of the \stylemap's resolution in editing scenarios, mixing specific parts of one image and another.
\Fref{fig:ablation_editing} shows that the 8$\times$8 \stylemap synthesizes the most plausible images in terms of seamlessness and preserving the identities of an original and reference image.
We see that when the spatial resolution is higher than $8\times8$, the edited parts are easily detected. 

Furthermore, we estimate \FIDlerp in different resolution models in \celebahq. The $8\times8$ model shows the best \FIDlerp value (9.97) than other resolution models; 10.72, 11.05, and 12.10 for $4\times4$, $16\times16$, and $32\times32$, respectively. We suppose that the larger resolution of \stylemap, the more likely projected latent from the encoder gets out of the latent space, which comes from a standard Gaussian distribution.
Considering the editing quality and \FIDlerp, we choose the 8$\times$8 resolution as our best model and use it consistently for all subsequent experiments.

\subsection{Real image projection}
\label{sec:real_image_projection}

In \Tref{tab:baseline_reconst}, we compare our approach with the state-of-the-art methods for real image projection. For both datasets, \stylemapgan achieves better reconstruction quality (MSE \& LPIPS) than all competitors except Image2StyleGAN. However, Image2StyleGAN fails to meet requirements for editing in that it produces spurious artifacts in latent interpolation (\FIDlerp and figures) and suffers from minutes of runtime. 
Our method also achieves the best \FIDlerp, which implicitly shows that our manipulation on the style space leads to the most realistic images.
Importantly, our method runs at least $100\times$ faster than the optimization-based baselines since a single feed-forward pass provides accurate projection thanks to the \stylemap, which is measured in a single V100 GPU.
SEAN also runs with a single feed-forward pass, but it requires ground-truth segmentation masks for both training and testing, which is a severe drawback for practical uses.

\subsection{Local editing}
\label{sec:local_editing}

We evaluate local editing performance regarding three aspects: detectability, faithfulness to the reference image in the mask, and preservation of the original image outside the mask. Figures \ref{fig:local_editing_comparison_celeb} and \ref{fig:local_editing_comparison_afhq} visually demonstrate that our method seamlessly composes the two images while others struggle. Since there is no metric for evaluating the last two aspects, we propose two quantitative metrics: MSE\textsubscript{src} and MSE\textsubscript{ref}. \Tref{tab:baseline_edit} shows that the results from our method are the hardest for the classifier to detect fakes, and both original and reference images are best reflected. Note that MSEs are not the sole measures, but AP should be considered together for the realness of the image.
\figComparisonCeleb
\figComparisonAFHQ
\tabBaselineEdit
\figunaligned
\subsection{Unaligned transplantation}
\label{sec:unaligned}
Here, we demonstrate a more flexible use case, unaligned transplantation (image blending), showing that our local editing does not require the masks on the original and the reference images to be aligned. We project the images to the \stylemaps and replace the designated region of the original \stylemap with the crop of the reference \stylemap even though they are in different locations. Users can specify what to replace. \Fref{fig:unaligned} shows examples of LSUN Car \& Church.

\section{Discussion and Conclusion}
Invertibility of GANs has been essential for editing real images with unconditional GAN models at a practical time, and it has not been properly answered yet.
To achieve this goal, we propose StyleMapGAN, which introduces explicit spatial dimensions to the latent space, called a \stylemap.
We show that our method based on the stylemap has a number of advantages over prior approaches through an extensive evaluation. It can accurately project real images in real-time into the latent space and synthesize high-quality output images by both interpolation and local editing.
We believe that improving fidelity by applying our latent representation to other methods such as conditional GANs (\eg, BigGAN~\cite{brock2018biggan}) or variational autoencoders~\cite{kingma2013vae} would be exciting future work.

\noindent\textbf{Acknowledgements.} The authors thank NAVER AI Lab researchers for constructive discussion. All experiments were conducted on NAVER Smart Machine Learning (NSML) platform~\cite{kim2018nsml, sung2017nsml}.

\newcommand{\figAblationLoss}{
\begin{figure}[h]
\begin{center}
\includegraphics[width=1.0\columnwidth]{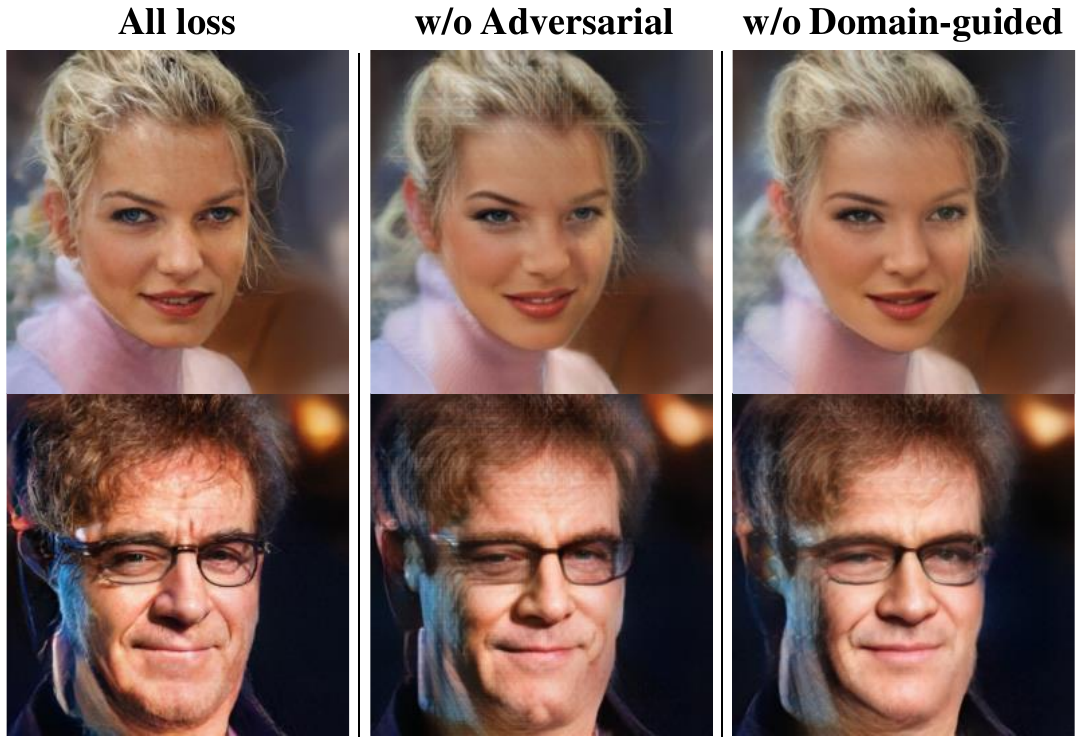}
\end{center}
\caption{This figure shows the interpolation results of different networks. Leftmost images are results of a network trained using whole losses. Second column images are generated by a network trained without random Gaussian noise, which is similar to AutoEncoder. A network trained without domain-guided loss generates rightmost column images.}
\label{fig:figablationloss}
\end{figure}
}

\newcommand{\figeditingimgtostylegantwo}{
\begin{figure*}[t]
\begin{center}
\includegraphics[width=1.0\linewidth]{supp_figures/1024/editing_img2stylegan2.jpg}
\end{center}

\caption{
Edit later
}
\label{fig:figeditingimgtostylegantwo}     
\end{figure*}
}

\newcommand{\mixshape}{
\begin{figure*}[t]
\newcommand{\hfive}{0.195}
    \centering
    \makebox[\hfive\linewidth][c]{Original}\hfill
    \makebox[\hfive\linewidth][c]{Reference}\hfill
    \makebox[\hfive\linewidth][c]{Mask}\hfill
    \makebox[\hfive\linewidth][c]{Copy All}\hfill
    \makebox[\hfive\linewidth][c]{Copy shape}\hfill

    \includegraphics[width=\hfive\linewidth]{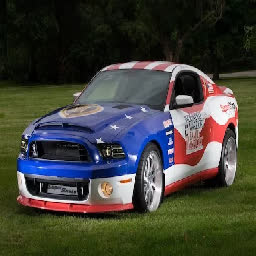}
    \includegraphics[width=\hfive\linewidth]{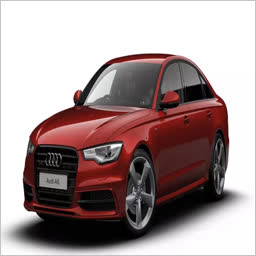}
    \includegraphics[width=\hfive\linewidth]{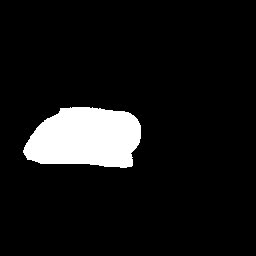}
    \includegraphics[width=\hfive\linewidth]{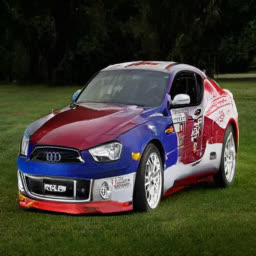}
    \includegraphics[width=\hfive\linewidth]{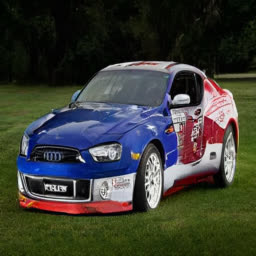}
    
\caption{4th column shows the transplantation of all identity including shape, texture, and color. The rightmost image shows the transplantation of structure alone.}
\label{fig:mix_shape}     
\end{figure*}
}
\newcommand{\figAblationMapping}{
\begin{figure*}[t]
\newcommand{\hfive}{0.195}
    \centering
    \makebox[\hfive\linewidth][c]{Original}\hfill
    \makebox[\hfive\linewidth][c]{Reference}\hfill
    \makebox[\hfive\linewidth][c]{AutoEncoder}\hfill
    \makebox[\hfive\linewidth][c]{Conv}\hfill
    \makebox[\hfive\linewidth][c]{MLP (Ours)}\hfill

    \includegraphics[width=\hfive\linewidth]{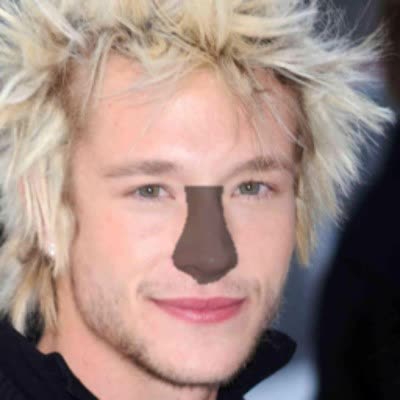}
    \includegraphics[width=\hfive\linewidth]{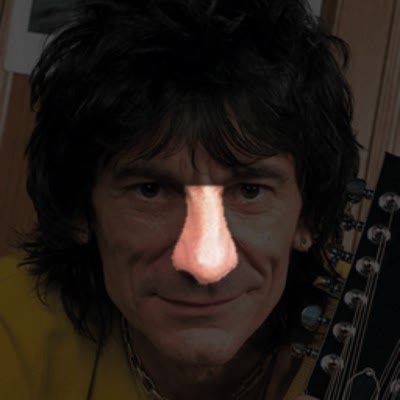}
    \includegraphics[width=\hfive\linewidth]{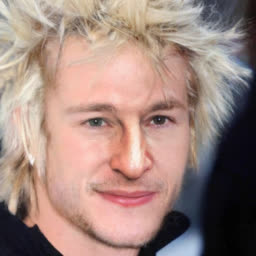}
    \includegraphics[width=\hfive\linewidth]{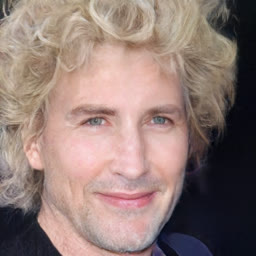}
    \includegraphics[width=\hfive\linewidth]{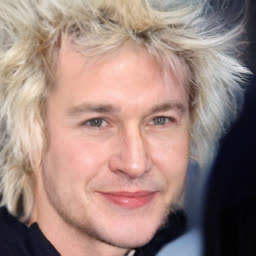}
\\

    \includegraphics[width=\hfive\linewidth]{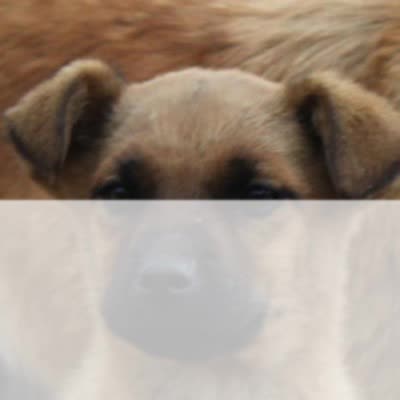}
    \includegraphics[width=\hfive\linewidth]{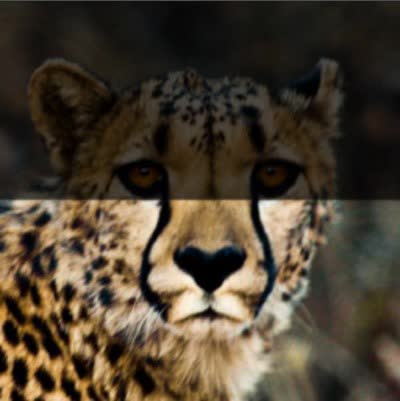}
    \includegraphics[width=\hfive\linewidth]{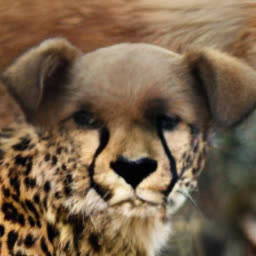}
    \includegraphics[width=\hfive\linewidth]{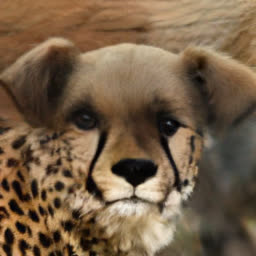}
    \includegraphics[width=\hfive\linewidth]{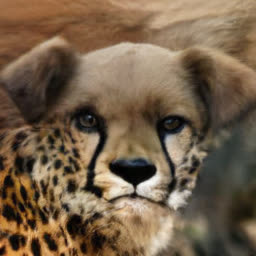}

\caption{
Local editing comparison across different mapping network architectures of StyleMapGAN. MLP-based architecture provides more natural images compared to autoencoder-based and convolution-based architecture.
}
\label{fig:figAblationMapping}     
\end{figure*}
}

\newcommand{\figspatialmixing}{
\begin{figure*}[t]
\centering
\includegraphics[width=1.0\linewidth]{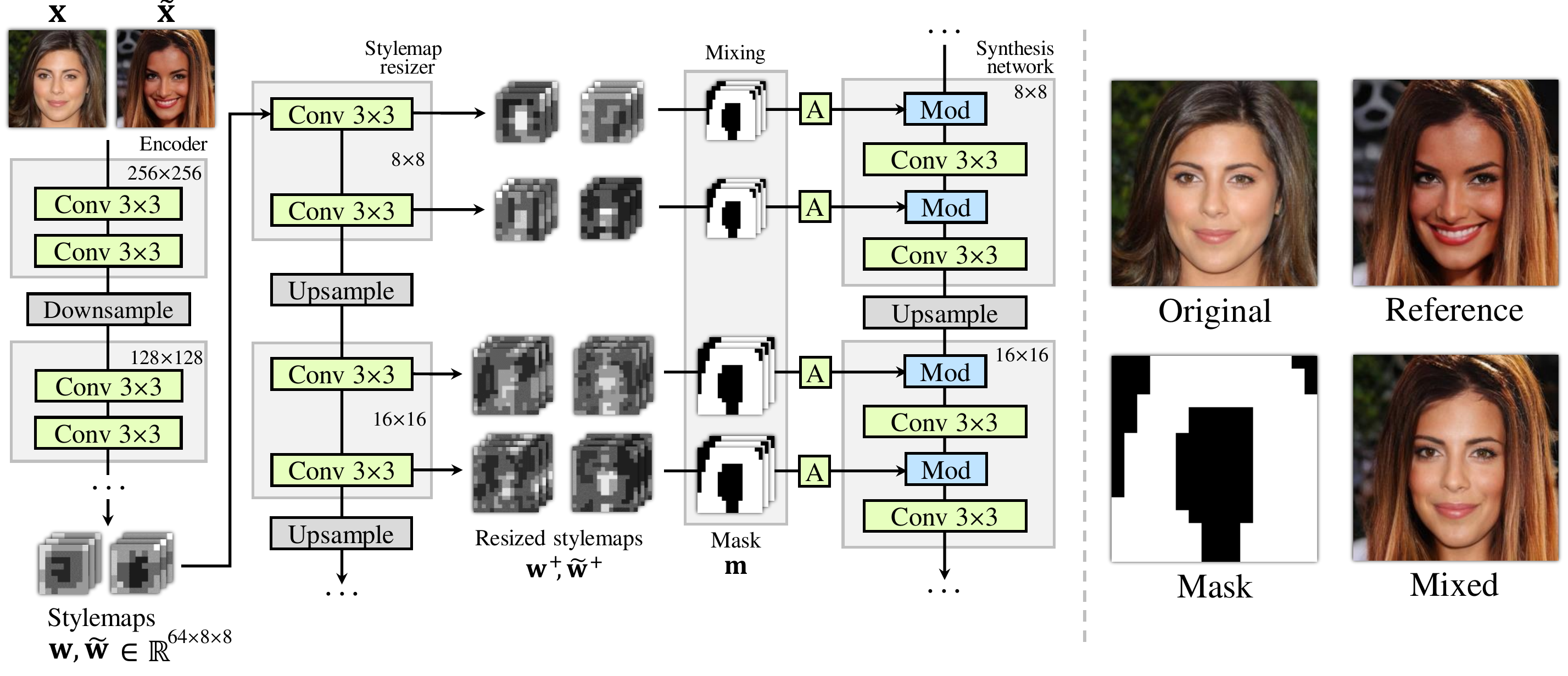}\\
\caption{Our local editing starts with a learned encoder for fast image-to-stylemap projection. We estimate the stylemaps $\w$ and $\wt$ of the original $\x$ and the reference $\xt$ and transform them into multiple resolutions through the learned stylemap resizer. For each resolution, we calculate the alpha blending of the two stylemaps using the user-defined binary mask $\m$. Finally, the learned generator produces the output using the spatially-mixed stylemaps. The right one shows an example generated using our method.} 
\label{fig:figspatialmixing}
\end{figure*}
}

\newcommand{\figRandomGeneration}{
\begin{figure*}[t]
\centering
\vspace{-3mm}
\makebox[0.45\linewidth][c]{\textbf{CelebA-HQ}, FID: \textbf{4.92}}
\makebox[0.45\linewidth][c]{\textbf{~~~AFHQ}, FID: \textbf{6.71}}\\
\hspace{3mm}\includegraphics[width=0.45\linewidth]{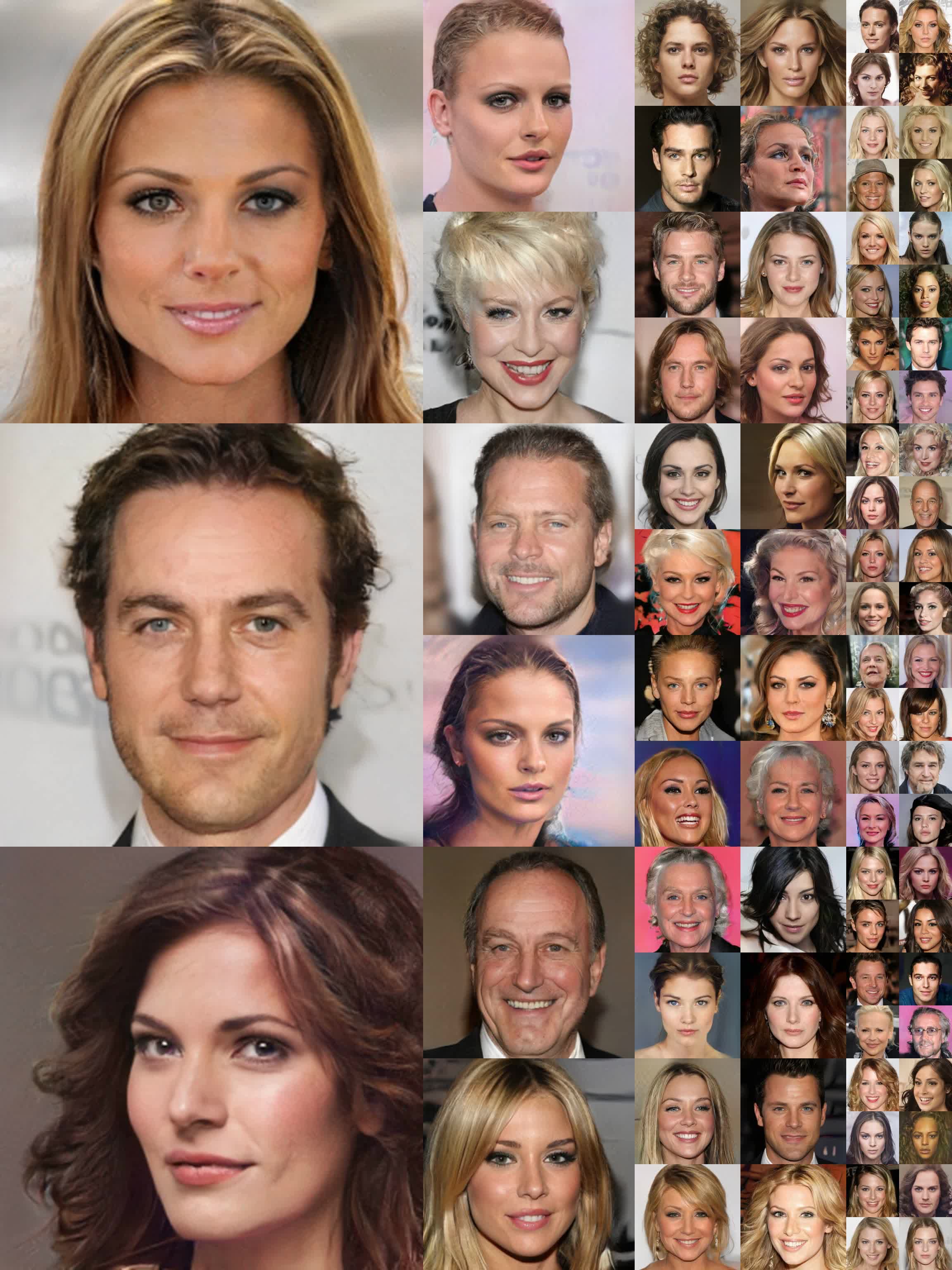}\hspace{3mm}
\includegraphics[width=0.45\linewidth]{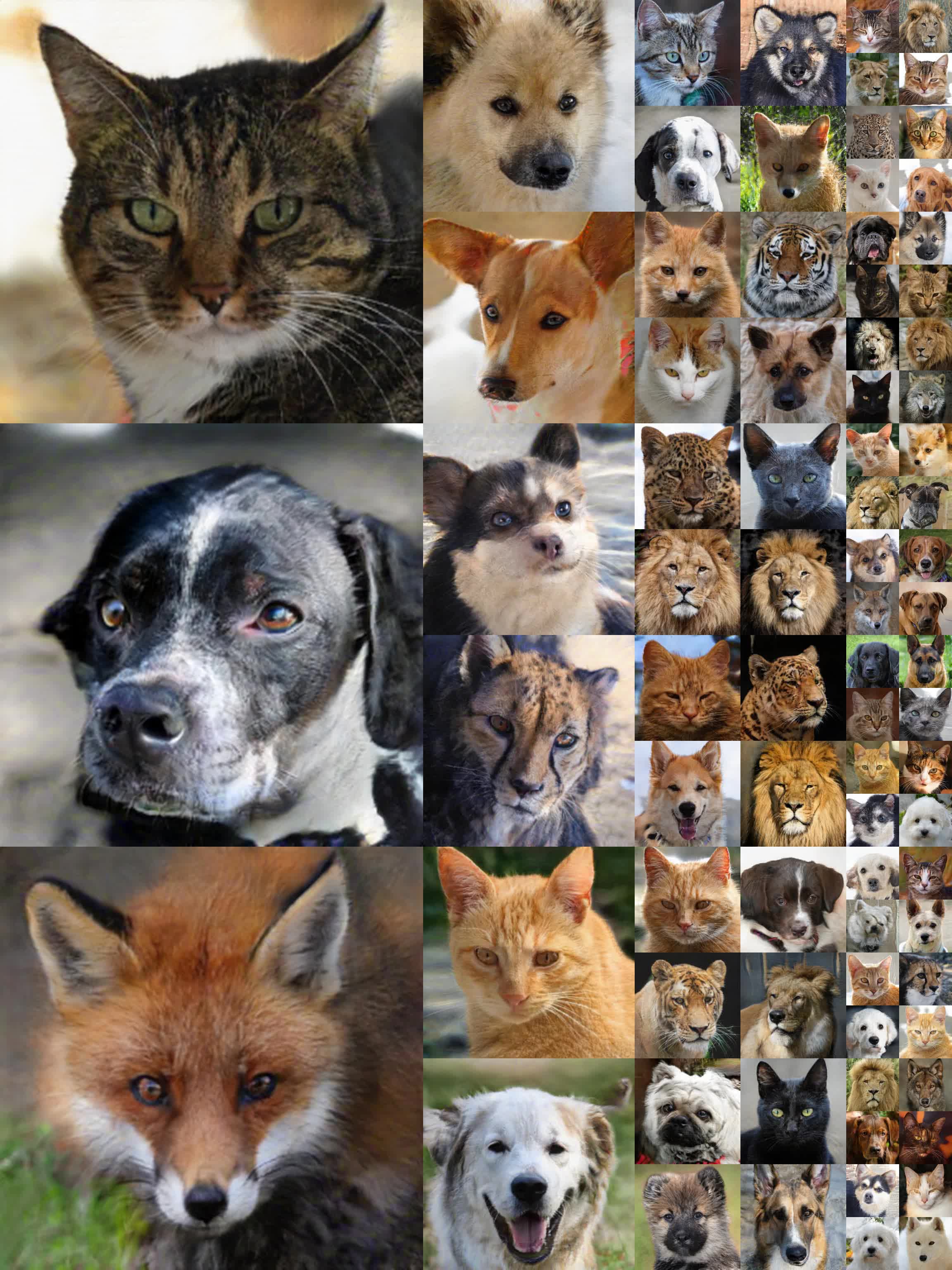}\\
\vspace{3mm}
\makebox[0.45\linewidth][c]{\textbf{LSUN Car}, FID: \textbf{4.15}}
\makebox[0.45\linewidth][c]{\textbf{LSUN Church}, FID: \textbf{2.95}}\vspace{1mm}\\
\hspace{3mm}\includegraphics[width=0.45\linewidth]{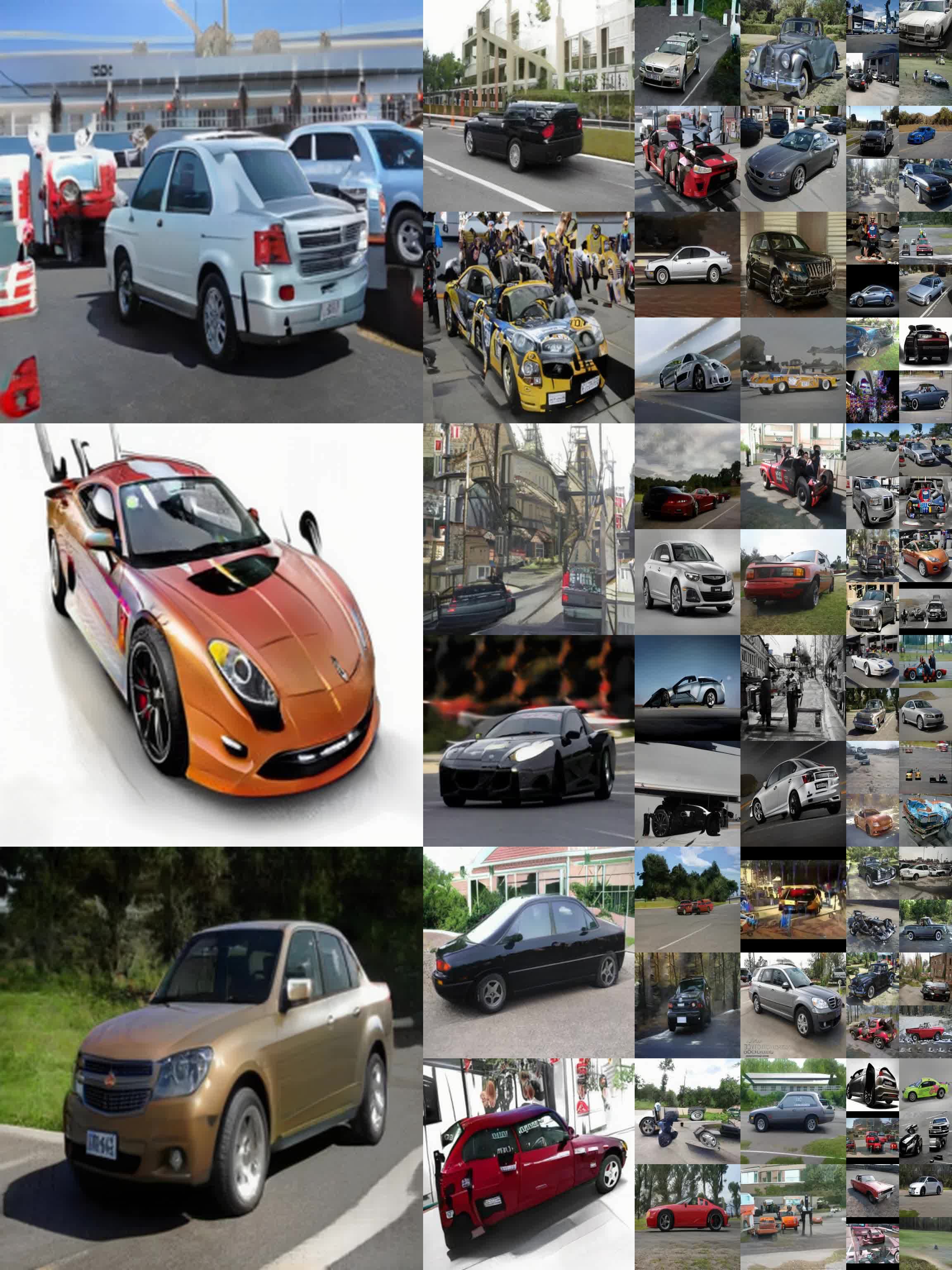}\hspace{3mm}
\includegraphics[width=0.45\linewidth]{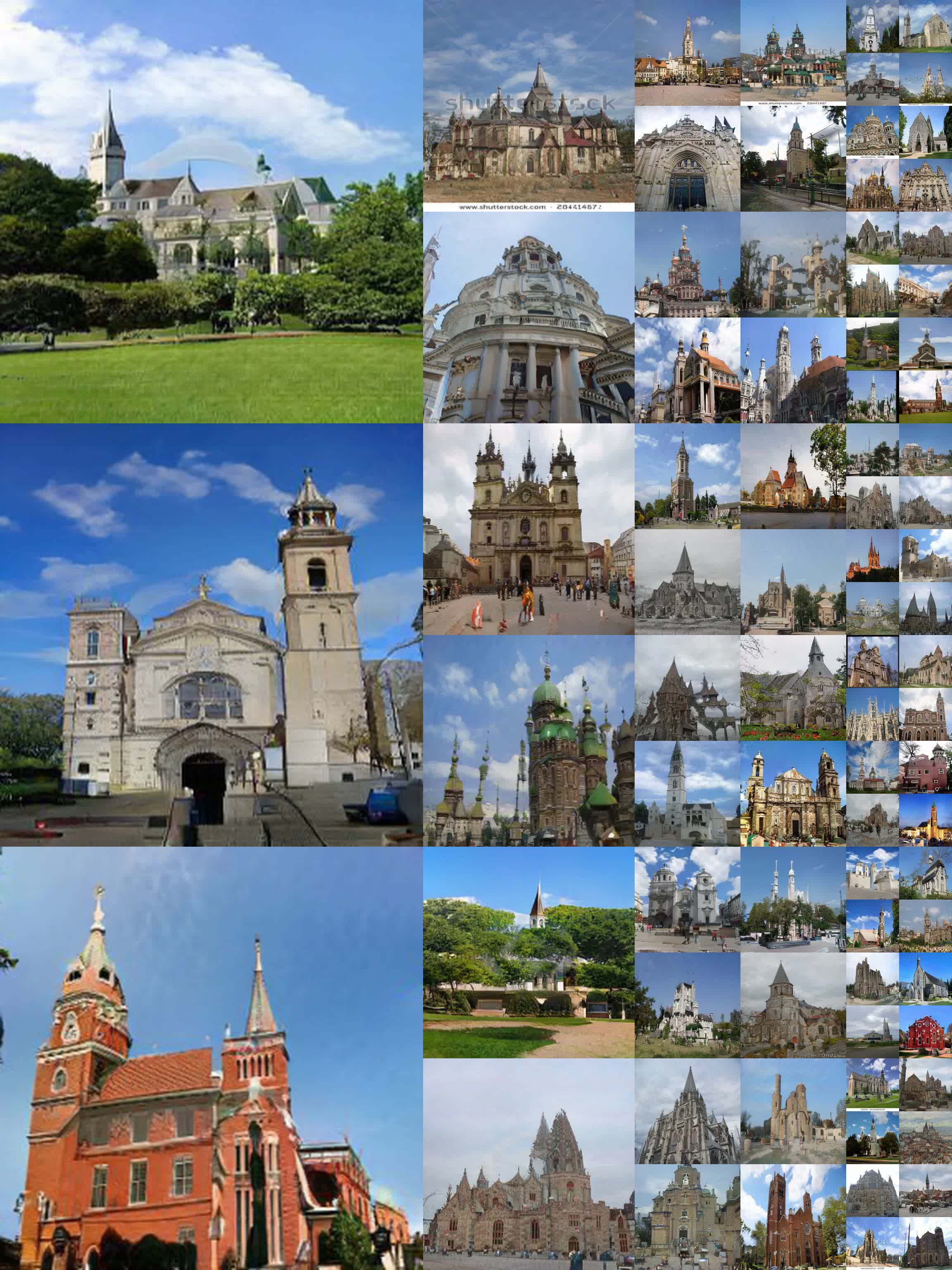}
\caption{Uncurated random generation results in four datasets.} 
\label{fig:figRandomGeneration}
\end{figure*}
}

\newcommand{\suppfigComparisonCeleb}{
\begin{figure*}[t]
    \begin{minipage}[]{\linewidth}
    
    \newcolumntype{h}{>{\centering\arraybackslash}p{23.8mm}}
    \setlength\tabcolsep{15pt} %
    \begin{tabular}{hhhhh}
    ~~~~Original & ~~~Reference & ~In-DomainGAN & SEAN & Ours %
    \end{tabular}
    
    \end{minipage}
    \centering
    \\
    \begin{subfigure}[h]{\htwo\linewidth}
        \includegraphics[width=\linewidth]{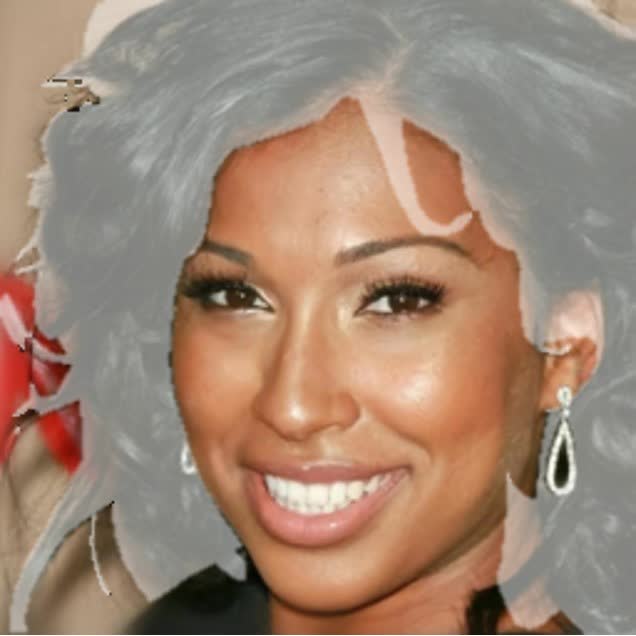}
    \end{subfigure}
    \begin{subfigure}[h]{\htwo\linewidth}
        \includegraphics[width=\linewidth]{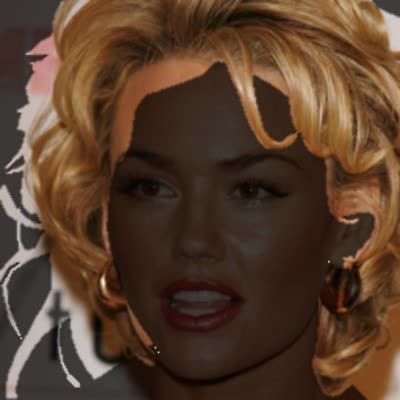}
    \end{subfigure}
    \begin{subfigure}[h]{\htwo\linewidth}
        \includegraphics[width=\linewidth]{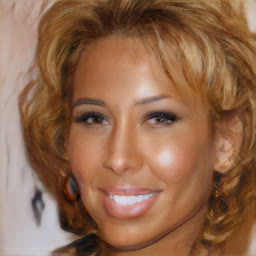}
    \end{subfigure}    
    \begin{subfigure}[h]{\htwo\linewidth}
        \includegraphics[width=\linewidth]{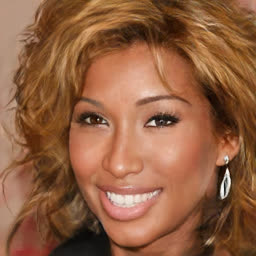}
    \end{subfigure}
    \begin{subfigure}[h]{\htwo\linewidth}
        \centering
        \includegraphics[width=\linewidth]{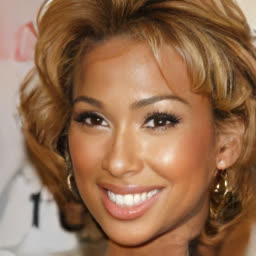}
    \end{subfigure}\\
    
        \begin{subfigure}[h]{\htwo\linewidth}
        \includegraphics[width=\linewidth]{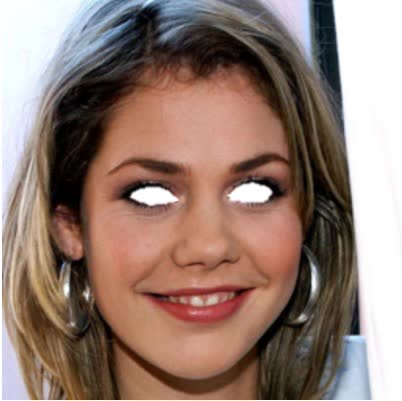}
    \end{subfigure}
    \begin{subfigure}[h]{\htwo\linewidth}
        \includegraphics[width=\linewidth]{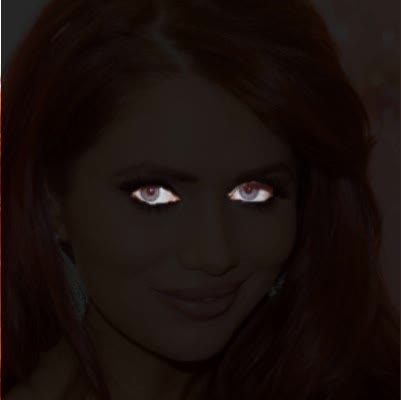}
    \end{subfigure}
    \begin{subfigure}[h]{\htwo\linewidth}
        \includegraphics[width=\linewidth]{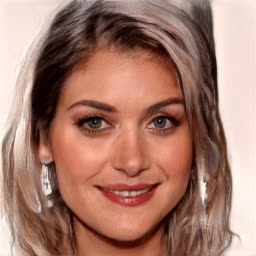}
    \end{subfigure}    
    \begin{subfigure}[h]{\htwo\linewidth}
        \includegraphics[width=\linewidth]{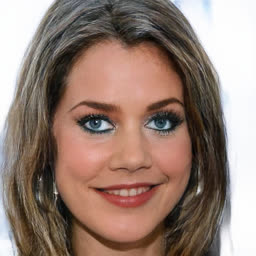}
    \end{subfigure}
    \begin{subfigure}[h]{\htwo\linewidth}
        \centering
        \includegraphics[width=\linewidth]{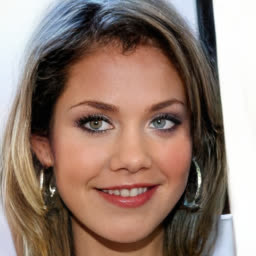}
    \end{subfigure}\\
    
        \begin{subfigure}[h]{\htwo\linewidth}
        \includegraphics[width=\linewidth]{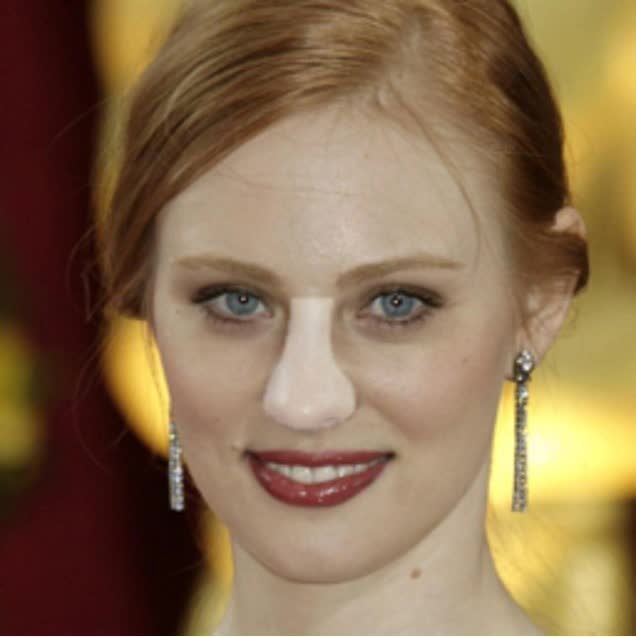}
    \end{subfigure}
    \begin{subfigure}[h]{\htwo\linewidth}
        \includegraphics[width=\linewidth]{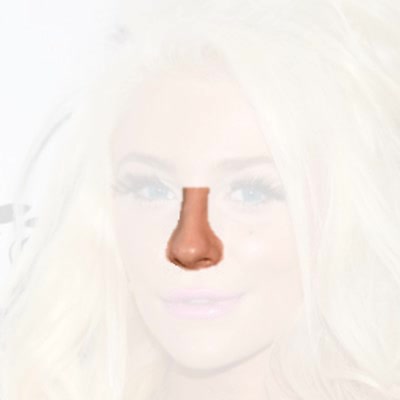}
    \end{subfigure}
    \begin{subfigure}[h]{\htwo\linewidth}
        \includegraphics[width=\linewidth]{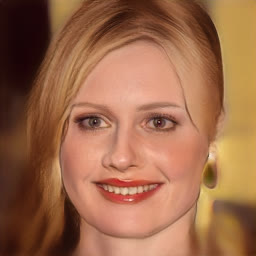}
    \end{subfigure}    
    \begin{subfigure}[h]{\htwo\linewidth}
        \includegraphics[width=\linewidth]{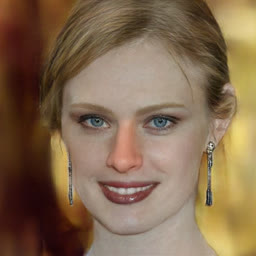}
    \end{subfigure}
    \begin{subfigure}[h]{\htwo\linewidth}
        \centering
        \includegraphics[width=\linewidth]{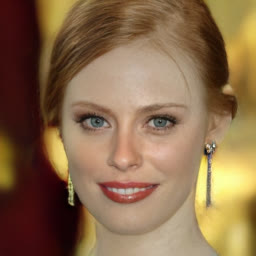}
    \end{subfigure}\\

        \begin{subfigure}[h]{\htwo\linewidth}
        \includegraphics[width=\linewidth]{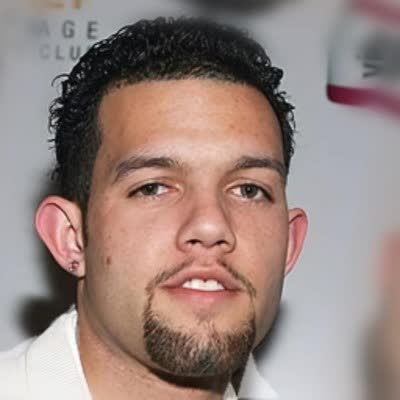}
    \end{subfigure}
    \begin{subfigure}[h]{\htwo\linewidth}
        \includegraphics[width=\linewidth]{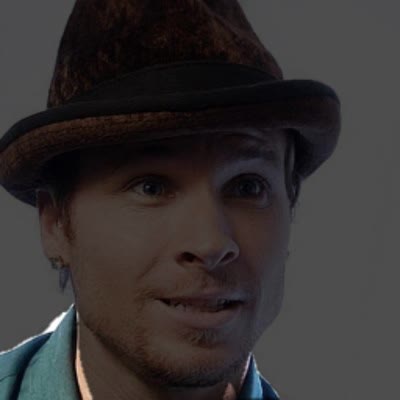}
    \end{subfigure}
    \begin{subfigure}[h]{\htwo\linewidth}
        \includegraphics[width=\linewidth]{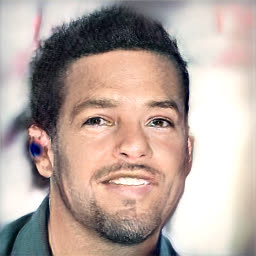}
    \end{subfigure}    
    \begin{subfigure}[h]{\htwo\linewidth}
        \includegraphics[width=\linewidth]{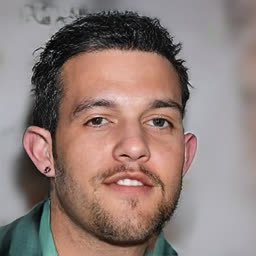}
    \end{subfigure}
    \begin{subfigure}[h]{\htwo\linewidth}
        \centering
        \includegraphics[width=\linewidth]{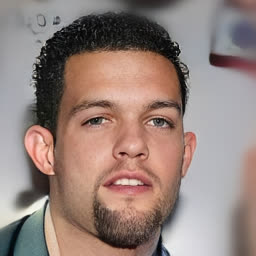}
    \end{subfigure}\\
    \begin{subfigure}[h]{\htwo\linewidth}
        \includegraphics[width=\linewidth]{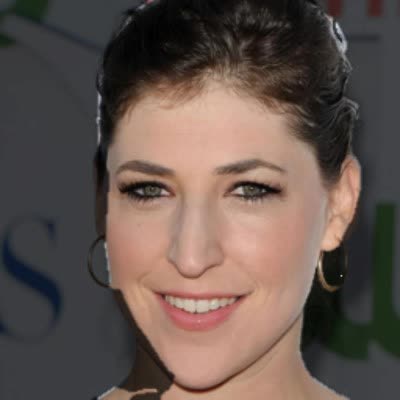}
    \end{subfigure}
    \begin{subfigure}[h]{\htwo\linewidth}
        \includegraphics[width=\linewidth]{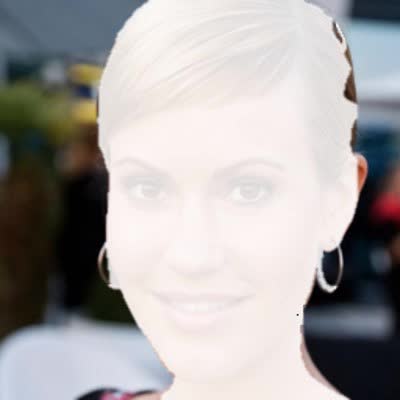}
    \end{subfigure}
    \begin{subfigure}[h]{\htwo\linewidth}
        \includegraphics[width=\linewidth]{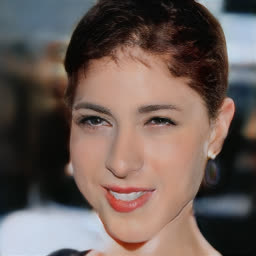}
    \end{subfigure}    
    \begin{subfigure}[h]{\htwo\linewidth}
        \includegraphics[width=\linewidth]{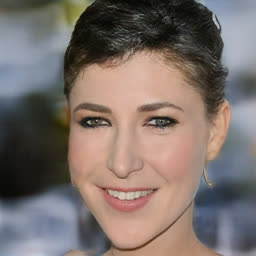}
    \end{subfigure}
    \begin{subfigure}[h]{\htwo\linewidth}
        \centering
        \includegraphics[width=\linewidth]{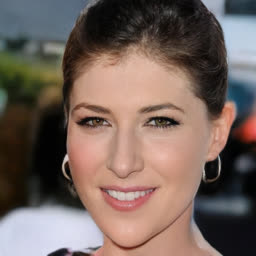}
    \end{subfigure}\\

\caption{Local editing comparison in CelebA-HQ. Each row edits hair, eyes, nose, cloth, and background. In-DomainGAN~\cite{zhu2020indomaingan} only optimizes region in the target mask and it changes identities of the original images. SEAN~\cite{zhu2020sean} tends to bring only the color and the texture of reference images, not the shape (especially on hair lines). Our method reflects the shape of the reference image as well and preserves the identity of the original image.
}
\label{fig:suppfigComparisonCeleb}
\end{figure*}
}

\newcommand{\suppfigInterpolation}{
\begin{figure*}[t]

    \begin{minipage}{\linewidth}
        \centering
        \hspace{3mm}
        \makebox[\himg\linewidth][c]{Input A}
        \makebox[\himg\linewidth][c]{Inversion A}
        \makebox[\himg\linewidth][c]{~~$\xleftarrow[]{~~~~~~~~~~~~~~~~~~~~~}$}\hfill
        \makebox[\himg\linewidth][c]{Interpolation}\hfill
        \makebox[\himg\linewidth][c]{$\xrightarrow[]{~~~~~~~~~~~~~~~~~~~~~}$~~}\hfill
        \makebox[\himg\linewidth][c]{Inversion B~}\hfill 
        \makebox[\himg\linewidth][c]{Input B~~~}\hfill \hspace{7mm} \\

        \rotatebox[origin=l]{90}{\makebox[0mm][l]{\hspace*{0.005\linewidth}\footnotesize{\raisebox{0.5mm}[0mm][0mm]{Image2StyleGAN}}}}\hspace{0.8mm}
        \includegraphics[width=\himg\linewidth]{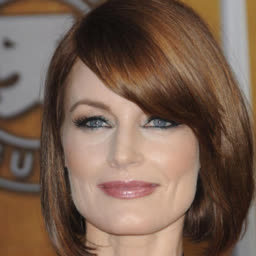}
        \includegraphics[width=\himg\linewidth]{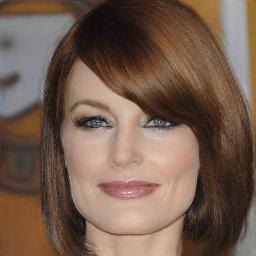}
        \includegraphics[width=\himg\linewidth]{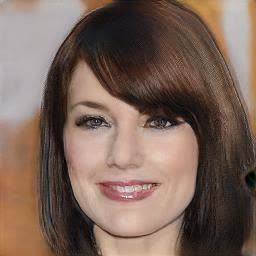}
        \includegraphics[width=\himg\linewidth]{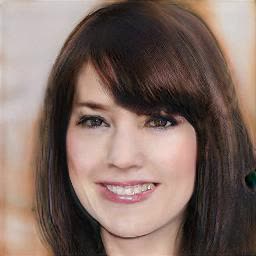}
        \includegraphics[width=\himg\linewidth]{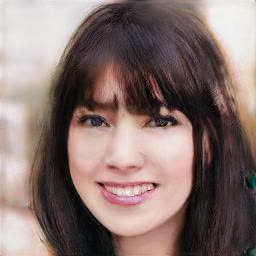}
        \includegraphics[width=\himg\linewidth]{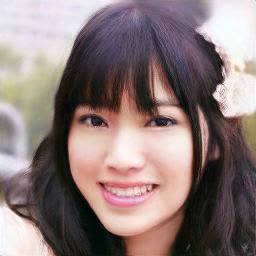}
        \includegraphics[width=\himg\linewidth]{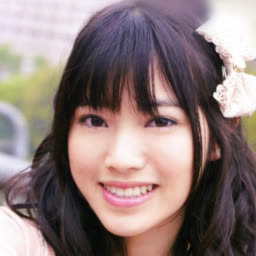} \\
        \vspace{-0.4mm}
        \rotatebox[origin=l]{90}{\makebox[0mm][l]{\hspace*{0.045\linewidth}\footnotesize{\raisebox{0.5mm}[0mm][0mm]{Ours}}}}\hspace{0.8mm}
        \includegraphics[width=\himg\linewidth]{supp_figures/w_interpolation/original/23.jpg}
        \includegraphics[width=\himg\linewidth]{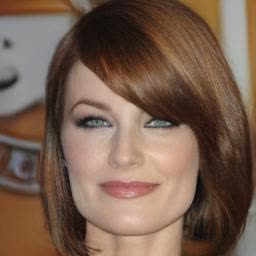}
        \includegraphics[width=\himg\linewidth]{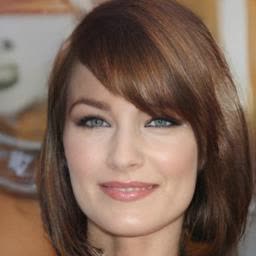}
        \includegraphics[width=\himg\linewidth]{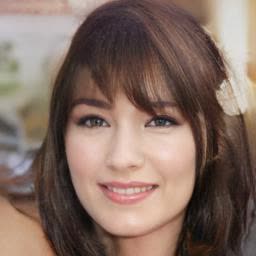}
        \includegraphics[width=\himg\linewidth]{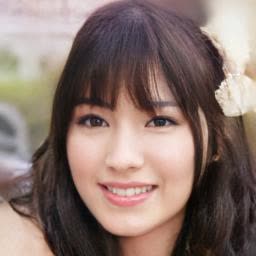}
        \includegraphics[width=\himg\linewidth]{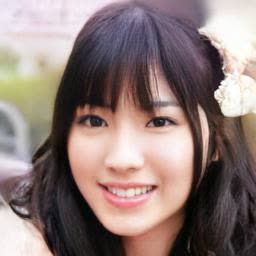}
        \includegraphics[width=\himg\linewidth]{supp_figures/w_interpolation/original/15.jpg} \\
        \vspace{1.0mm}
        \rotatebox[origin=l]{90}{\makebox[0mm][l]{\hspace*{0.005\linewidth}\footnotesize{\raisebox{0.5mm}[0mm][0mm]{Image2StyleGAN}}}}\hspace{0.8mm}
        \includegraphics[width=\himg\linewidth]{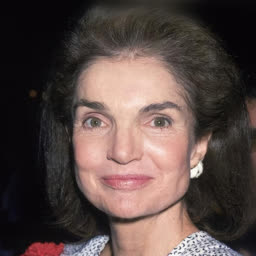}
        \includegraphics[width=\himg\linewidth]{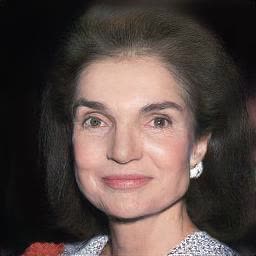}
        \includegraphics[width=\himg\linewidth]{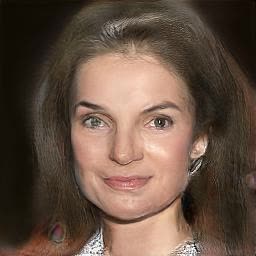}
        \includegraphics[width=\himg\linewidth]{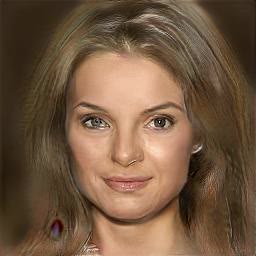}
        \includegraphics[width=\himg\linewidth]{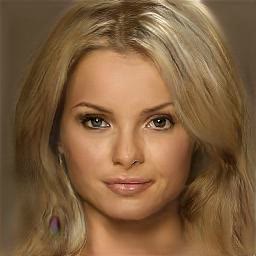}
        \includegraphics[width=\himg\linewidth]{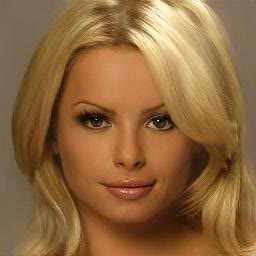}
        \includegraphics[width=\himg\linewidth]{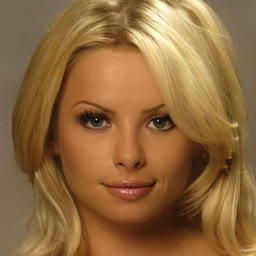} \\
        \vspace{-0.4mm}
        \rotatebox[origin=l]{90}{\makebox[0mm][l]{\hspace*{0.045\linewidth}\footnotesize{\raisebox{0.5mm}[0mm][0mm]{Ours}}}}\hspace{0.8mm}
        \includegraphics[width=\himg\linewidth]{supp_figures/w_interpolation/original/124.jpg}
        \includegraphics[width=\himg\linewidth]{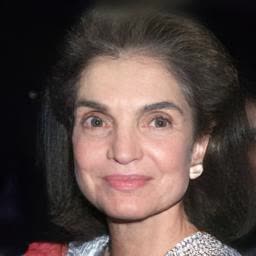}
        \includegraphics[width=\himg\linewidth]{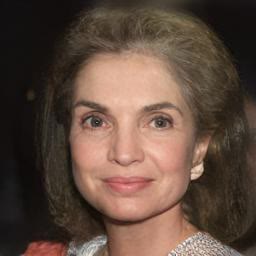}
        \includegraphics[width=\himg\linewidth]{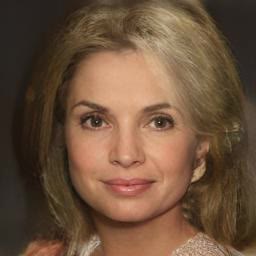}
        \includegraphics[width=\himg\linewidth]{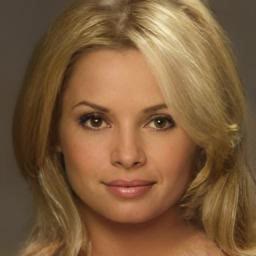}
        \includegraphics[width=\himg\linewidth]{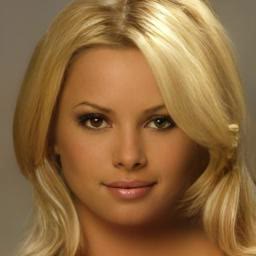}
        \includegraphics[width=\himg\linewidth]{supp_figures/w_interpolation/original/67.jpg} \\
        \vspace{1.0mm}
        \rotatebox[origin=l]{90}{\makebox[0mm][l]{\hspace*{0.005\linewidth}\footnotesize{\raisebox{0.5mm}[0mm][0mm]{Image2StyleGAN}}}}\hspace{0.8mm}
        \includegraphics[width=\himg\linewidth]{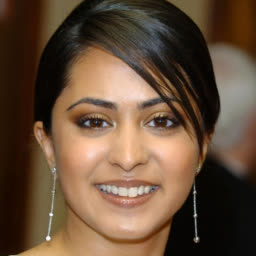}
        \includegraphics[width=\himg\linewidth]{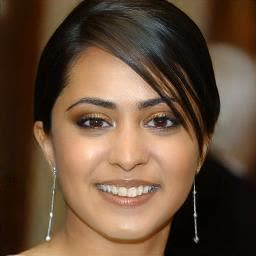}
        \includegraphics[width=\himg\linewidth]{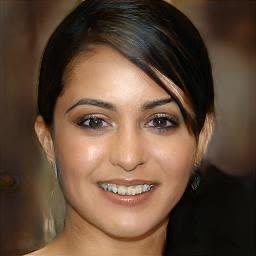}
        \includegraphics[width=\himg\linewidth]{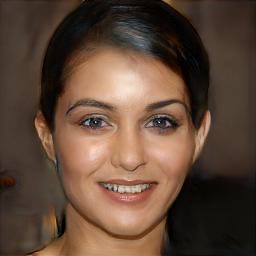}
        \includegraphics[width=\himg\linewidth]{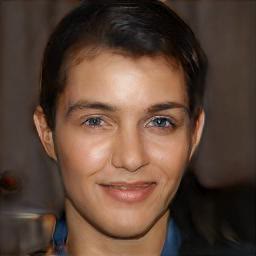}
        \includegraphics[width=\himg\linewidth]{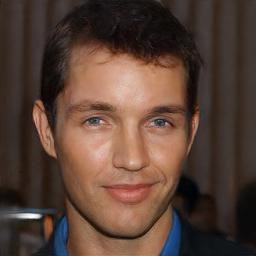}
        \includegraphics[width=\himg\linewidth]{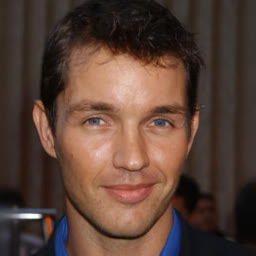} \\
        \vspace{-0.4mm}
        \rotatebox[origin=l]{90}{\makebox[0mm][l]{\hspace*{0.045\linewidth}\footnotesize{\raisebox{0.5mm}[0mm][0mm]{Ours}}}}\hspace{0.8mm}
        \includegraphics[width=\himg\linewidth]{supp_figures/w_interpolation/original/84.jpg}
        \includegraphics[width=\himg\linewidth]{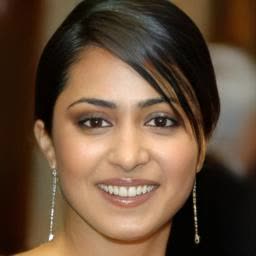}
        \includegraphics[width=\himg\linewidth]{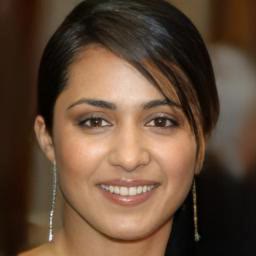}
        \includegraphics[width=\himg\linewidth]{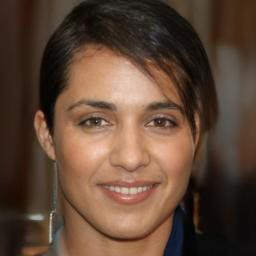}
        \includegraphics[width=\himg\linewidth]{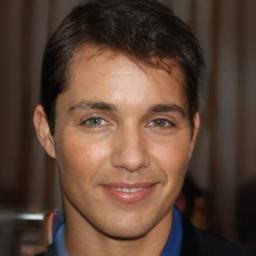}
        \includegraphics[width=\himg\linewidth]{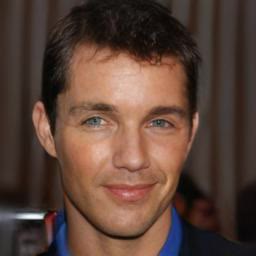}
        \includegraphics[width=\himg\linewidth]{supp_figures/w_interpolation/original/20.jpg} \\
        \vspace{1.0mm}
        \rotatebox[origin=l]{90}{\makebox[0mm][l]{\hspace*{0.005\linewidth}\footnotesize{\raisebox{0.5mm}[0mm][0mm]{Image2StyleGAN}}}}\hspace{0.8mm}
        \includegraphics[width=\himg\linewidth]{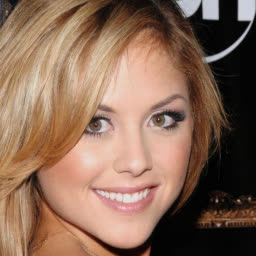}
        \includegraphics[width=\himg\linewidth]{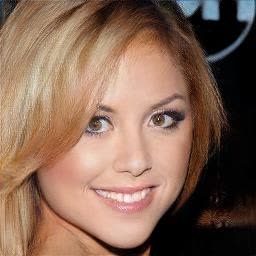}
        \includegraphics[width=\himg\linewidth]{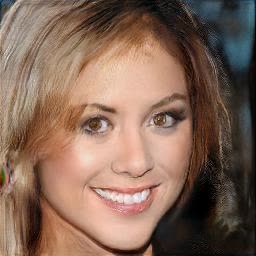}
        \includegraphics[width=\himg\linewidth]{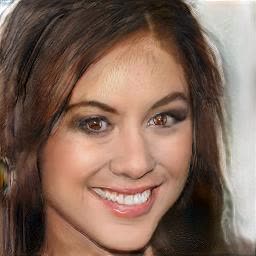}
        \includegraphics[width=\himg\linewidth]{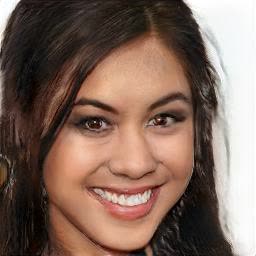}
        \includegraphics[width=\himg\linewidth]{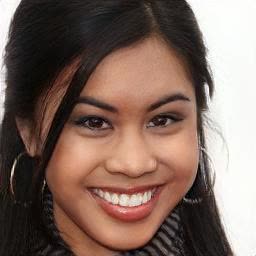}
        \includegraphics[width=\himg\linewidth]{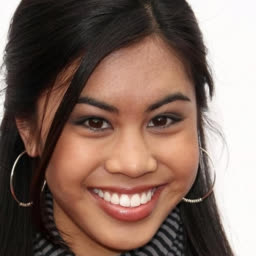} \\
        \vspace{-0.4mm}
        \rotatebox[origin=l]{90}{\makebox[0mm][l]{\hspace*{0.045\linewidth}\footnotesize{\raisebox{0.5mm}[0mm][0mm]{Ours}}}}\hspace{0.8mm}
        \includegraphics[width=\himg\linewidth]{supp_figures/w_interpolation/original/150.jpg}
        \includegraphics[width=\himg\linewidth]{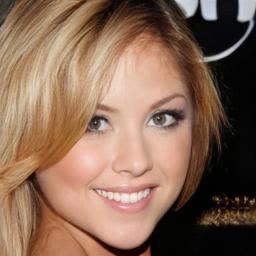}
        \includegraphics[width=\himg\linewidth]{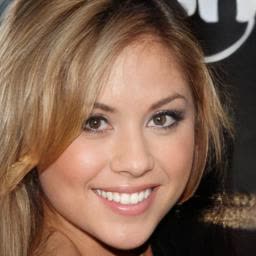}
        \includegraphics[width=\himg\linewidth]{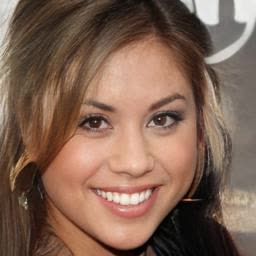}
        \includegraphics[width=\himg\linewidth]{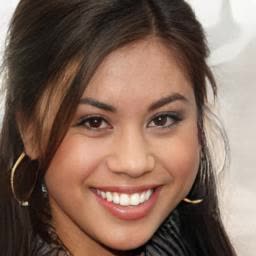}
        \includegraphics[width=\himg\linewidth]{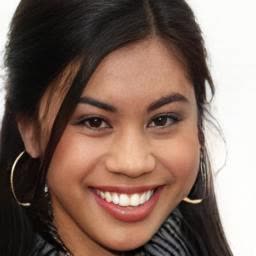}
        \includegraphics[width=\himg\linewidth]{supp_figures/w_interpolation/original/118.jpg} \\
        \vspace{1.0mm}

    \end{minipage}
\caption{Comparison with Image2StyleGAN for interpolation quality. Image2StyleGAN shows rugged and discontinuous interpolations, even though the reconstruction quality is visually good. Our method produces clearer and smoother interpolations, which reveal our superiority in both pixel-level and semantic-level (\ie, the semantics the original latent space) reconstruction. Note that the reconstruction speed of our method is $2000\times$ faster than Image2StyleGAN.
}
\label{fig:suppfigInterpolation}
\end{figure*}
}

\newcommand{\suppfigReconstruction}{
\renewcommand{\htwo}{0.190}
\begin{figure*}[t]
    \begin{minipage}[]{\linewidth}
    
    \newcolumntype{h}{>{\centering\arraybackslash}p{23.8mm}}
    \setlength\tabcolsep{15pt} %
    \begin{tabular}{hhhhh}
    ~~~~Original & ~~~ALAE & ~In-DomainGAN & ~~SEAN & Ours %
    \end{tabular}

    \end{minipage}
    \centering
    \\

    \begin{subfigure}[h]{\htwo\linewidth}
        \includegraphics[width=\linewidth]{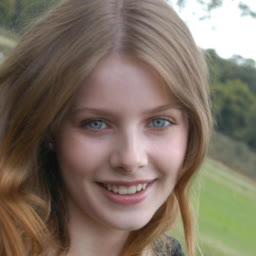}
    \end{subfigure}
    \begin{subfigure}[h]{\htwo\linewidth}
        \includegraphics[width=\linewidth]{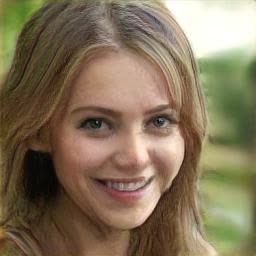}
    \end{subfigure}
    \begin{subfigure}[h]{\htwo\linewidth}
        \includegraphics[width=\linewidth]{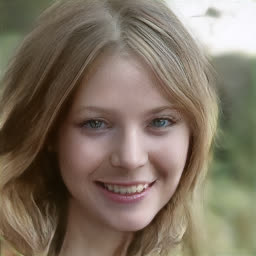}
    \end{subfigure}    
    \begin{subfigure}[h]{\htwo\linewidth}
        \includegraphics[width=\linewidth]{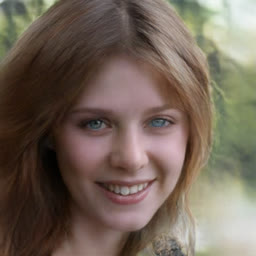}
    \end{subfigure}
    \begin{subfigure}[h]{\htwo\linewidth}
        \centering
        \includegraphics[width=\linewidth]{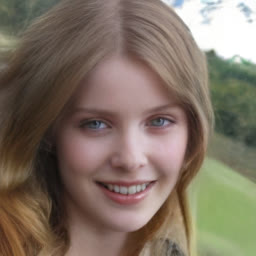}
    \end{subfigure}\\

    \begin{subfigure}[h]{\htwo\linewidth}
        \includegraphics[width=\linewidth]{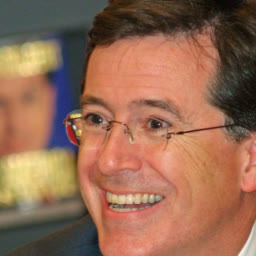}
    \end{subfigure}
    \begin{subfigure}[h]{\htwo\linewidth}
        \includegraphics[width=\linewidth]{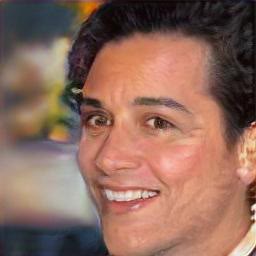}
    \end{subfigure}
    \begin{subfigure}[h]{\htwo\linewidth}
        \includegraphics[width=\linewidth]{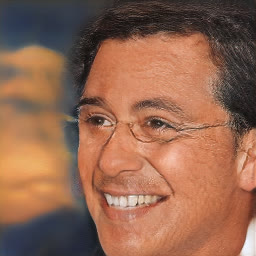}
    \end{subfigure}    
    \begin{subfigure}[h]{\htwo\linewidth}
        \includegraphics[width=\linewidth]{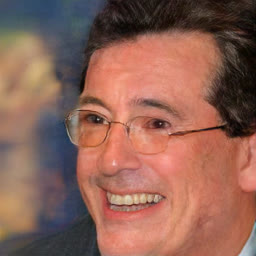}
    \end{subfigure}
    \begin{subfigure}[h]{\htwo\linewidth}
        \centering
        \includegraphics[width=\linewidth]{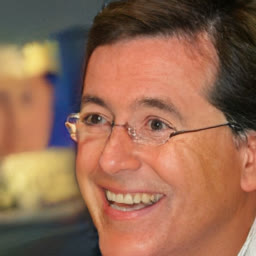}
    \end{subfigure}\\

    \begin{subfigure}[h]{\htwo\linewidth}
        \includegraphics[width=\linewidth]{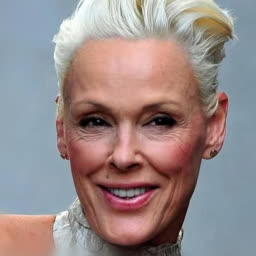}
    \end{subfigure}
    \begin{subfigure}[h]{\htwo\linewidth}
        \includegraphics[width=\linewidth]{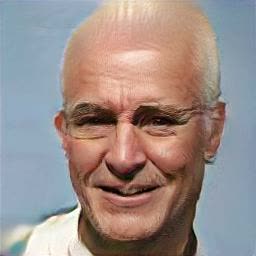}
    \end{subfigure}
    \begin{subfigure}[h]{\htwo\linewidth}
        \includegraphics[width=\linewidth]{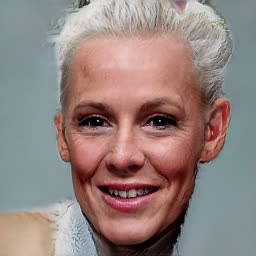}
    \end{subfigure}    
    \begin{subfigure}[h]{\htwo\linewidth}
        \includegraphics[width=\linewidth]{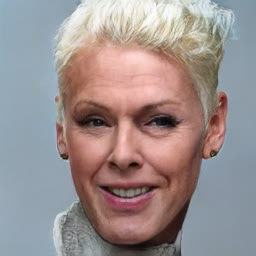}
    \end{subfigure}
    \begin{subfigure}[h]{\htwo\linewidth}
        \centering
        \includegraphics[width=\linewidth]{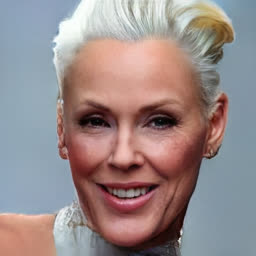}
    \end{subfigure}\\

    \begin{subfigure}[h]{\htwo\linewidth}
        \includegraphics[width=\linewidth]{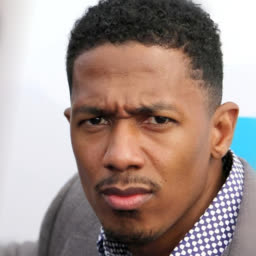}
    \end{subfigure}
    \begin{subfigure}[h]{\htwo\linewidth}
        \includegraphics[width=\linewidth]{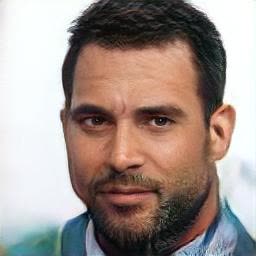}
    \end{subfigure}
    \begin{subfigure}[h]{\htwo\linewidth}
        \includegraphics[width=\linewidth]{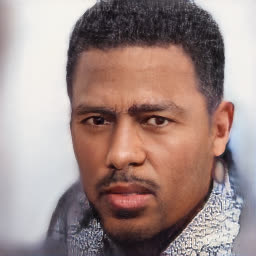}
    \end{subfigure}    
    \begin{subfigure}[h]{\htwo\linewidth}
        \includegraphics[width=\linewidth]{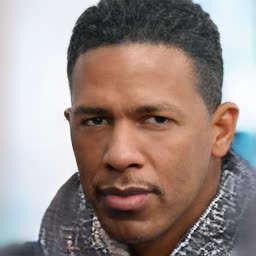}
    \end{subfigure}
    \begin{subfigure}[h]{\htwo\linewidth}
        \centering
        \includegraphics[width=\linewidth]{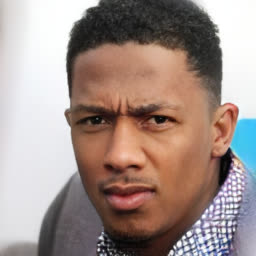}
    \end{subfigure}\\

    \begin{subfigure}[h]{\htwo\linewidth}
        \includegraphics[width=\linewidth]{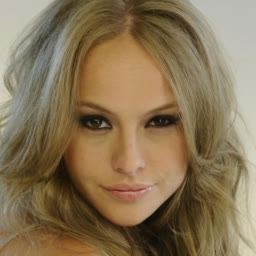}
    \end{subfigure}
    \begin{subfigure}[h]{\htwo\linewidth}
        \includegraphics[width=\linewidth]{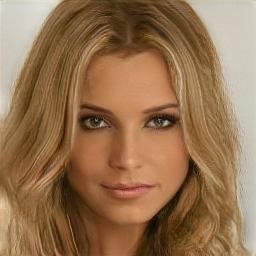}
    \end{subfigure}
    \begin{subfigure}[h]{\htwo\linewidth}
        \includegraphics[width=\linewidth]{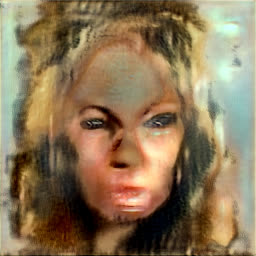}
    \end{subfigure}    
    \begin{subfigure}[h]{\htwo\linewidth}
        \includegraphics[width=\linewidth]{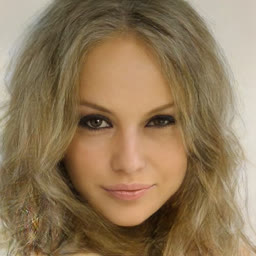}
    \end{subfigure}
    \begin{subfigure}[h]{\htwo\linewidth}
        \centering
        \includegraphics[width=\linewidth]{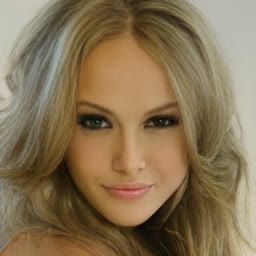}
    \end{subfigure}\\

    \begin{subfigure}[h]{\htwo\linewidth}
        \includegraphics[width=\linewidth]{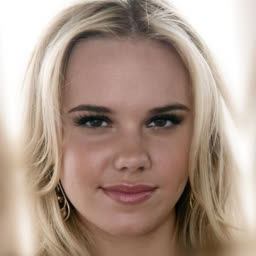}
    \end{subfigure}
    \begin{subfigure}[h]{\htwo\linewidth}
        \includegraphics[width=\linewidth]{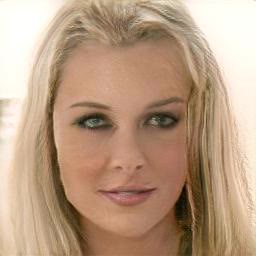}
    \end{subfigure}
    \begin{subfigure}[h]{\htwo\linewidth}
        \includegraphics[width=\linewidth]{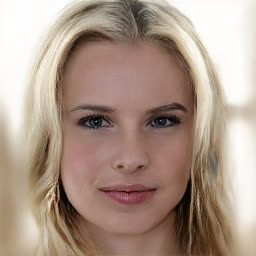}
    \end{subfigure}    
    \begin{subfigure}[h]{\htwo\linewidth}
        \includegraphics[width=\linewidth]{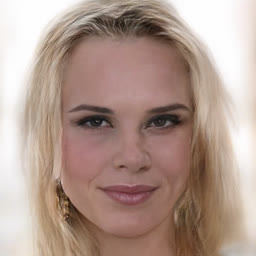}
    \end{subfigure}
    \begin{subfigure}[h]{\htwo\linewidth}
        \centering
        \includegraphics[width=\linewidth]{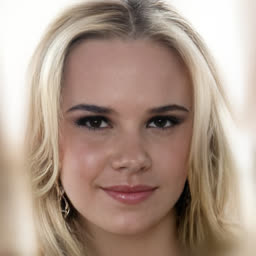}
    \end{subfigure}\\

\caption{Reconstruction results in encoder-based methods. ALAE~\cite{pidhorskyi2020alae} does not preserve identities in the original images. In-DomainGAN~\cite{zhu2020indomaingan} has better reconstruction results than ALAE, but it sometimes fails to generate human-like images as shown in the second-last row. Note that In-DomainGAN requires additional optimization steps which take seconds. SEAN~\cite{zhu2020sean} fails to preserve the shape of original images (\eg, background, hair curl, and cloth). Our method reconstructs images well not only the color and texture but also the shape.}
\label{fig:suppfigReconstruction}
\end{figure*}
}

\newcommand{\suppfigUnalignedCar}{
    
\begin{figure*}[t]
\centering
\begin{minipage}[t]{\linewidth}
\makebox[\hh][c]{}\hspace{0mm}%
\makebox[\h][c]{\textbf{\footnotesize{\ Reference}}}\hspace{3mm}
\rotatebox[origin=l]{90}{\makebox[0mm][l]{\hspace*{0.045\linewidth}\textbf{\footnotesize{\raisebox{0.5mm}[0mm][0mm]{Original}}}}}\hspace{0.8mm}%
\includegraphics[height=\h]{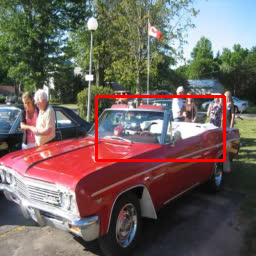}
\includegraphics[height=\h]{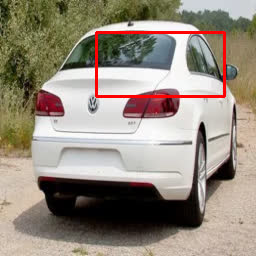}
\includegraphics[height=\h]{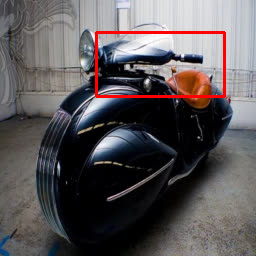}
\includegraphics[height=\h]{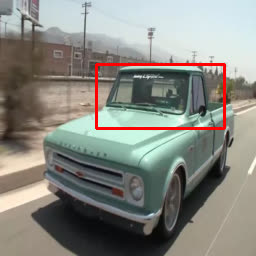}
\includegraphics[height=\h]{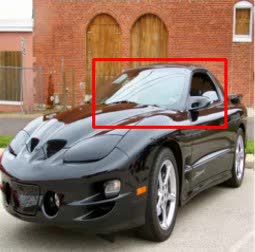}
\vspace{1.mm}\\
\makebox[\hh][c]{}\hspace{0.3mm}%
\includegraphics[height=\h]{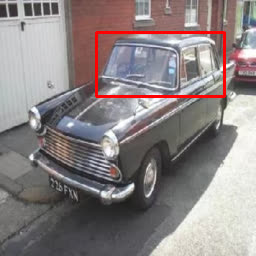}\hspace{3.5mm}
\includegraphics[height=\h]{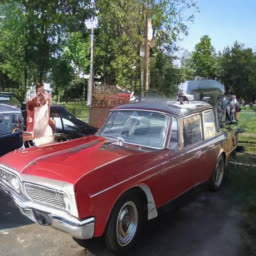}
\includegraphics[height=\h]{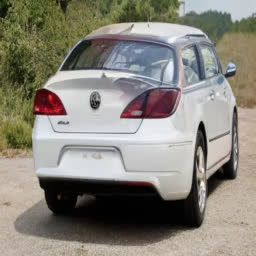}
\includegraphics[height=\h]{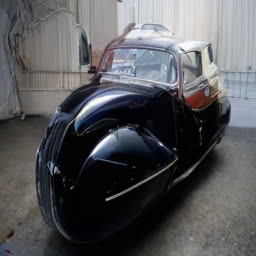}
\includegraphics[height=\h]{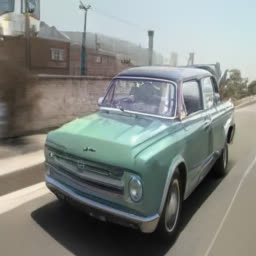}
\includegraphics[height=\h]{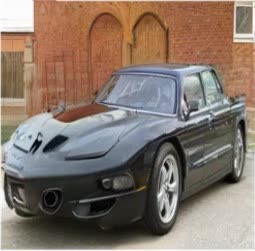}
\vspace{3.5mm}\\
\makebox[\hh][c]{}\hspace{0mm}%
\makebox[\h][c]{\textbf{\footnotesize{\ Reference}}}\hspace{3mm}
\rotatebox[origin=l]{90}{\makebox[0mm][l]{\hspace*{0.045\linewidth}\textbf{\footnotesize{\raisebox{.5mm}[0mm][0mm]{Original}}}}}\hspace{0.8mm}%
\includegraphics[height=\h]{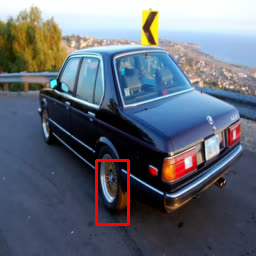}
\includegraphics[height=\h]{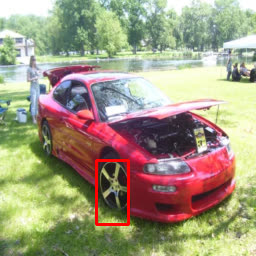}
\includegraphics[height=\h]{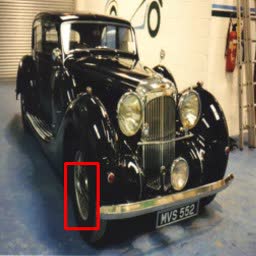}
\includegraphics[height=\h]{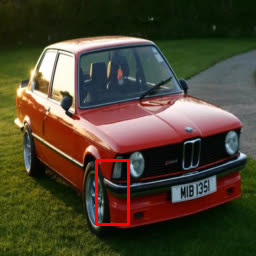}
\includegraphics[height=\h]{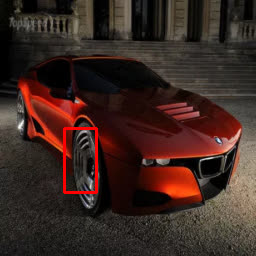}
\vspace{1.mm}\\
\makebox[\hh][c]{}\hspace{0.5mm}%
\includegraphics[height=\h]{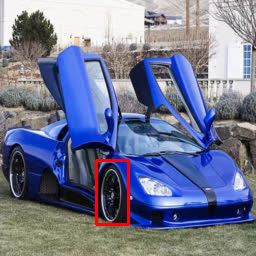}\hspace{3.3mm}
\includegraphics[height=\h]{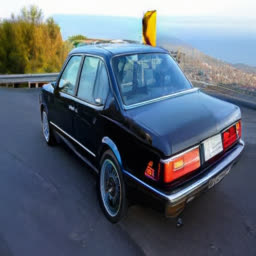}
\includegraphics[height=\h]{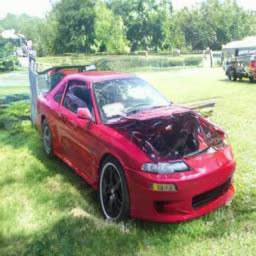}
\includegraphics[height=\h]{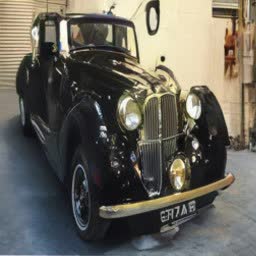}
\includegraphics[height=\h]{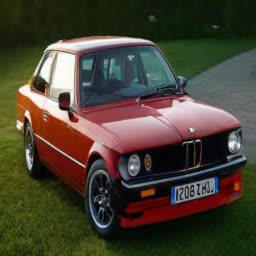}
\includegraphics[height=\h]{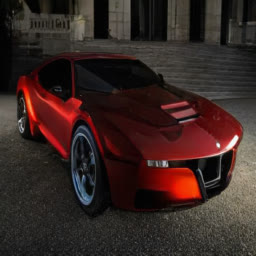}\vspace{3.5mm}\\
\makebox[\hh][c]{}\hspace{0mm}%
\makebox[\h][c]{\textbf{\footnotesize{\ Reference}}}\hspace{3mm}
\rotatebox[origin=l]{90}{\makebox[0mm][l]{\hspace*{0.045\linewidth}\textbf{\footnotesize{\raisebox{.5mm}[0mm][0mm]{Original}}}}}\hspace{0.8mm}%
\includegraphics[height=\h]{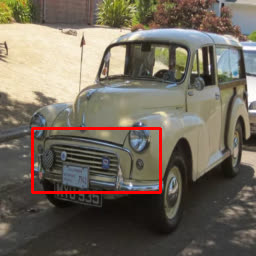}
\includegraphics[height=\h]{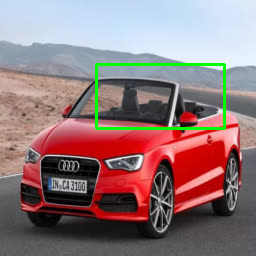}
\includegraphics[height=\h]{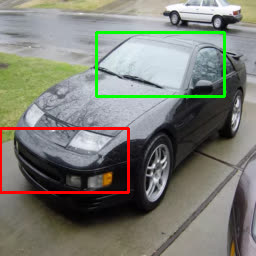}
\includegraphics[height=\h]{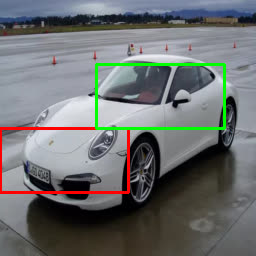}
\includegraphics[height=\h]{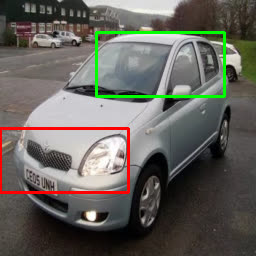}
\vspace{1.mm}\\
\makebox[\hh][c]{}\hspace{0.5mm}%
\includegraphics[height=\h]{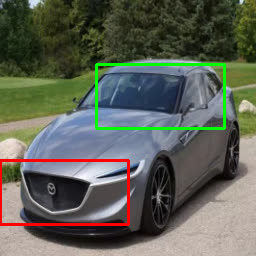}\hspace{3.3mm}
\includegraphics[height=\h]{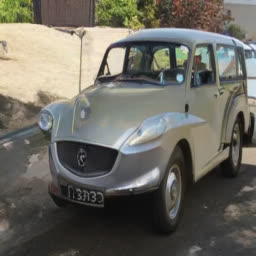}
\includegraphics[height=\h]{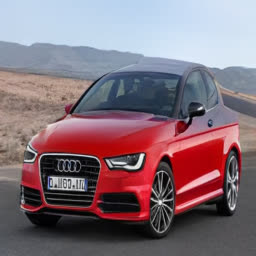}
\includegraphics[height=\h]{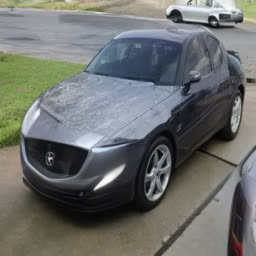}
\includegraphics[height=\h]{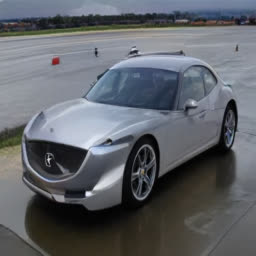}
\includegraphics[height=\h]{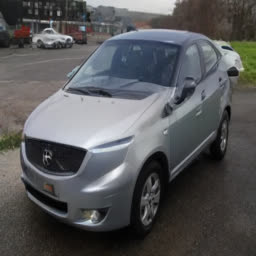}\\
\end{minipage}

\caption{Transplantation results of our method in LSUN Car. We transplant the cabin, wheel, and bumper.
}
\label{fig:suppfigUnalignedCar}
\end{figure*}

}

\newcommand{\suppfigUnalignedChurch}{
\begin{figure*}[t]
\centering
\begin{minipage}[t]{\linewidth}
\makebox[\hh][c]{}\hspace{0mm}%
\makebox[\h][c]{\textbf{\footnotesize{\ Reference}}}\hspace{3mm}
\rotatebox[origin=l]{90}{\makebox[0mm][l]{\hspace*{0.045\linewidth}\textbf{\footnotesize{\raisebox{0.5mm}[0mm][0mm]{Original}}}}}\hspace{0.8mm}%
\includegraphics[height=\h]{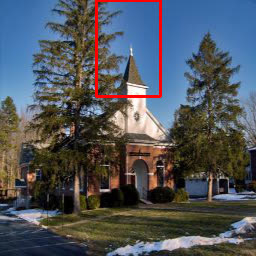}
\includegraphics[height=\h]{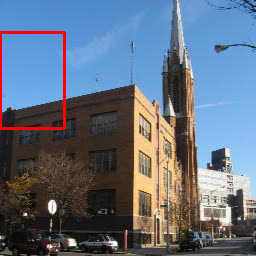}
\includegraphics[height=\h]{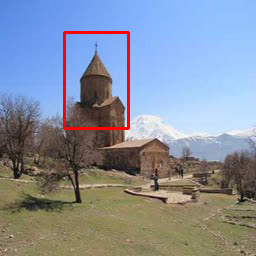}
\includegraphics[height=\h]{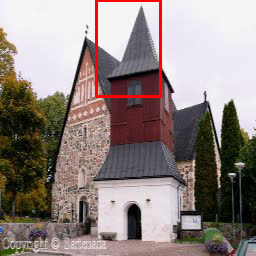}
\includegraphics[height=\h]{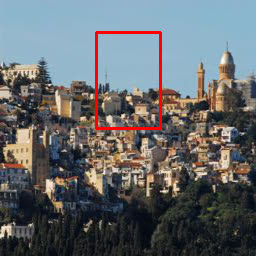}
\vspace{1.mm}\\
\makebox[\hh][c]{}\hspace{0.3mm}%
\includegraphics[height=\h]{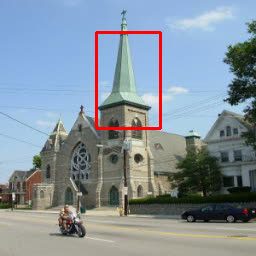}\hspace{3.5mm}
\includegraphics[height=\h]{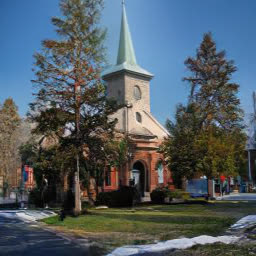}
\includegraphics[height=\h]{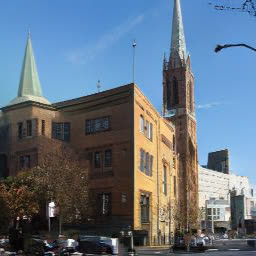}
\includegraphics[height=\h]{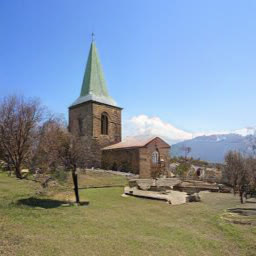}
\includegraphics[height=\h]{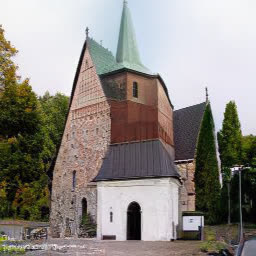}
\includegraphics[height=\h]{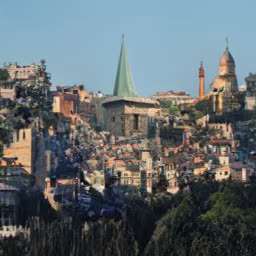}
\vspace{3.5mm}\\
\makebox[\hh][c]{}\hspace{0mm}%
\makebox[\h][c]{\textbf{\footnotesize{\ Reference}}}\hspace{3mm}
\rotatebox[origin=l]{90}{\makebox[0mm][l]{\hspace*{0.045\linewidth}\textbf{\footnotesize{\raisebox{.5mm}[0mm][0mm]{Original}}}}}\hspace{0.8mm}%
\includegraphics[height=\h]{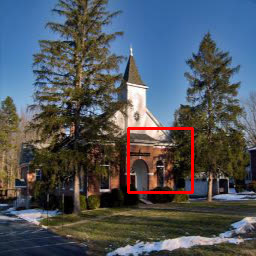}
\includegraphics[height=\h]{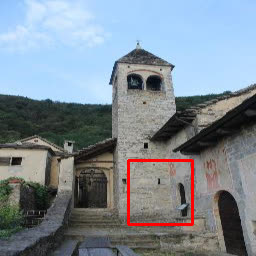}
\includegraphics[height=\h]{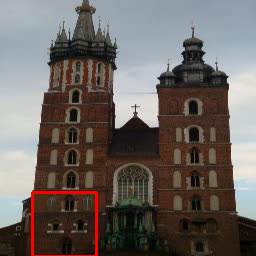}
\includegraphics[height=\h]{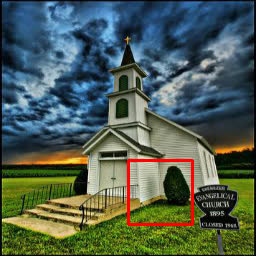}
\includegraphics[height=\h]{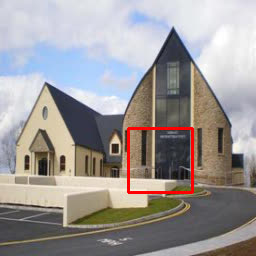}
\vspace{1.mm}\\
\makebox[\hh][c]{}\hspace{0.5mm}%
\includegraphics[height=\h]{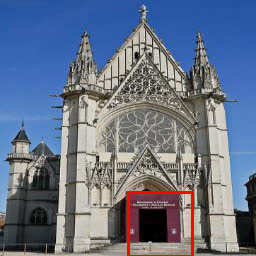}\hspace{3.3mm}
\includegraphics[height=\h]{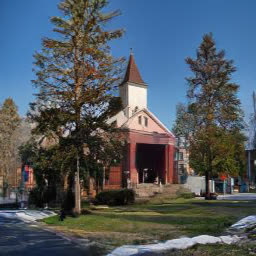}
\includegraphics[height=\h]{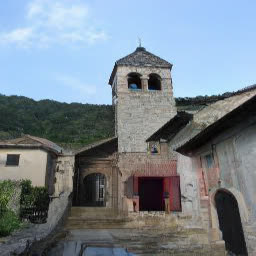}
\includegraphics[height=\h]{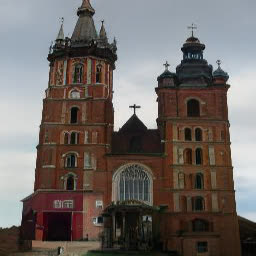}
\includegraphics[height=\h]{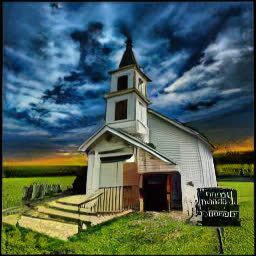}
\includegraphics[height=\h]{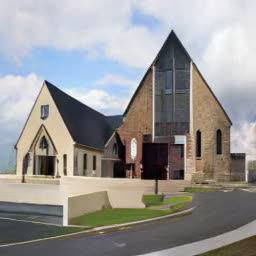}\vspace{3.5mm}\\
\makebox[\hh][c]{}\hspace{0mm}%
\makebox[\h][c]{\textbf{\footnotesize{\ Reference}}}\hspace{3mm}
\rotatebox[origin=l]{90}{\makebox[0mm][l]{\hspace*{0.045\linewidth}\textbf{\footnotesize{\raisebox{.5mm}[0mm][0mm]{Original}}}}}\hspace{0.8mm}%
\includegraphics[height=\h]{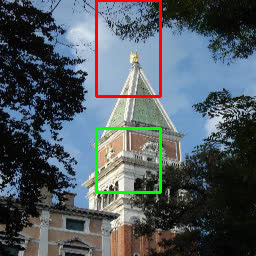}
\includegraphics[height=\h]{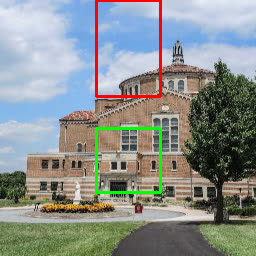}
\includegraphics[height=\h]{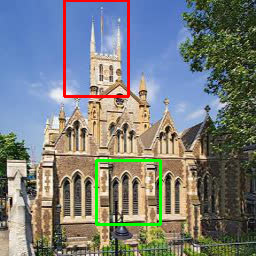}
\includegraphics[height=\h]{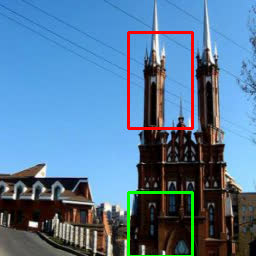}
\includegraphics[height=\h]{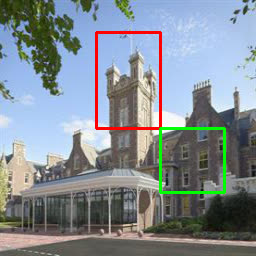}
\vspace{1.mm}\\
\makebox[\hh][c]{}\hspace{0.5mm}%
\includegraphics[height=\h]{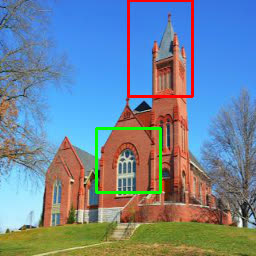}\hspace{3.3mm}
\includegraphics[height=\h]{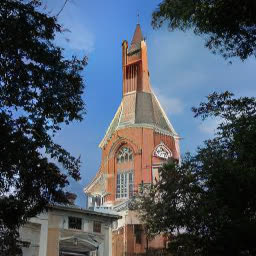}
\includegraphics[height=\h]{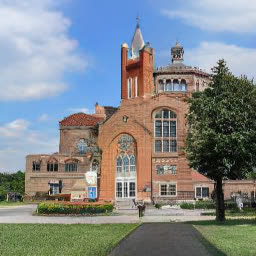}
\includegraphics[height=\h]{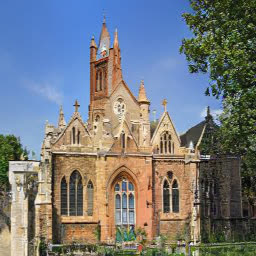}
\includegraphics[height=\h]{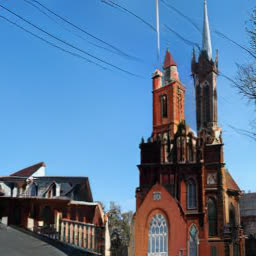}
\includegraphics[height=\h]{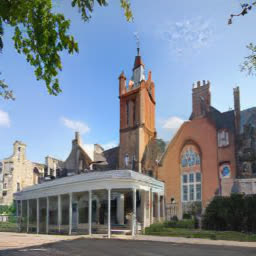}\\
\end{minipage}

\caption{Transplanting tower, gate and windows in LSUN Church by our method.
Our method can transplant the arbitrary number and location of areas in reference images to the original images. Note that our method adjusts color tone and structure of the same references regarding the original images.
}
\label{fig:suppfigUnalignedChurch}
\end{figure*}
}

\newcommand{\suppfigStyleMixing}{
\begin{figure*}[t]
\centering
\includegraphics[width=1.0\linewidth]{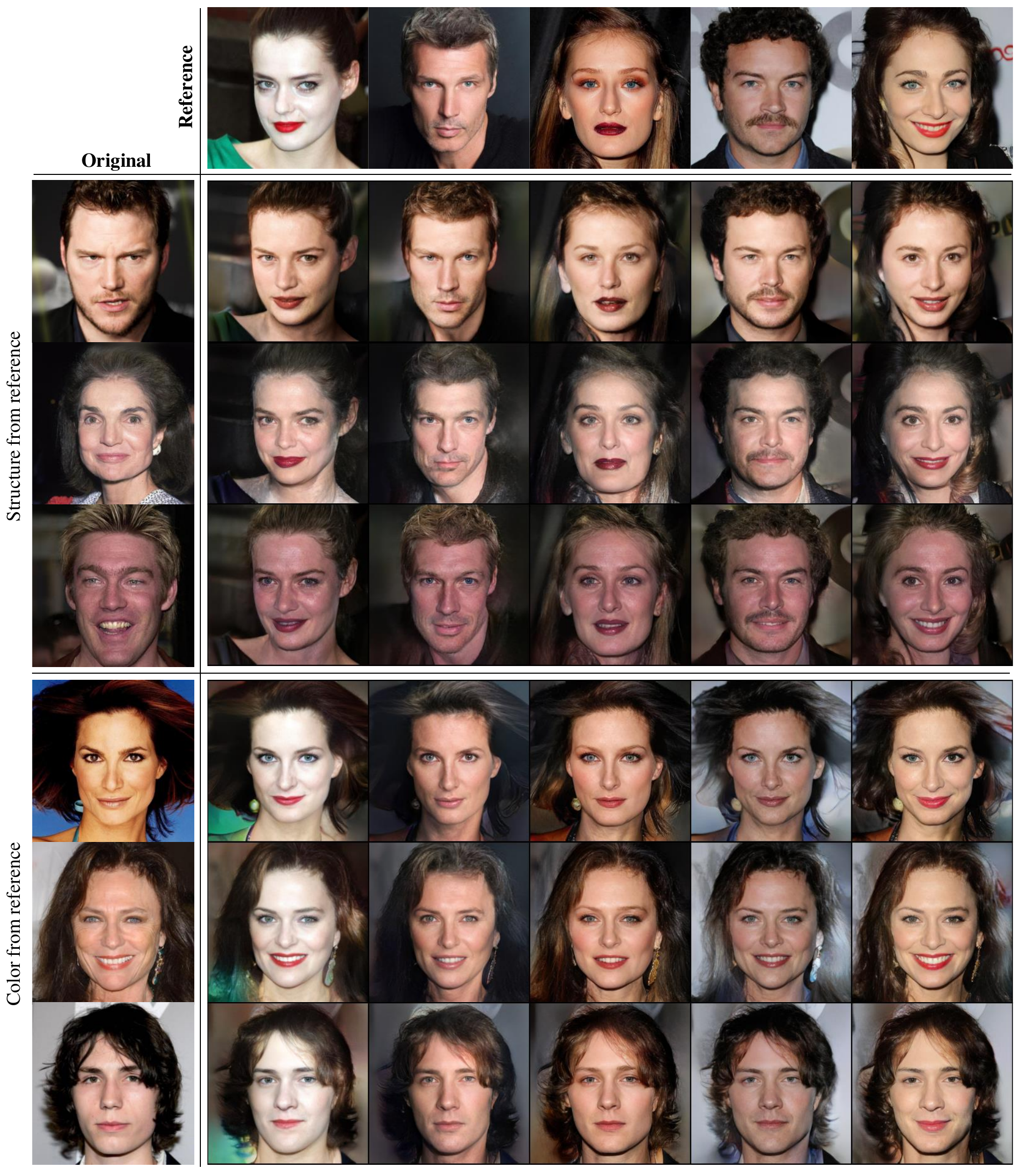}\\
\caption{The results of style mixing in our method.
Please refer to \ref{sec:supp_stylemixing} for details.
} 
\label{fig:suppfigStyleMixing}
\end{figure*}
}

\newcommand{\suppfigSemantic}{
\begin{figure*}[t]
\centering
\vspace{-8mm}
\includegraphics[width=1.0\linewidth]{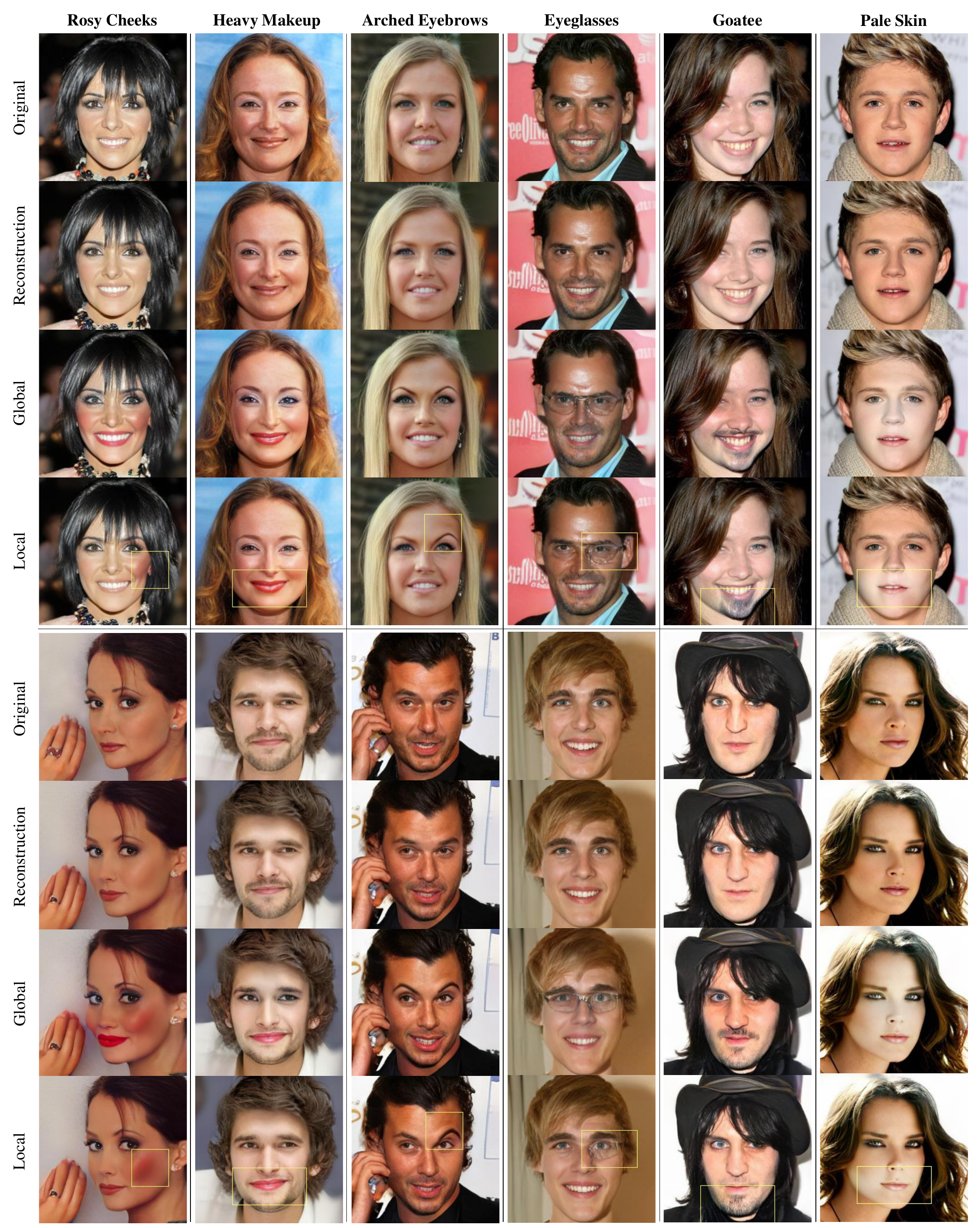}\\

\caption{The results of the global and local version of semantic manipulation in our method. In local manipulation, we apply the semantic direction in the yellow box area.} 
\label{fig:suppfigSemantic}
\end{figure*}
}

\newcommand{\suppfigFailInter}{
\begin{figure*}[t]
        
    \hspace{-5mm}    
    \begin{minipage}{1.05\linewidth}
        \centering
        \hspace{0.2mm}
        \makebox[\himg\linewidth][c]{~~~~~~~~~Input A}
        \makebox[\himg\linewidth][c]{~~~~~~~~~Inversion A}
        \makebox[\himg\linewidth][c]{~~~~~~~$\xleftarrow[]{~~~~~~~~~~~~~~~~~~~}$}\hfill
        \makebox[\himg\linewidth][c]{~~Interpolation}\hfill
        \makebox[\himg\linewidth][c]{$\xrightarrow[]{~~~~~~~~~~~~~~~~~~~}$~~}\hfill
        \makebox[\himg\linewidth][c]{Inversion B~~~~~~}\hfill 
        \makebox[\himg\linewidth][c]{Input B~~~~~~~~~~~}\hfill \\

        \includegraphics[width=\himg\linewidth]{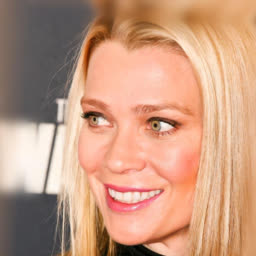}
        \includegraphics[width=\himg\linewidth]{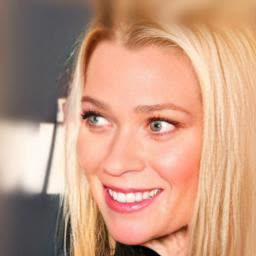}
        \includegraphics[width=\himg\linewidth]{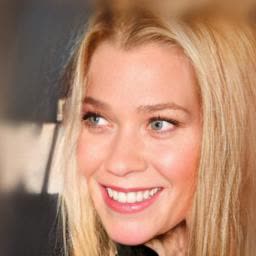}
        \includegraphics[width=\himg\linewidth]{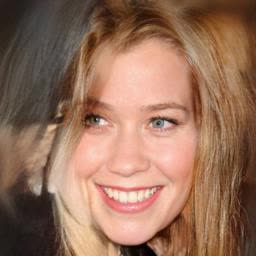}
        \includegraphics[width=\himg\linewidth]{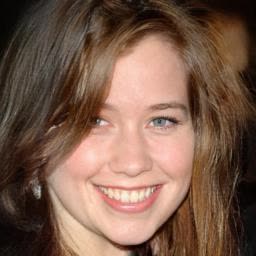}
        \includegraphics[width=\himg\linewidth]{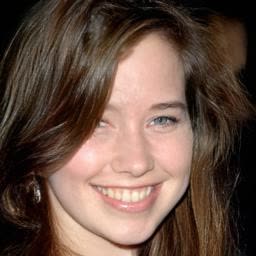}
        \includegraphics[width=\himg\linewidth]{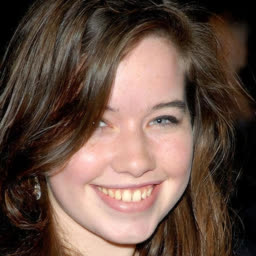} \\
        \vspace{1.0mm}

        \includegraphics[width=\himg\linewidth]{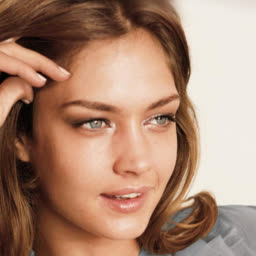}
        \includegraphics[width=\himg\linewidth]{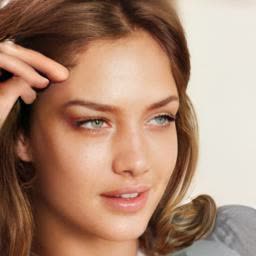}
        \includegraphics[width=\himg\linewidth]{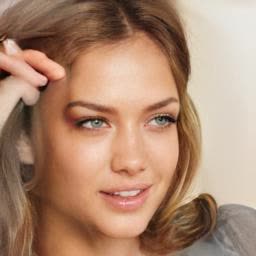}
        \includegraphics[width=\himg\linewidth]{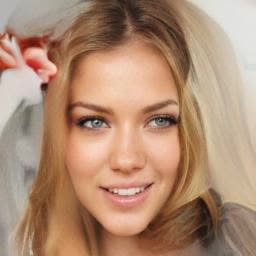}
        \includegraphics[width=\himg\linewidth]{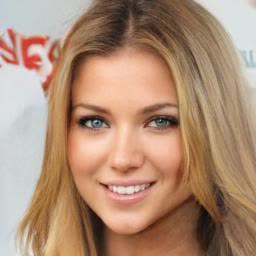}
        \includegraphics[width=\himg\linewidth]{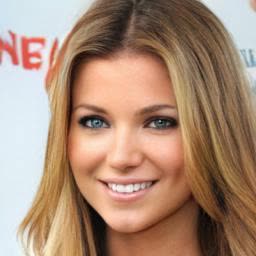}
        \includegraphics[width=\himg\linewidth]{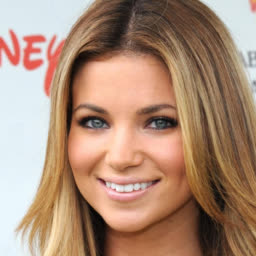} \\
        \vspace{1.0mm}

    \end{minipage}
\caption{Failure cases of interpolation in our method due to extreme pose difference.}
\label{fig:suppfigFailInter}
\end{figure*}
}

\newcommand{\suppfigFailTrans}{

\newcommand{\hfail}{27.7685mm}
\begin{figure*}[t]
\centering
\begin{minipage}[t]{\linewidth}
\makebox[\hh][c]{}\hspace{0mm}%
\makebox[\hfail][c]{\textbf{\small{\ Reference~~~~~}}}\hspace{-1.0mm}
\rotatebox[origin=l]{90}{\makebox[0mm][l]{\hspace*{0.045\linewidth}\textbf{\small{\raisebox{0.5mm}[0mm][0mm]{Original~~}}}}}\hspace{1.0mm}%
\includegraphics[height=\hfail]{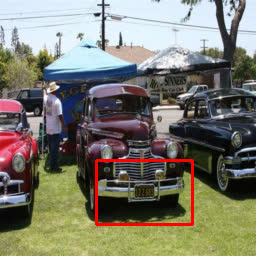}
\includegraphics[height=\hfail]{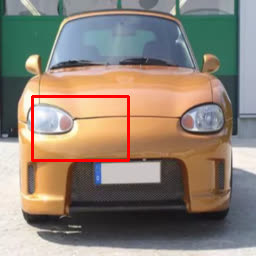}
\includegraphics[height=\hfail]{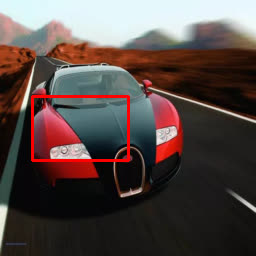}
\includegraphics[height=\hfail]{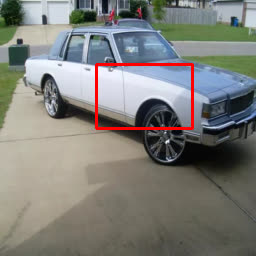}
\includegraphics[height=\hfail]{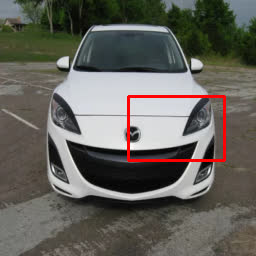}
\vspace{1mm}\\
\makebox[\hh][c]{}\hspace{-2.0mm}%
\includegraphics[height=\hfail]{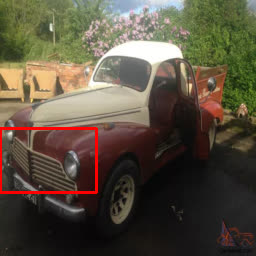}\hspace{2.0mm}
\includegraphics[height=\hfail]{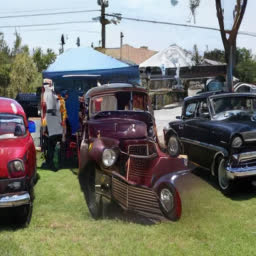}
\includegraphics[height=\hfail]{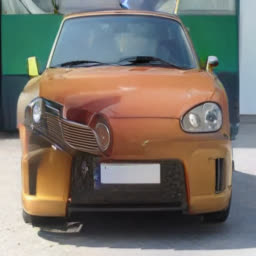}
\includegraphics[height=\hfail]{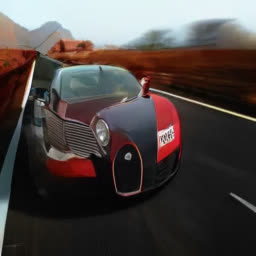}
\includegraphics[height=\hfail]{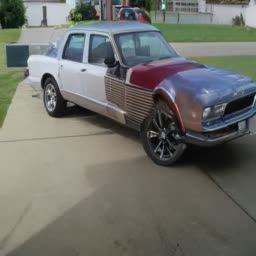}
\includegraphics[height=\hfail]{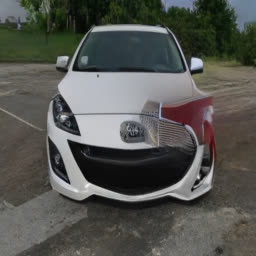}

\end{minipage}

\caption{Failure cases of transplantation in our method due to different sizes of the masks.
}
\label{fig:suppfigFailTrans}
\end{figure*}
}

\newcommand{\taboneKcomparison}{
\renewcommand{\himg}{0.131}
\begin{table*}[t]
\centering
\begin{minipage}{\linewidth}
    \centering
    \hspace{6mm}
    \makebox[\himg\linewidth][c]{Input A}
    \makebox[\himg\linewidth][c]{Inversion A}
    \makebox[\himg\linewidth][c]{$\xleftarrow[]{~~~~~~~~~~~~~~~~~~~}$}\hfill
    \makebox[\himg\linewidth][c]{Interpolation}\hfill
    \makebox[\himg\linewidth][c]{$\xrightarrow[]{~~~~~~~~~~~~~~~~~~~}$~~}\hfill
    \makebox[\himg\linewidth][c]{Inversion B~~~~}\hfill 
    \makebox[\himg\linewidth][c]{Input B~~~~~~}\hfill 
    
    \rotatebox{90}{\makebox[20mm][c]{\small{~~~~~Im2StyleGAN (\arch{a})}}}\vspace{-0.35mm}\hspace{0.5mm}
    \includegraphics[width=\himg\linewidth]{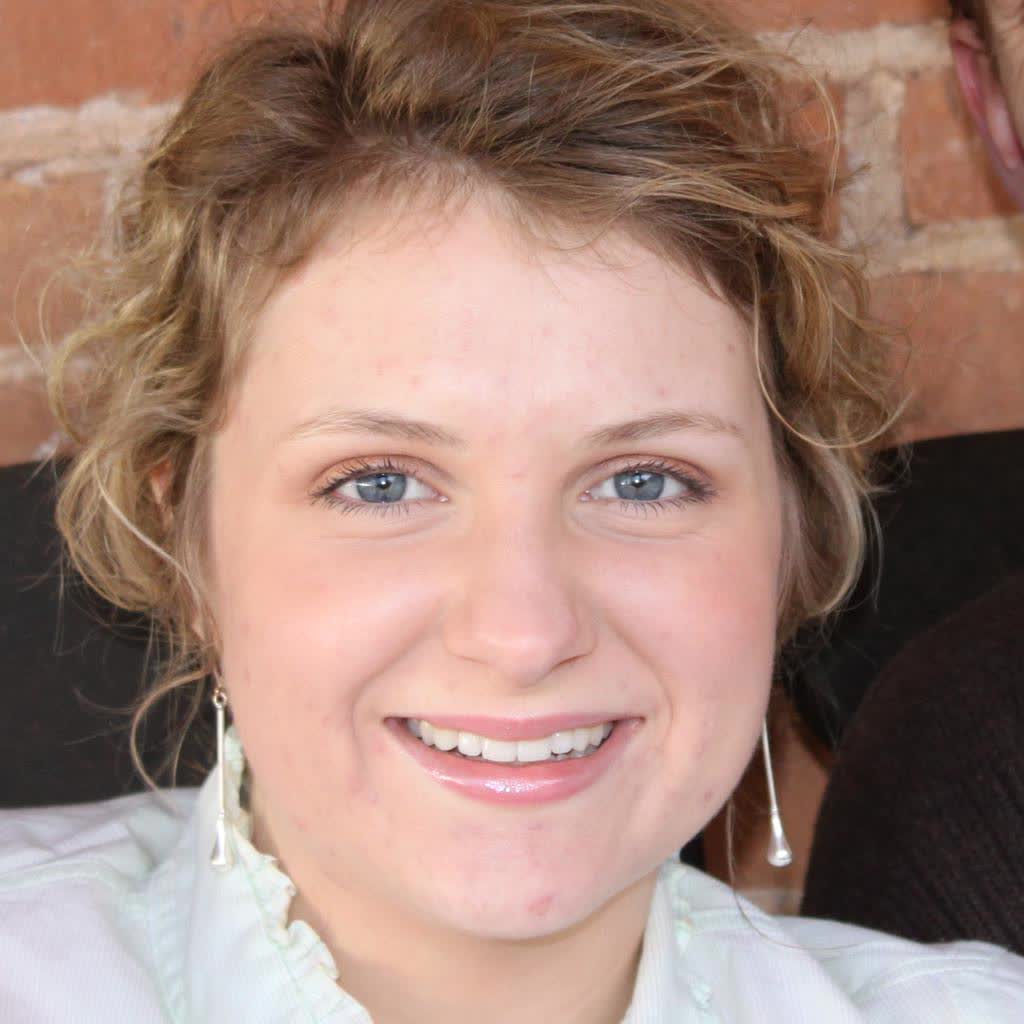}
    \includegraphics[width=\himg\linewidth]{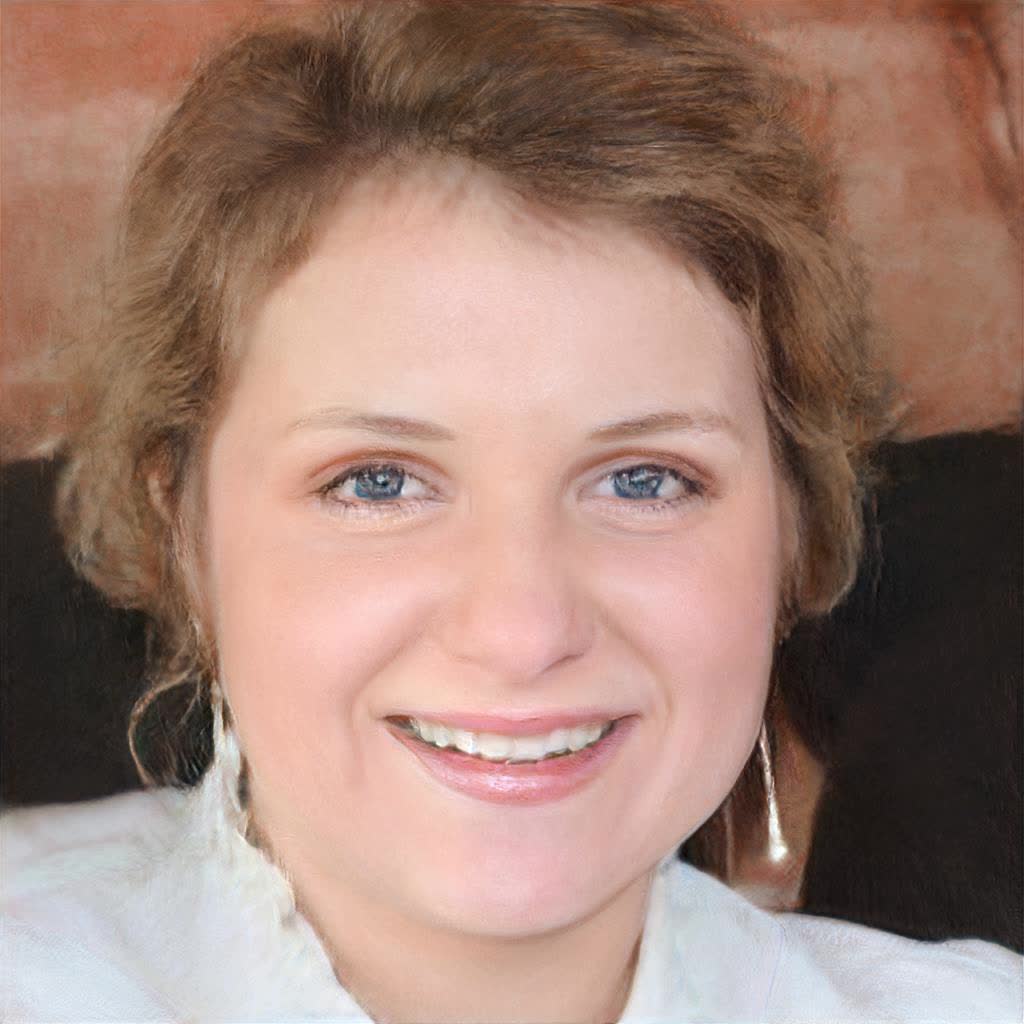}
    \includegraphics[width=\himg\linewidth]{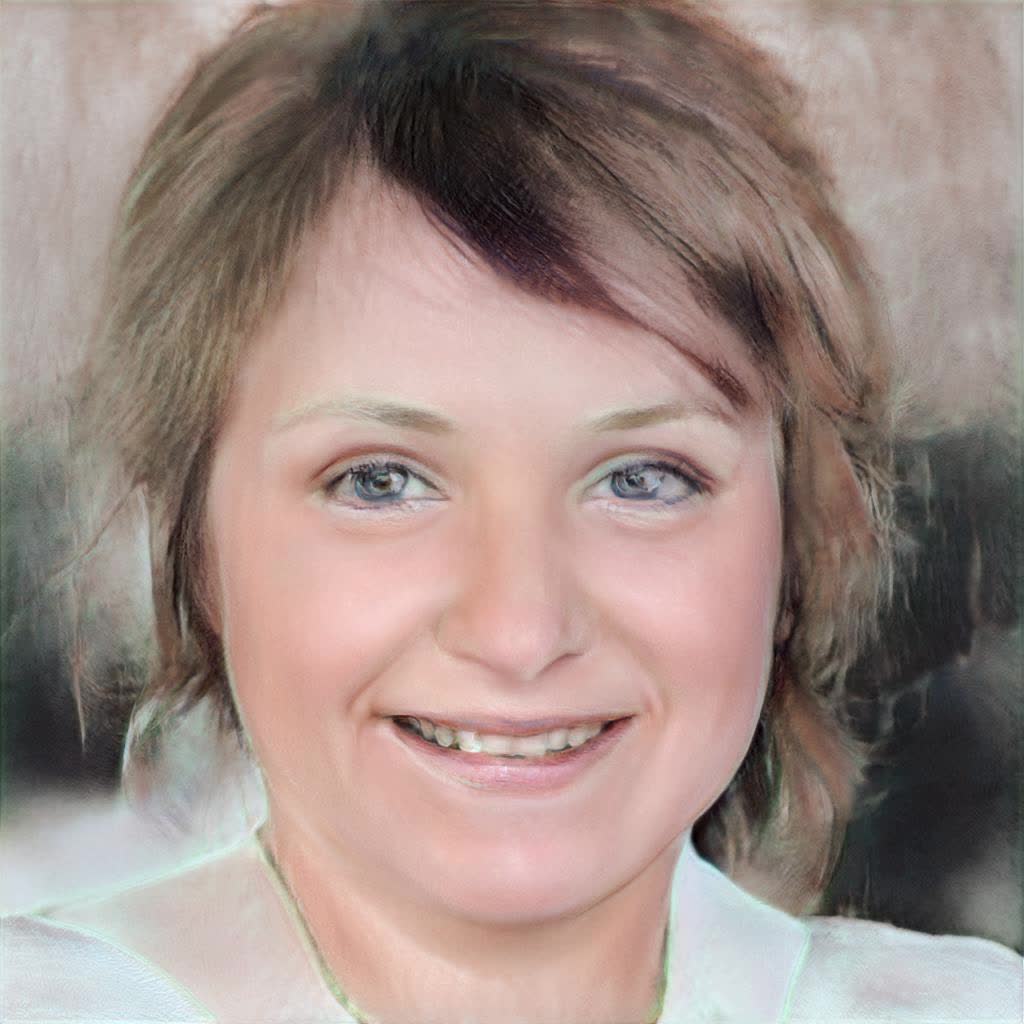}
    \includegraphics[width=\himg\linewidth]{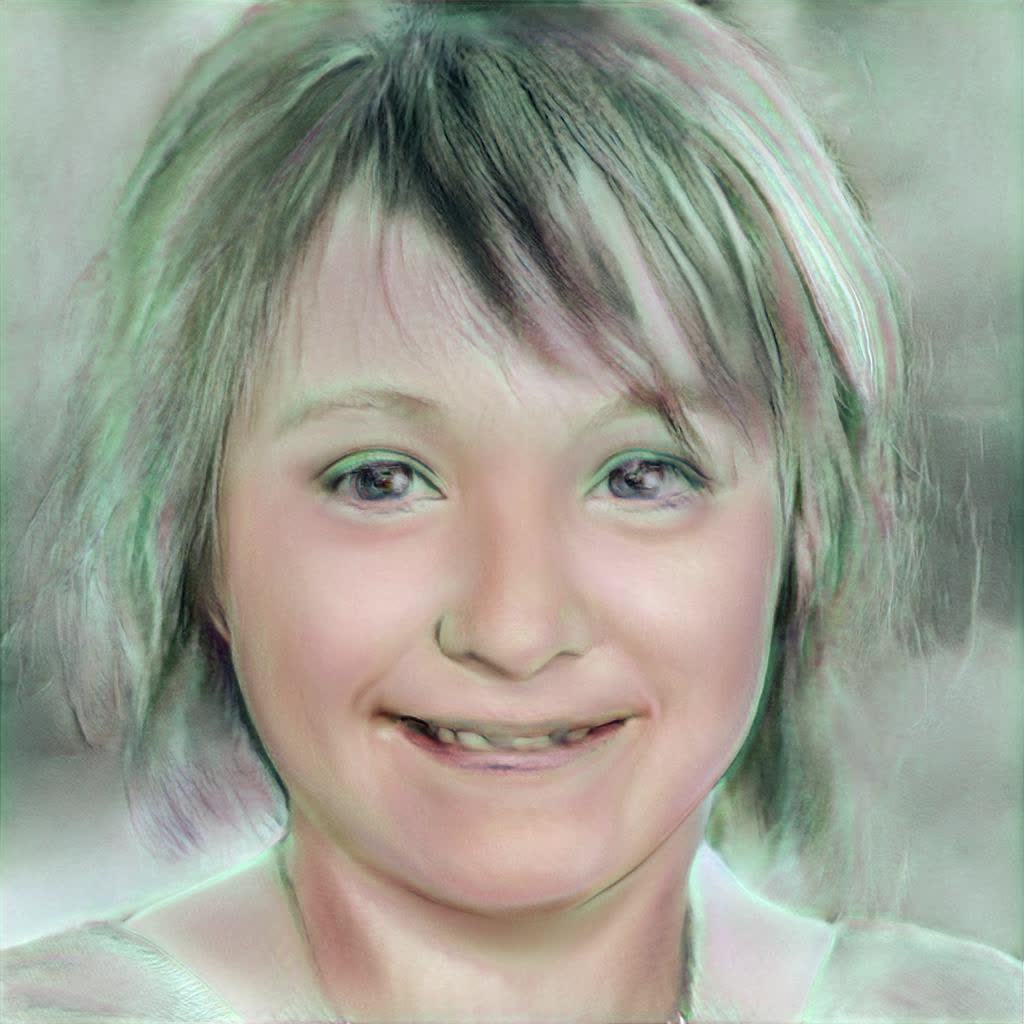}
    \includegraphics[width=\himg\linewidth]{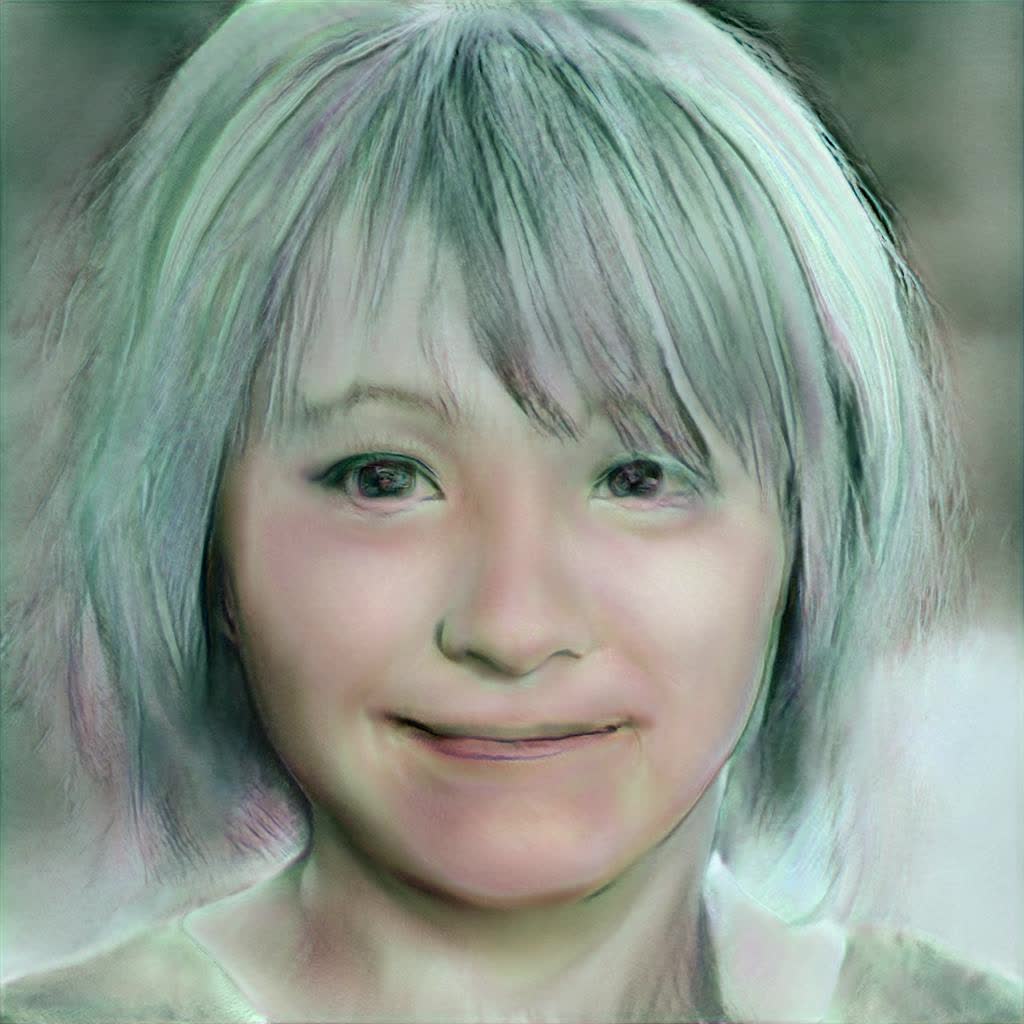}
    \includegraphics[width=\himg\linewidth]{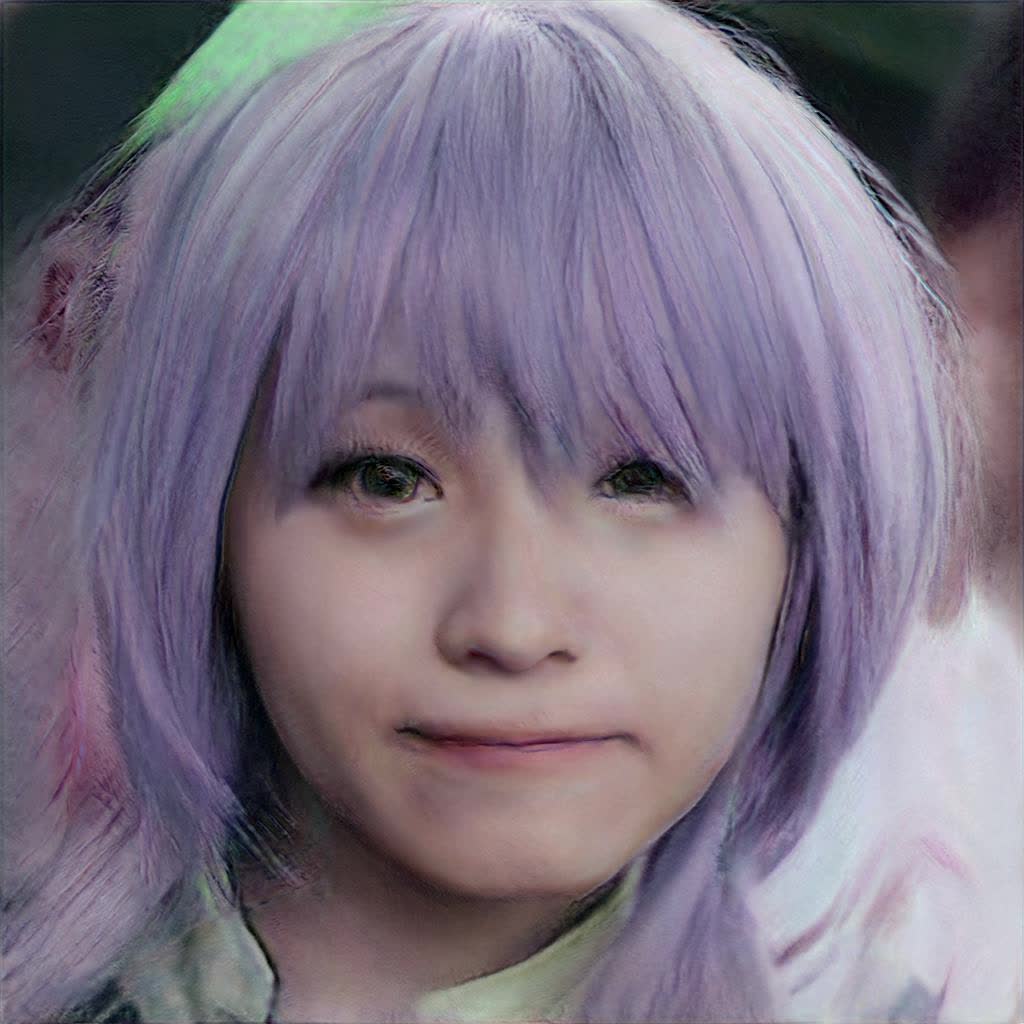}
    \includegraphics[width=\himg\linewidth]{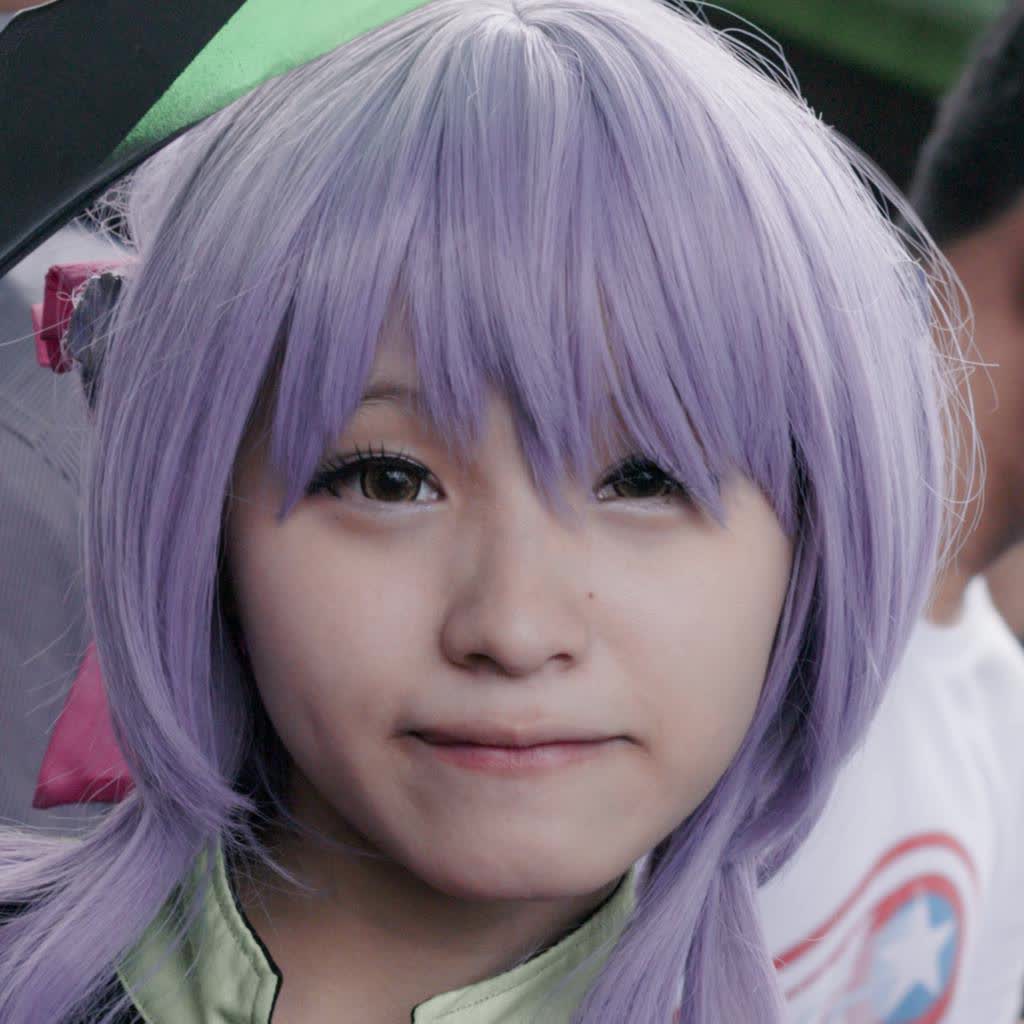}
    
    \hspace{-1.0mm}
     \rotatebox{90}{\makebox[20mm][c]{\small{~~~Ours-Light (\arch{e})}}}\vspace{-0.4mm}\hspace{0.65mm}
    \includegraphics[width=\himg\linewidth]{supp_figures/1024/w_interpolation/src.jpg}
    \includegraphics[width=\himg\linewidth]{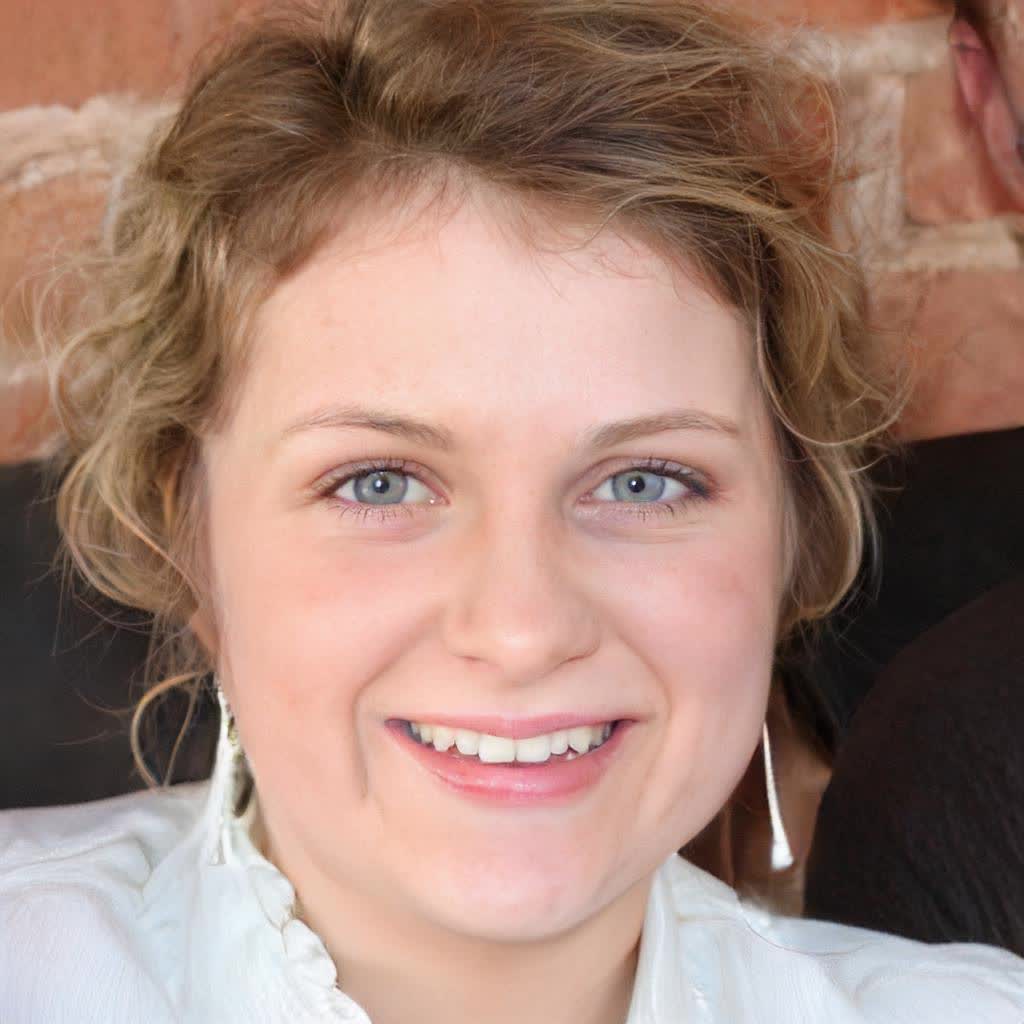}
    \includegraphics[width=\himg\linewidth]{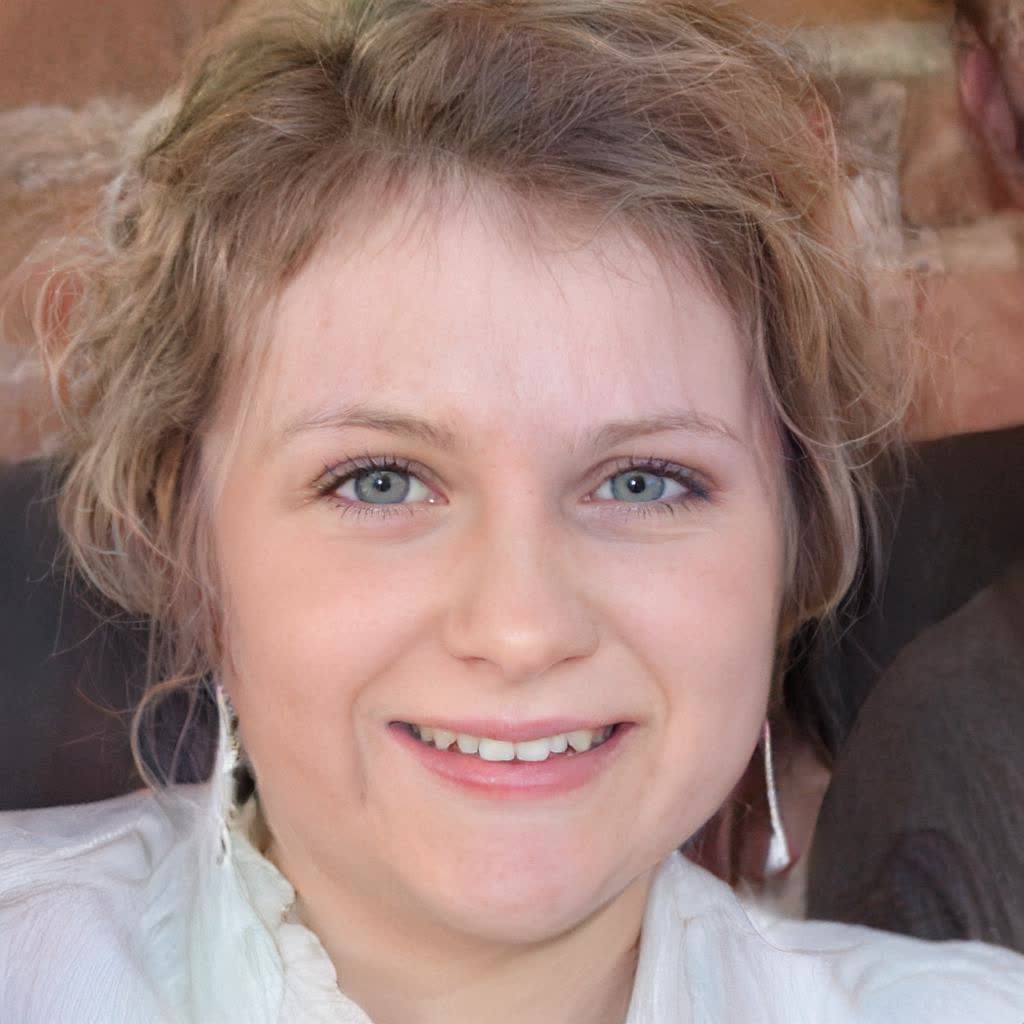}
    \includegraphics[width=\himg\linewidth]{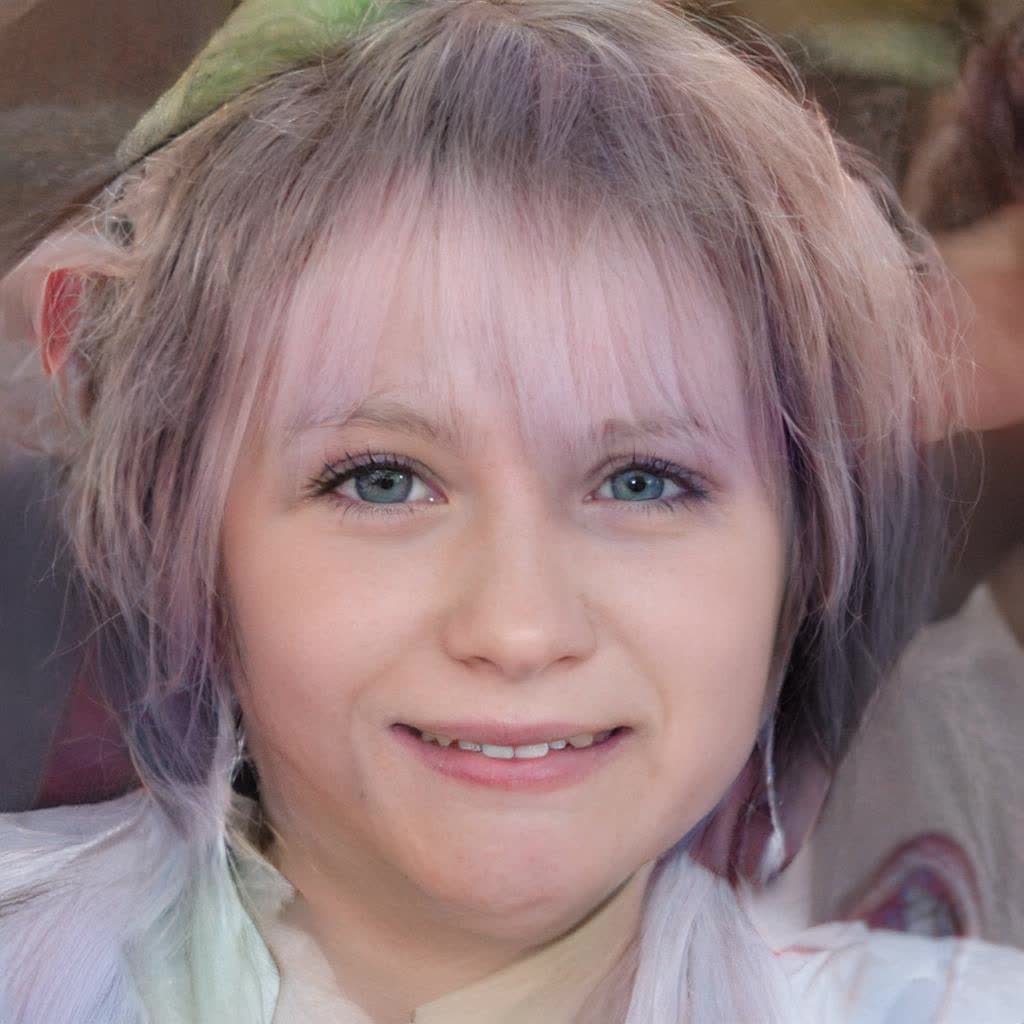}
    \includegraphics[width=\himg\linewidth]{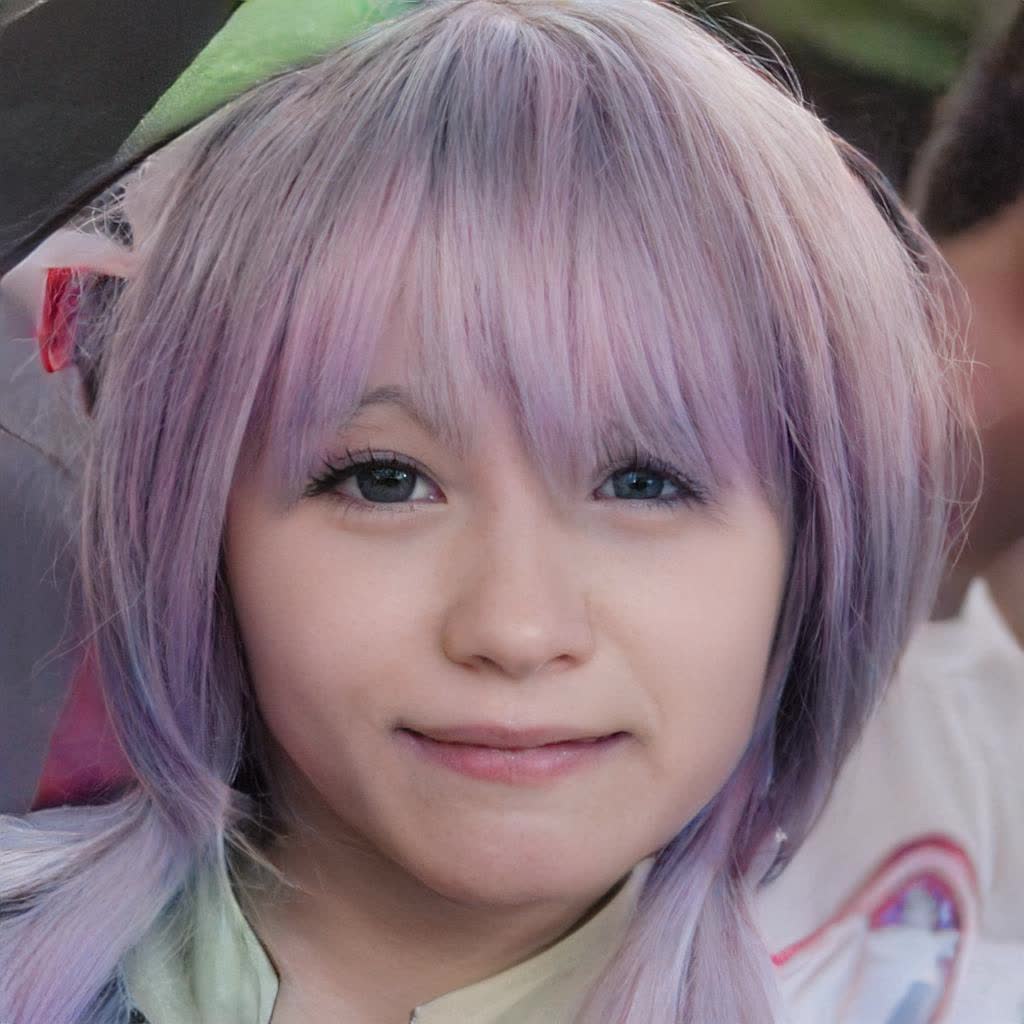}
    \includegraphics[width=\himg\linewidth]{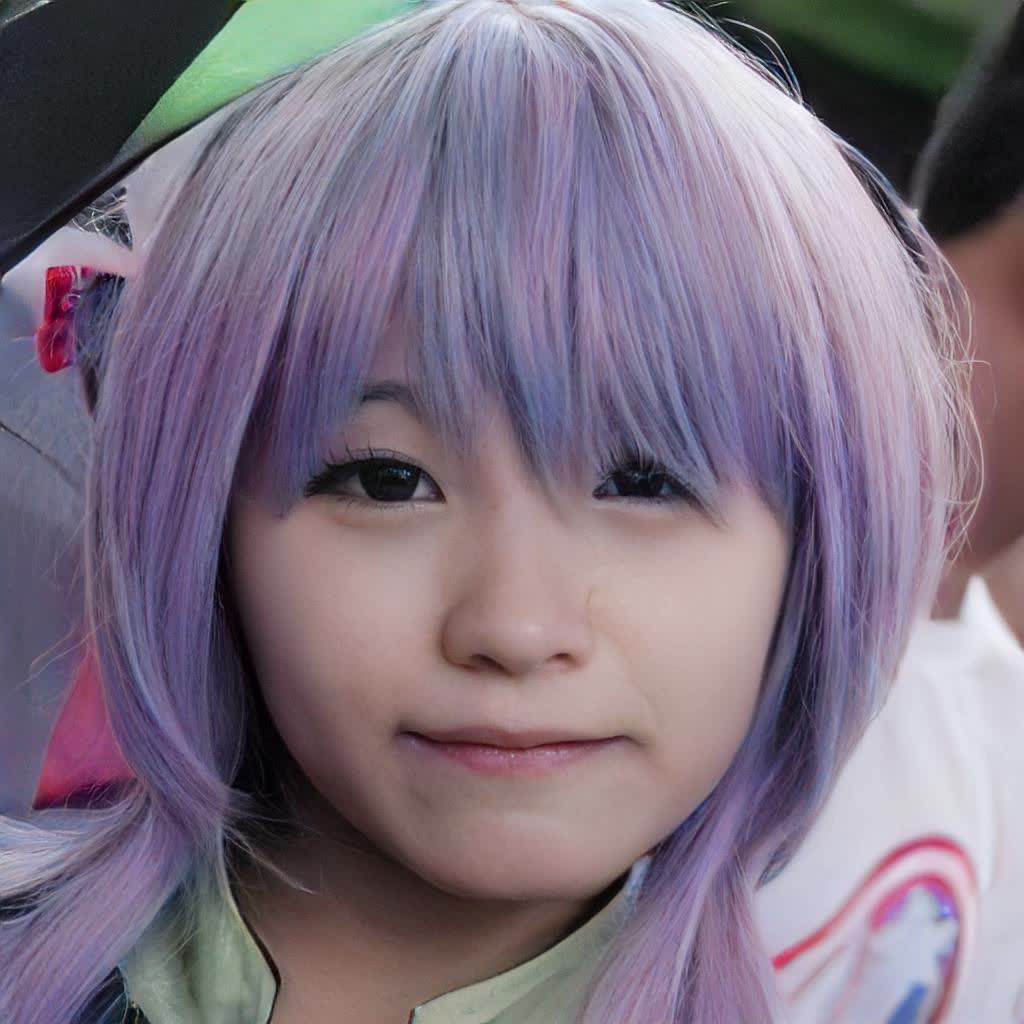}
    \includegraphics[width=\himg\linewidth]{supp_figures/1024/w_interpolation/ref.jpg}

    \hspace{-0.9mm}
     \rotatebox{90}{\makebox[20mm][c]{\small{~~~Ours (\arch{f})}}}\vspace{0mm}\hspace{0.65mm}
    \includegraphics[width=\himg\linewidth]{supp_figures/1024/w_interpolation/src.jpg}
    \includegraphics[width=\himg\linewidth]{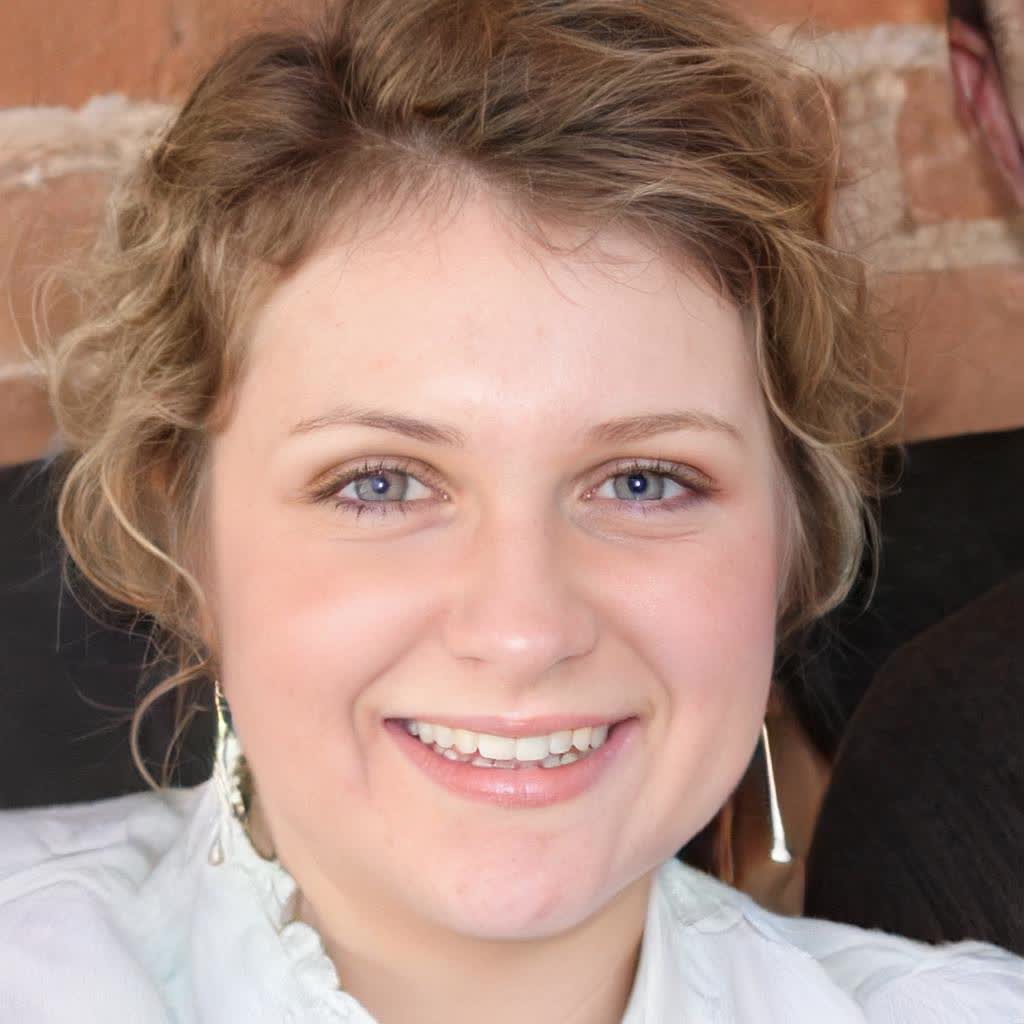}
    \includegraphics[width=\himg\linewidth]{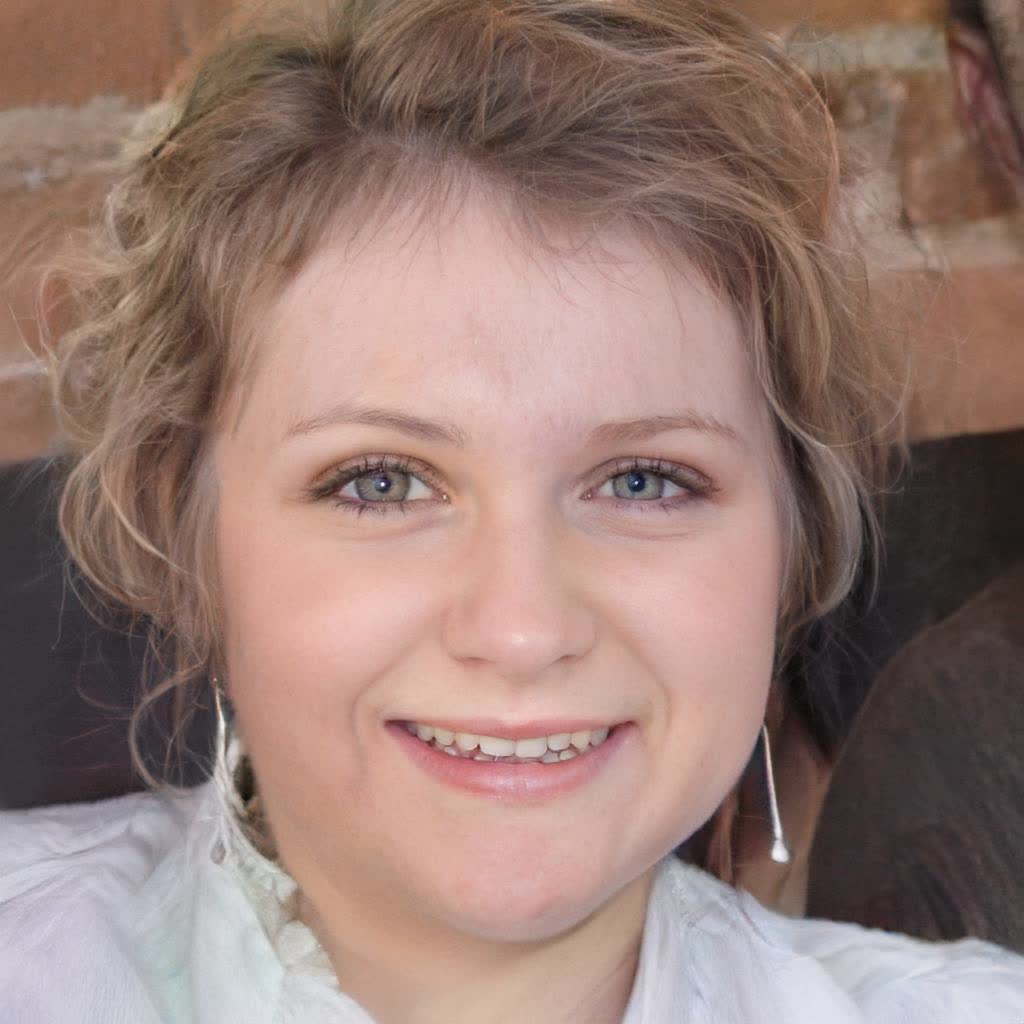}
    \includegraphics[width=\himg\linewidth]{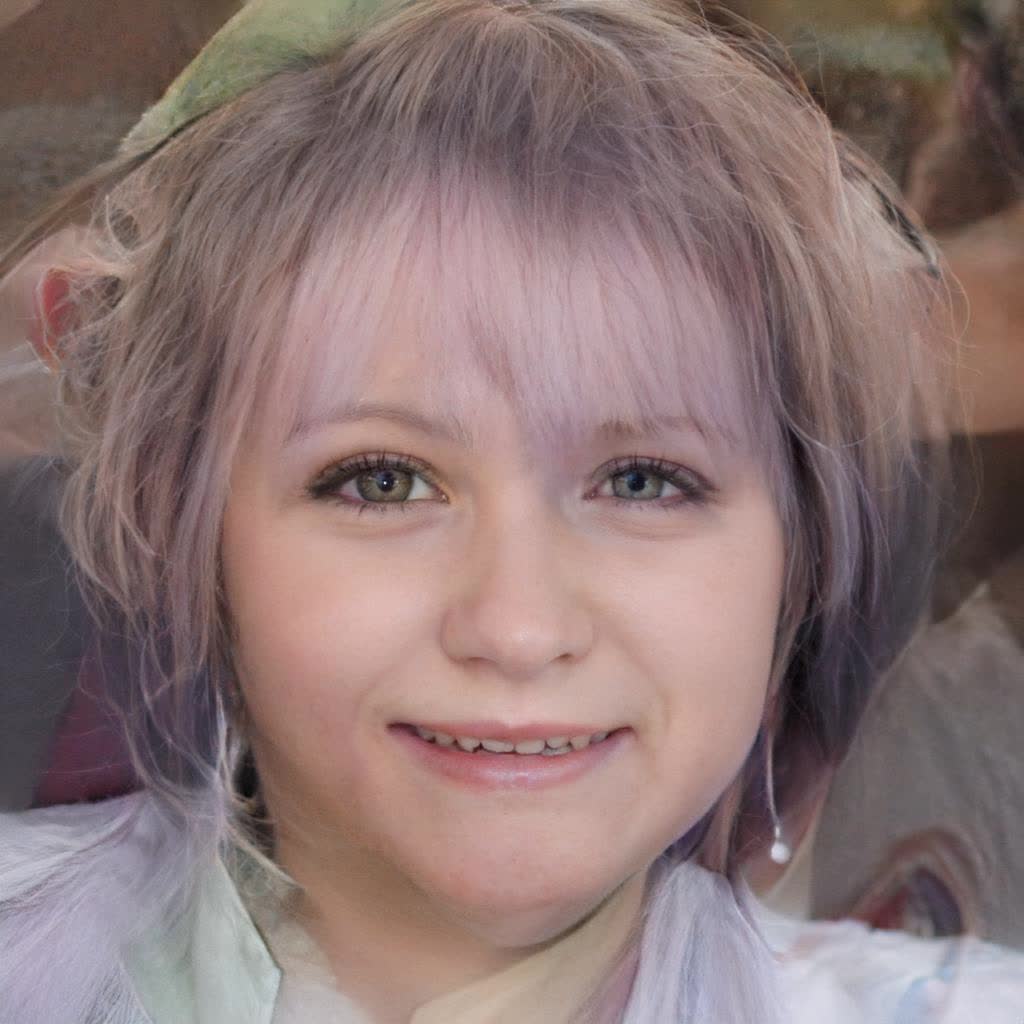}
    \includegraphics[width=\himg\linewidth]{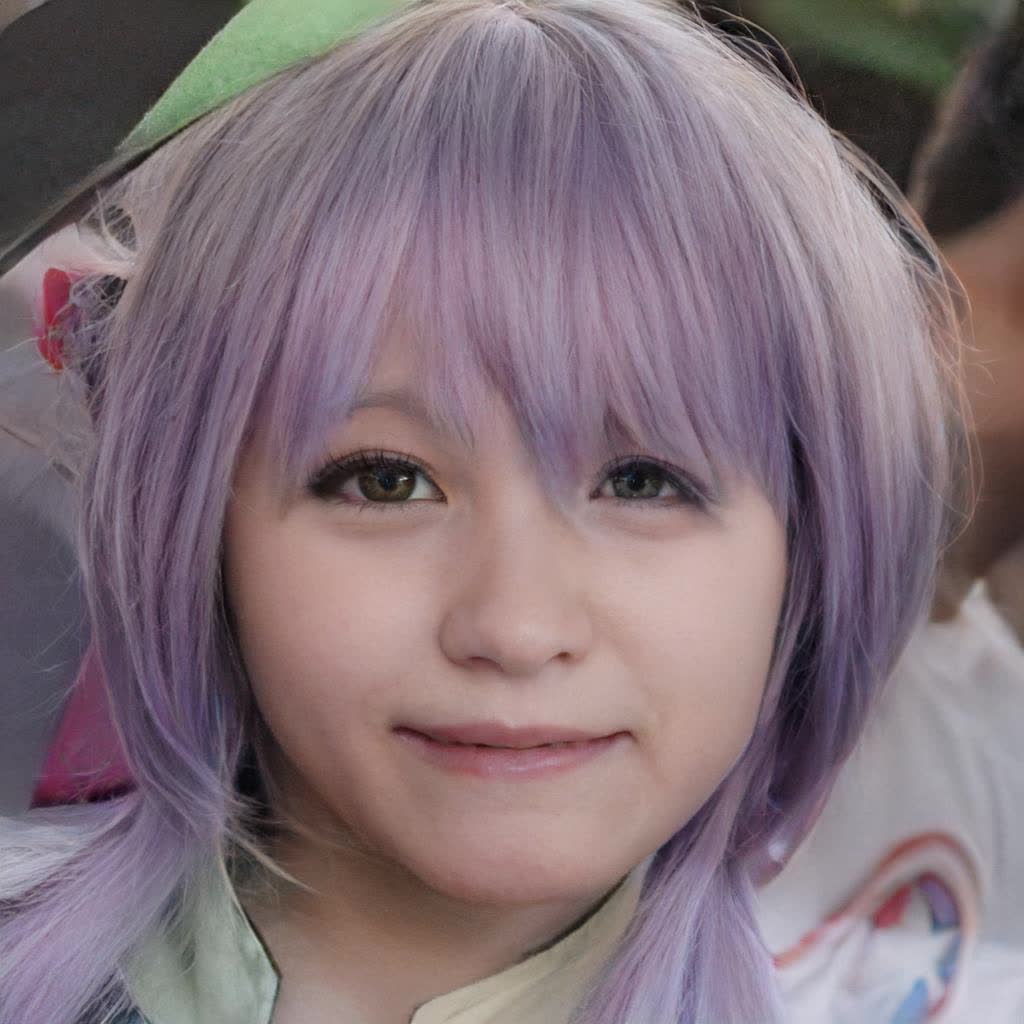}
    \includegraphics[width=\himg\linewidth]{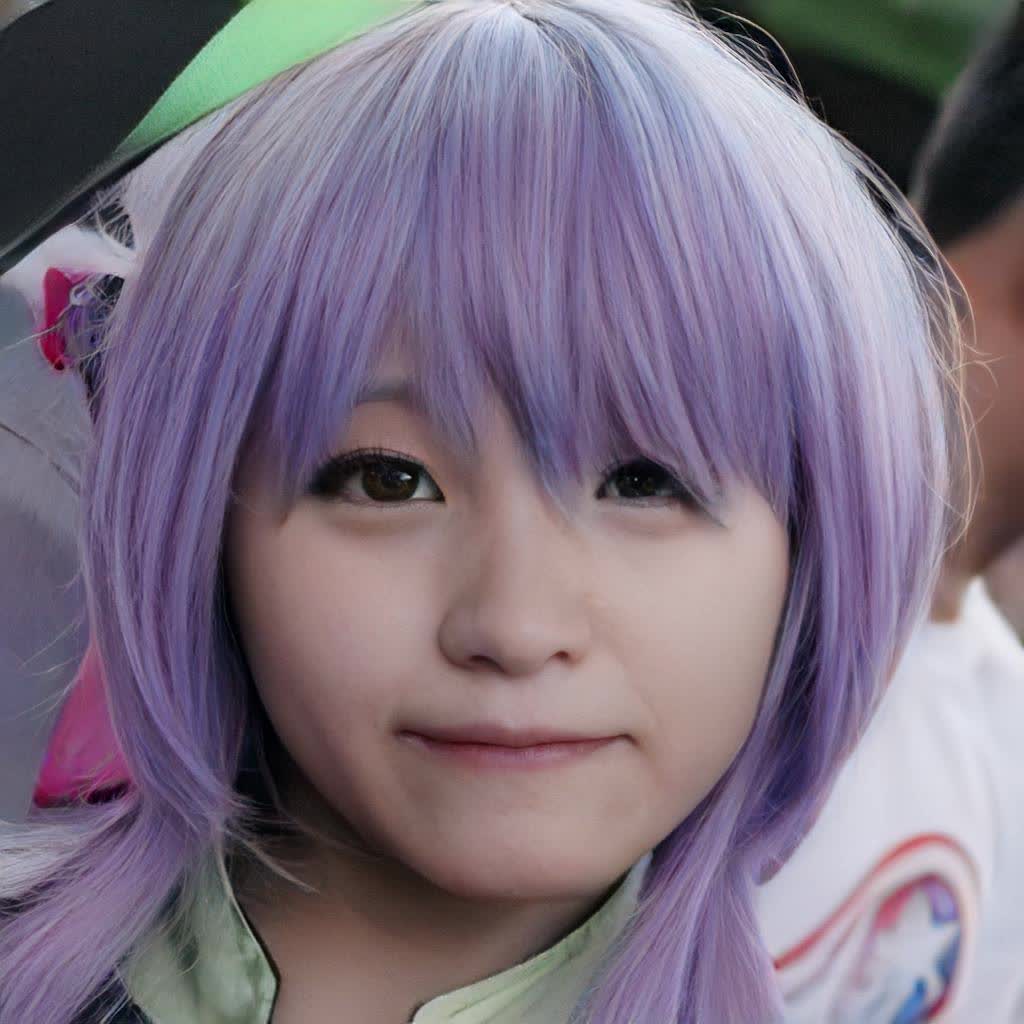}
    \includegraphics[width=\himg\linewidth]{supp_figures/1024/w_interpolation/ref.jpg}
 
    \vspace{2mm}
    
\end{minipage}
\newcolumntype{x}{>{\centering\arraybackslash\hspace{0pt}}p{10.5mm}}
\resizebox{1.0\linewidth}{!}{
\begin{tabular}{|l|l|c|c|c|c|c|c|c|c|}
\hline
~~~Network & Projection & Joint learning & G param(M) & runtime(s) & G GPU(GB) & MSE & LPIPS  & \FIDlerp \\
\hline\hline
\arch{a} StyleGAN2 & Image2StyleGAN & \xmark & 30.4  & 454.7 & 2.1 & 0.021 & 0.468 & 38.00  \\
\arch{b} StyleGAN2 & StyleGAN2 & \xmark & 30.4  & 142.2  & 2.1 & 0.093 & 0.467 & 34.65  \\
\hline\hline
\arch{d} StyleMapGAN-Light & Encoder & \xmark  & 18.6  &  0.253   &  3.0   & 0.071 & 0.546 & 201.18 \\
\arch{e} StyleMapGAN-Light & Encoder & \cmark  &  18.6  & 0.253  &  3.0   & 0.017 & 0.347 & \textbf{13.52} \\
\arch{f} StyleMapGAN & Encoder & \cmark  & 46.4 &  0.249 & 3.1 & \textbf{0.016} &  \textbf{0.344} & 13.68  \\
\hline
\end{tabular}
}
\caption{Comparison with StyleGAN2 on $1024\times1024$ FFHQ. We also explored the effect of other components such as generator size and joint learning. StyleMapGAN-Light has reduced the number of channels of the stylemap resizer. $32\times32$ stylemap is used for the large size of the image. ``G'' denotes the generator.}
\label{tab:1K_tab}
\end{table*}
}

\newcommand{\tabLoss}{
\begin{table}[h]
\small
\centering
\resizebox{1.0\columnwidth}{!}{
\begin{tabular}{|l|c|c|c|c|}
\hline
Removed loss & MSE & LPIPS & FID & \FIDlerp \\
\hline\hline
Adversarial loss & \textbf{0.009} & \textbf{0.137} & 278.87 & 12.99 \\
Domain-guided loss & 0.013 & 0.193 & 5.11  & 16.84 \\

Latent reconstruction loss & 0.021 & 0.220 & \textbf{4.43}  & 10.08 \\

Image reconstruction loss & 0.029 & 0.254 & 5.01  & 10.29 \\
Perceptual loss & 0.033 & 0.304 & 5.34  & 13.33 \\

R1 regularization & 0.097 & 0.403 & 31.82  & 14.56 \\

\hline \hline
Train with all losses & 0.023 & 0.237 & 4.72  & \textbf{9.97} \\
\hline

\end{tabular}
}
\caption{Loss ablation study removing one loss at a
time. We used CelebA-HQ, $256\times256$ image, and $8\times8$ stylemap.}
\label{tab:tabloss}
\end{table}
}

\appendix
\twocolumn[{
\renewcommand\twocolumn[1][]{#1}
\maketitle
\begin{center}
    \centering 
\centering
\includegraphics[width=1.0\linewidth]{supp_figures/spatial_mixing.pdf}\\
\captionof{figure}{Our local editing starts with a learned encoder for fast image-to-stylemap projection. We estimate the stylemaps $\w$ and $\wt$ of the original $\x$ and the reference $\xt$ and transform them into multiple resolutions through the learned stylemap resizer. For each resolution, we calculate the alpha blending of the two stylemaps using the user-defined binary mask $\m$. Finally, the learned generator produces the output using the spatially-mixed stylemaps. The right one shows an example generated using our method.} 
\label{fig:figspatialmixing}
\end{center}
}]
\section{Local editing in $\w^+$ space}
\label{sec:w_plus_blending}

This section illustrates how we perform local editing using StyleMapGAN. Although we already described the local editing method in Section 3.3 of the paper, it is impossible to edit in detail due to the coarse mask resolution ($8 \times 8$). Contrary to the previous method, we propose a local editing method in $\w^+$ space. Regardless of the resolution of stylemap ($\w$), we can exploit detailed masks with resized stylemaps ($\w^+$) in high resolutions.

\smallskip

\Fref{fig:figspatialmixing} shows the overview of blending on the $\w^+$ space.
The edited $i$-th resized \stylemap \textbf{\"w}$^+$ is an alpha blending of $\w^+$ and $\wt^+$:
\begin{equation}
\text{\textbf{\"w}}^+_i=\m_i \otimes {\wt_i}^+ \oplus (1-\m_i) \otimes {\w_i}^+
\label{eqn::local_editing_w_plus}
\end{equation}
where $i$-th resized mask $\m_i$ is shrunk by max pooling. Although the mask's shape does not align with the $8\times8$ \stylemap, we can precisely blend the two images on the $\w^+$ space.

\taboneKcomparison

\section{Experiments in the high-resolution dataset}
\label{sec:1K_comparison}

We evaluate our model on FFHQ at $1024\times1024$ resolution. Baseline is StyleGAN2, and we also test Image2StyleGAN (\arch{a}). StyleGAN2 official pretrained network is used for a fair comparison. StyleMapGAN adopts $32\times32$ stylemap for the high-resolution dataset, compared to $8\times8$ stylemap for $256\times256$ image. StyleMapGAN-Light (\arch{e}) is a light version of StyleMapGAN; it reduces the number of parameters of the generator. Another training setting (\arch{d}) is sequential learning, which trains the generator first and then trains the encoder. In \Tref{tab:1K_tab}, we used the same protocol as the paper to calculate MSE, LPIPS, and \FIDlerp. The number of training images for FFHQ is 69K, and we limited the test and validation set to 500 images.

\paragraph{Comparison with baselines.}
As shown in \Tref{tab:1K_tab}, Image2StyleGAN reconstructs the image well, but it struggles with poor interpolation quality. Low \FIDlerp, rugged interpolation results and lengthy runtime shows Image2StyleGAN is not suitable for image editing tasks. StyleMapGAN outperforms baselines in all metrics, and even StyleMapGAN-Light shows astonishing results.

\paragraph{StyleMapGAN-Light} is $2.5\times$ smaller than the original version. Stylemap resizer accounts for a large portion of the network's size, so we reduce the number of channels of feature maps in the stylemap resizer. The reconstruction image lacks some detail, but StyleMapGAN-Light still outperforms baselines, and \FIDlerp is even better than the original version. Please see our code to refer to the number of channels.

\paragraph{Joint learning} is important when training StyleMapGAN. It makes training stable and network performance better. Training the encoder after training the generator fails to reconstruct images. We speculate the reason why joint learning is better than sequential learning as follows. In joint learning, the generator and the encoder affect each other. The generator generates an image that would be easy to reconstruct by the encoder. The structure of the encoder is a stack of convolutional layers, which makes the projected stylemap is prone to have local correspondence: Partial change in the stylemap leads to local editing on the image. Through joint learning, the mapping network in the generator also learns to make the stylemap from Gaussian distribution have the local correspondence.

\section{Implementation details}
\label{sec:implementation_details}
\figAblationMapping
\paragraph{Architecture.}
We follow StyleGAN2~\cite{karras2020stylegan2} regarding the discriminator architecture and the feature map counts in the convolutional layers of the synthesis network. Our mapping network is an MLP with eight fully connected layers followed by a reshape layer. The channel sizes are 64, except the last being 4,096. Our encoder adopts the discriminator architecture until the $8\times8$ layer and without minibatch discrimination~\cite{salimans2016improved}.

\paragraph{Training.}
We jointly train the generator, the encoder, and the discriminator. It is simpler and leads to more stable training and higher performance than separately training the adversarial networks and the encoder, as described in \sref{sec:1K_comparison}. 
For the rest, we mostly follow the settings of StyleGAN2, \eg, the discriminator architecture, R1 regularization~\cite{mescheder2018r1reg} in the discriminator using $\gamma = 10$, Adam~\cite{kingma2014adam} optimizer with 0.002 learning rate, $\beta_{1}=0.0$ and $\beta_{2}=0.99$, an exponential moving average of the generator and the encoder, leaky ReLU~\cite{maas2013lrelu}, equalized learning rate~\cite{karras2017progressivegan} for all layers, random horizontal flip for augmentation, and reducing the learning rate by two orders~\cite{karras2019stylegan} of magnitude for the mapping network.
Our code is based on unofficial PyTorch implementation of StyleGAN2\footnote{https://github.com/rosinality/stylegan2-pytorch}.
All \stylemapgan variants at $256\times256$ are trained for two weeks on 5M images with 2 Tesla V100 GPUs using a minibatch size of 16. In \sref{sec:1K_comparison}, $1024\times1024$ models are trained for one week on 2.5M images with 8 Tesla V100 GPUs using a minibatch size of 16.
We note that most cases keep slowly improving until 10M images.
Our code is publicly available online for reproducibility\footnote{https://github.com/naver-ai/StyleMapGAN}.

\paragraph{Mapping network design for stylemap.}

There are several choices when designing a mapping network. We can remove the mapping network so that our method does not generate images from the standard Gaussian distribution and uses only real images for training like autoencoder~\cite{hinton2006reducing}. As shown in \Fref{fig:figAblationMapping}, autoencoder fails to produce realistic images using the projected \stylemap. It seems to copy and paste between two images on RGB space. The autoencoder uses only images as input, which is a discrete variable. On the contrary, our method uses not only images but also the latent from Gaussian distribution, which is a continuous space. If we mix two latent codes for editing the image, training with continuous latent space can cover more latent values than discrete latent space.

\medskip

Alternatively, we can easily think of convolutional layers due to the spatial dimensions of the \stylemap. But, the mapping network with convolutional layers struggles in reconstruction so that the edited results images are quite different from the original images. We assume that there is such a limit because the convolutional layer's mapping is bounded to the local area.
On the other hand, each weight and input in MLP are fully-connected so that it can make a more flexible latent space.

\section{Loss details}
\label{sec:contribution_loss}
In Section 3.2 of the main paper, we briefly introduced six losses. In this section, we provide details of the losses and their responsibilities. Some losses degrade reconstruction quality (MSE, LPIPS~\cite{zhang2018lpips}), but we need every loss for the best editing quality (\FIDlerp). We can obtain the best \FIDlerp by training with all losses. \Tref{tab:tabloss} shows the quantitative results of the ablation study. The coefficients of all loss terms are set to 1.

\figAblationLoss
\tabLoss

\paragraph{Adversarial loss.}

The discriminator tries to classify fake images as fake, which are generated randomly from Gaussian distribution or reconstruction of input images. On the contrary, the generator fools the discriminator by producing more realistic images. Generation from the continuous space increases generation power in terms of smooth interpolation. Without adversarial loss related to the mapping network, we can not obtain a smooth manifold of latent space as mentioned in \sref{sec:implementation_details}. \Fref{fig:figablationloss} also shows unnatural interpolation results and checkerboard artifacts if we don't use adversarial loss. We use the non-saturating loss~\cite{goodfellow2014gan} as our adversarial loss.

\paragraph{Domain-guided loss.}

Domain-guided loss is introduced by In-DomainGAN~\cite{zhu2020indomaingan}. We use an adversarial training manner on fake images generated from real images via the encoder and the generator. The discriminator tries to classify generated images as fake while the encoder and the generator attempt to fool the discriminator. The loss pushes the projected latent code to remain in the original latent space of GAN, which facilitates smooth real-image editing by exploiting GAN's properties (\eg, smooth interpolation). Without domain-guided loss, interpolation results are blurry as shown in \Fref{fig:figablationloss}.

\paragraph{Latent reconstruction loss.}

The goal of the encoder is to find the latent code which generates the target image. When we generate a fake image from Gaussian distribution, we know the pair of the latent code and the generated image. Using that supervision, we train the encoder like other approaches~\cite{ulyanov2017takes, srivastava2017veegan, pidhorskyi2020alae, zhu2020indomaingan}. The encoder tries to project images in the semantic domain of the original latent space and alleviates strong bias against pixel-level reconstruction.
 
\paragraph{Image reconstruction loss.}

To make the output image visually identical to the input image, we minimize differences between them at pixel-level. If we do not use this loss function, visual reconstruction fails as ALAE~\cite{pidhorskyi2020alae} does.

\paragraph{Perceptual loss.}

Image reconstruction loss often makes the encoder overfit and output blurry images. Several approaches~\cite{abdal2019image2stylegan, abdal2020image2stylegan, zhu2020indomaingan} adopt perceptual loss~\cite{johnson2016perceptual}, which exploits the features extracted by VGG~\cite{simonyan2015deep}, for perceptual-level reconstruction. We use LPIPS~\cite{zhang2018lpips} for perceptual loss, which has better feature representation.

\paragraph{R1 regularization.}

R1 regularization~\cite{mescheder2018r1reg} makes training stable. We find that lazy regularization~\cite{karras2020stylegan2} is enough and apply it every 16 steps for the discriminator. Without this loss function, performance degrades in all metrics.

\section{Additional results}
\label{sec:additional_results}
In this section, we show extensive qualitative results. \sref{sec:supp_random_generation} illustrates randomly generated images to show that the generation capability of our method does not degenerate compared to the baseline. \sref{sec:supp_reconstruction} and \sref{sec:supp_local_editing} provide expanded comparison on reconstruction and local editing, respectively. 
\sref{sec:supp_unaligned} shows additional unaligned transplantation examples. Our method is applicable to other latent-based editing methods as shown in \sref{sec:supp_semantic} and \sref{sec:supp_stylemixing}. Lastly, we discuss the limitations (\sref{sec:supp_failure}) of our method.

\subsection{Random generation}
\label{sec:supp_random_generation}
The primary objective of GANs is generating high-fidelity images from random Gaussian noise. We show random generation results for each dataset: CelebA-HQ~\cite{karras2017progressivegan}, AFHQ~\cite{choi2020starganv2}, and LSUN Car \& Church~\cite{yu2015lsun}. We use $8 \times 8$ resolution of stylemap except for AFHQ, in which case $16 \times 16$ resolution provides much better generation quality as shown in Table 2 of the main paper. To generate high-quality images, we use the truncation trick~\cite{brock2018biggan, kingma2018glow, Marchesi2017MegapixelSI} with $\psi = 0.7$. \Fref{fig:figRandomGeneration} shows uncurated images and FID values. In CelebA-HQ and AFHQ, we use the same FID protocol as in the main experiments; the number of generated samples equal to that of training samples. On the other hand, LSUN consists of a lot of training images so that we use 50k images randomly chosen from the training set; the number of generated samples is 50k, too. Low FIDs reveal that our method has 
satisfactory generation capability.

\figRandomGeneration

\medskip

\subsection{Image projection \& Interpolation}
\label{sec:supp_reconstruction}

Although encoder-based methods project images into latent space in real time, their projection quality falls short of expectations. \Fref{fig:suppfigReconstruction} shows the projection quality comparison between our method and other encoder-based baselines (ALAE~\cite{pidhorskyi2020alae}, In-DomainGAN~\cite{zhu2020indomaingan}, and SEAN~\cite{zhu2020sean}). 

\smallskip

\Fref{fig:suppfigInterpolation} shows projection and interpolation results of our method and Image2StyleGAN~\cite{abdal2019image2stylegan}. Although Image2StyleGAN reconstructs the input images in high-fidelity, it struggles in latent interpolation because its projection on $\w^+$ drifts from the learned latent space $\w$ of the generator.

\suppfigReconstruction

\suppfigInterpolation

\medskip

\subsection{Local editing}
\label{sec:supp_local_editing}
\Fref{fig:suppfigComparisonCeleb} shows local editing comparison with competitors. We eject two competitors (Structured Noise~\cite{alharbi2020structurednoise} and Editing in Style~\cite{collins2020editinginstyle}) due to their poor results, as shown in Figure 4 of the main paper. It is because they do not target editing real images but target fake images.

\suppfigComparisonCeleb

\medskip

\subsection{Unaligned transplantation}
\label{sec:supp_unaligned}

Figure~\ref{fig:suppfigUnalignedCar} and \ref{fig:suppfigUnalignedChurch} show unaligned transplantation results in LSUN Car \& Church~\cite{yu2015lsun}. Our method can transplant the arbitrary number and location of areas in reference images to the original images. Note that our method adjusts the color tone and structure of the same reference regarding the original images.

\suppfigUnalignedCar
\suppfigUnalignedChurch

\medskip

\subsection{Semantic manipulation}
\label{sec:supp_semantic}

\suppfigSemantic

We exploit InterFaceGAN~\cite{shen2020interfacegan} to find the semantic boundary in the latent space. Our method can change the semantic attribute using a certain direction derived from the boundary. We apply the direction on stylemap ($\w$ space). \Fref{fig:suppfigSemantic} shows two versions of semantic manipulation. The global version is a typical way to manipulate attributes. The local version is only available in our method due to the spatial dimensions of the stylemap. We apply the semantic direction on the specified location in the $\w$ space. It allows us not to change the undesired area of the original image regardless of attribute correlation. For example, ``Rosy Cheeks'' makes lips red and ``Goatee'' changes the color of noses in the global version but not in the local version as shown in \Fref{fig:suppfigSemantic}. Furthermore, we can change part of attributes such as lip makeup from ``Heavy Makeup'' and beard from ``Goatee''. It alleviates the hard labor of highly granular labeling.
Swap Autoencoder~\cite{park2020swapping} shows region editing that the structure code also can be manipulated locally. However, it can not apply region editing on some attributes (\eg, ``Pale Skin'') which are related to color and textures due to the absence of spatial dimensions in the texture code.

\medskip

\subsection{Style mixing}
\label{sec:supp_stylemixing}
\suppfigStyleMixing 

StyleGAN~\cite{karras2019stylegan} proposed the style mixing method, which copies a specified subset of styles from the reference image. We operate style mixing in resized stylemaps ($\w^+$). Unlike StyleGAN, our generator has color and texture information in the resized stylemaps of low resolution ($8 \times 8$). On the other hand, it generates overall structure through other resolutions ($16^2 - 256^2$). If we want to bring the color and texture styles from the reference image, we replace $8 \times 8$ resized stylemaps by reference. \Fref{fig:suppfigStyleMixing} shows the examples.

\mixshape
Using style mixing and unaligned transplantation, we can transfer local structure only as shown in Figure~\ref{fig:mix_shape}. We use the original image on the first resized stylemap and the reference image for the remaining resolutions in the target region.

\medskip

\subsection{Failure cases}
\label{sec:supp_failure}

Our method has a limitation when original and reference images have different poses and target semantic sizes. \Fref{fig:suppfigFailInter} shows failure cases on different poses. Especially, hair is not interpolated smoothly. \Fref{fig:suppfigFailTrans} shows failure cases on different target semantic sizes. The sizes and poses of the bumper vary, and our method can not transplant it naturally. This limitation gets worse when the resolution of the stylemap increases. Resolving this problem would be interesting future work.

\suppfigFailInter
\suppfigFailTrans

{\small
\bibliographystyle{ieee_fullname}
\bibliography{egbib}

\begin{thebibliography}{10}\itemsep=-1pt

\bibitem{abdal2019image2stylegan}
Rameen Abdal, Yipeng Qin, and Peter Wonka.
\newblock Image2stylegan: How to embed images into the stylegan latent space?
\newblock In {\em CVPR}, 2019.

\bibitem{abdal2020image2stylegan}
Rameen Abdal, Yipeng Qin, and Peter Wonka.
\newblock Image2stylegan++: How to edit the embedded images?
\newblock In {\em CVPR}, 2020.

\bibitem{alharbi2020structurednoise}
Yazeed Alharbi and Peter Wonka.
\newblock Disentangled image generation through structured noise injection.
\newblock In {\em CVPR}, 2020.

\bibitem{Bau_2019}
David Bau, Hendrik Strobelt, William Peebles, Jonas Wulff, Bolei Zhou, Jun-Yan
  Zhu, and Antonio Torralba.
\newblock Semantic photo manipulation with a generative image prior.
\newblock {\em ACM Transactions on Graphics}, 2019.

\bibitem{bau2018gan}
David Bau, Jun-Yan Zhu, Hendrik Strobelt, Bolei Zhou, Joshua~B Tenenbaum,
  William~T Freeman, and Antonio Torralba.
\newblock Gan dissection: Visualizing and understanding generative adversarial
  networks.
\newblock {\em ICLR}, 2019.

\bibitem{brock2018biggan}
Andrew Brock, Jeff Donahue, and Karen Simonyan.
\newblock Large scale gan training for high fidelity natural image synthesis.
\newblock In {\em ICLR}, 2019.

\bibitem{brock2016neuralphotoediting}
Andrew Brock, Theodore Lim, James~M Ritchie, and Nick Weston.
\newblock Neural photo editing with introspective adversarial networks.
\newblock {\em arXiv preprint arXiv:1609.07093}, 2016.

\bibitem{chang2018pairedcyclegan}
Huiwen Chang, Jingwan Lu, Fisher Yu, and Adam Finkelstein.
\newblock Pairedcyclegan: Asymmetric style transfer for applying and removing
  makeup.
\newblock In {\em CVPR}, 2018.

\bibitem{choi2018stargan}
Yunjey Choi, Minje Choi, Munyoung Kim, Jung-Woo Ha, Sunghun Kim, and Jaegul
  Choo.
\newblock Stargan: Unified generative adversarial networks for multi-domain
  image-to-image translation.
\newblock In {\em CVPR}, 2018.

\bibitem{choi2020starganv2}
Yunjey Choi, Youngjung Uh, Jaejun Yoo, and Jung-Woo Ha.
\newblock Stargan v2: Diverse image synthesis for multiple domains.
\newblock In {\em CVPR}, 2020.

\bibitem{collins2020editinginstyle}
Edo Collins, Raja Bala, Bob Price, and Sabine Susstrunk.
\newblock Editing in style: Uncovering the local semantics of gans.
\newblock In {\em CVPR}, 2020.

\bibitem{donahue2016bigan}
Jeff Donahue, Philipp Kr{\"a}henb{\"u}hl, and Trevor Darrell.
\newblock Adversarial feature learning.
\newblock {\em arXiv preprint arXiv:1605.09782}, 2016.

\bibitem{donahue2019bigbigan}
Jeff Donahue and Karen Simonyan.
\newblock Large scale adversarial representation learning.
\newblock In {\em NeurIPS}, 2019.

\bibitem{dumoulin2016ali}
Vincent Dumoulin, Ishmael Belghazi, Ben Poole, Olivier Mastropietro, Alex Lamb,
  Martin Arjovsky, and Aaron Courville.
\newblock Adversarially learned inference.
\newblock {\em arXiv preprint arXiv:1606.00704}, 2016.

\bibitem{goetschalckx2019gananalyze}
Lore Goetschalckx, Alex Andonian, Aude Oliva, and Phillip Isola.
\newblock Ganalyze: Toward visual definitions of cognitive image properties.
\newblock In {\em ICCV}, 2019.

\bibitem{goodfellow2014gan}
Ian Goodfellow, Jean Pouget-Abadie, Mehdi Mirza, Bing Xu, David Warde-Farley,
  Sherjil Ozair, Aaron Courville, and Yoshua Bengio.
\newblock Generative adversarial networks.
\newblock In {\em NeurIPS}, 2014.

\bibitem{harkonen2020ganspace}
Erik H{\"a}rk{\"o}nen, Aaron Hertzmann, Jaakko Lehtinen, and Sylvain Paris.
\newblock Ganspace: Discovering interpretable gan controls.
\newblock {\em arXiv preprint}, 2020.

\bibitem{heusel2017fid}
Martin Heusel, Hubert Ramsauer, Thomas Unterthiner, Bernhard Nessler, and Sepp
  Hochreiter.
\newblock Gans trained by a two time-scale update rule converge to a local nash
  equilibrium.
\newblock In {\em NeurIPS}, 2017.

\bibitem{hinton2006reducing}
Geoffrey~E Hinton and Ruslan~R Salakhutdinov.
\newblock Reducing the dimensionality of data with neural networks.
\newblock {\em science}, 2006.

\bibitem{huh2020pix2latent}
Minyoung Huh, Richard Zhang, Jun-Yan Zhu, Sylvain Paris, and Aaron Hertzmann.
\newblock Transforming and projecting images into class-conditional generative
  networks.
\newblock {\em arXiv preprint arXiv:2005.01703}, 2020.

\bibitem{isola2017pix2pix}
Phillip Isola, Jun-Yan Zhu, Tinghui Zhou, and Alexei~A Efros.
\newblock Image-to-image translation with conditional adversarial nets.
\newblock In {\em CVPR}, 2017.

\bibitem{jahanian2019steerability}
Ali Jahanian, Lucy Chai, and Phillip Isola.
\newblock On the ``steerability" of generative adversarial networks.
\newblock In {\em ICLR}, 2020.

\bibitem{johnson2016perceptual}
Justin Johnson, Alexandre Alahi, and Li Fei-Fei.
\newblock Perceptual losses for real-time style transfer and super-resolution.
\newblock In {\em ECCV}, 2016.

\bibitem{karras2017progressivegan}
Tero Karras, Timo Aila, Samuli Laine, and Jaakko Lehtinen.
\newblock Progressive growing of gans for improved quality, stability, and
  variation.
\newblock In {\em ICLR}, 2018.

\bibitem{karras2019stylegan}
Tero Karras, Samuli Laine, and Timo Aila.
\newblock A style-based generator architecture for generative adversarial
  networks.
\newblock In {\em CVPR}, 2019.

\bibitem{karras2020stylegan2}
Tero Karras, Samuli Laine, Miika Aittala, Janne Hellsten, Jaakko Lehtinen, and
  Timo Aila.
\newblock Analyzing and improving the image quality of stylegan.
\newblock In {\em CVPR}, 2020.

\bibitem{kim2019tag2pix}
Hyunsu Kim, Ho~Young Jhoo, Eunhyeok Park, and Sungjoo Yoo.
\newblock Tag2pix: Line art colorization using text tag with secat and changing
  loss.
\newblock In {\em ICCV}, 2019.

\bibitem{kim2018nsml}
Hanjoo Kim, Minkyu Kim, Dongjoo Seo, Jinwoong Kim, Heungseok Park, Soeun Park,
  Hyunwoo Jo, KyungHyun Kim, Youngil Yang, Youngkwan Kim, et~al.
\newblock Nsml: Meet the mlaas platform with a real-world case study.
\newblock {\em arXiv preprint arXiv:1810.09957}, 2018.

\bibitem{Kim2020U-GAT-IT:}
Junho Kim, Minjae Kim, Hyeonwoo Kang, and Kwang~Hee Lee.
\newblock U-gat-it: Unsupervised generative attentional networks with adaptive
  layer-instance normalization for image-to-image translation.
\newblock In {\em ICLR}, 2020.

\bibitem{kingma2014adam}
Diederik~P Kingma and Jimmy Ba.
\newblock Adam: A method for stochastic optimization.
\newblock In {\em ICLR}, 2015.

\bibitem{kingma2018glow}
Diederik~P. Kingma and Prafulla Dhariwal.
\newblock Glow: Generative flow with invertible 1x1 convolutions.
\newblock In {\em NeurIPS}, 2018.

\bibitem{kingma2013vae}
Diederik~P Kingma and Max Welling.
\newblock Auto-encoding variational bayes.
\newblock {\em arXiv preprint}, 2013.

\bibitem{lample2017fadernetwork}
Guillaume Lample, Neil Zeghidour, Nicolas Usunier, Antoine Bordes, Ludovic
  Denoyer, and Marc'Aurelio Ranzato.
\newblock Fader networks: Manipulating images by sliding attributes.
\newblock In {\em NeurIPS}, 2017.

\bibitem{larsen2016vaegan}
Anders Boesen~Lindbo Larsen, S{\o}ren~Kaae S{\o}nderby, Hugo Larochelle, and
  Ole Winther.
\newblock Autoencoding beyond pixels using a learned similarity metric.
\newblock In {\em ICML}, 2016.

\bibitem{CelebAMask-HQ}
Cheng-Han Lee, Ziwei Liu, Lingyun Wu, and Ping Luo.
\newblock Maskgan: Towards diverse and interactive facial image manipulation.
\newblock In {\em CVPR}, 2020.

\bibitem{luo2017autoencodergan}
Junyu Luo, Yong Xu, Chenwei Tang, and Jiancheng Lv.
\newblock Learning inverse mapping by autoencoder based generative adversarial
  nets.
\newblock In {\em ICNIP}, 2017.

\bibitem{ma2018invertibility}
Fangchang Ma, Ulas Ayaz, and Sertac Karaman.
\newblock Invertibility of convolutional generative networks from partial
  measurements.
\newblock In {\em NeurIPS}, 2018.

\bibitem{maas2013lrelu}
Andrew~L Maas, Awni~Y Hannun, and Andrew~Y Ng.
\newblock Rectifier nonlinearities improve neural network acoustic models.
\newblock In {\em ICML}, 2013.

\bibitem{Marchesi2017MegapixelSI}
M. Marchesi.
\newblock Megapixel size image creation using generative adversarial networks.
\newblock {\em ArXiv}, abs/1706.00082, 2017.

\bibitem{mescheder2018r1reg}
Lars Mescheder, Sebastian Nowozin, and Andreas Geiger.
\newblock Which training methods for gans do actually converge?
\newblock In {\em ICML}, 2018.

\bibitem{noguchi2019smalldatagan}
Atsuhiro Noguchi and Tatsuya Harada.
\newblock Image generation from small datasets via batch statistics adaptation.
\newblock In {\em ICCV}, 2019.

\bibitem{park2019spade}
Taesung Park, Ming-Yu Liu, Ting-Chun Wang, and Jun-Yan Zhu.
\newblock Semantic image synthesis with spatially-adaptive normalization.
\newblock In {\em CVPR}, 2019.

\bibitem{park2020swapping}
Taesung Park, Jun-Yan Zhu, Oliver Wang, Jingwan Lu, Eli Shechtman, Alexei~A.
  Efros, and Richard Zhang.
\newblock Swapping autoencoder for deep image manipulation.
\newblock In {\em NeurIPS}, 2020.

\bibitem{perarnau2016icgan}
Guim Perarnau, Joost Van De~Weijer, Bogdan Raducanu, and Jose~M {\'A}lvarez.
\newblock Invertible conditional gans for image editing.
\newblock {\em arXiv preprint}, 2016.

\bibitem{pidhorskyi2020alae}
Stanislav Pidhorskyi, Donald Adjeroh, and Gianfranco Doretto.
\newblock Adversarial latent autoencoders.
\newblock In {\em CVPR}, 2020.

\bibitem{salimans2016improved}
Tim Salimans, Ian Goodfellow, Wojciech Zaremba, Vicki Cheung, Alec Radford, and
  Xi Chen.
\newblock Improved techniques for training gans.
\newblock In {\em NeurIPS}, 2016.

\bibitem{shen2020interfacegan}
Yujun Shen, Jinjin Gu, Xiaoou Tang, and Bolei Zhou.
\newblock Interpreting the latent space of gans for semantic face editing.
\newblock In {\em CVPR}, 2020.

\bibitem{shen2020closedformgan}
Yujun Shen and Bolei Zhou.
\newblock Closed-form factorization of latent semantics in gans.
\newblock {\em arXiv preprint}, 2020.

\bibitem{shocher2020semantic}
Assaf Shocher, Yossi Gandelsman, Inbar Mosseri, Michal Yarom, Michal Irani,
  William~T Freeman, and Tali Dekel.
\newblock Semantic pyramid for image generation.
\newblock In {\em CVPR}, 2020.

\bibitem{simonyan2015deep}
Karen Simonyan and Andrew Zisserman.
\newblock Very deep convolutional networks for large-scale image recognition.
\newblock In {\em ICLR}, 2015.

\bibitem{srivastava2017veegan}
Akash Srivastava, Lazar Valkov, Chris Russell, Michael~U. Gutmann, and Charles
  Sutton.
\newblock Veegan: Reducing mode collapse in gans using implicit variational
  learning.
\newblock In {\em NeurIPS}, 2017.

\bibitem{sung2017nsml}
Nako Sung, Minkyu Kim, Hyunwoo Jo, Youngil Yang, Jingwoong Kim, Leonard Lausen,
  Youngkwan Kim, Gayoung Lee, Donghyun Kwak, Jung-Woo Ha, et~al.
\newblock Nsml: A machine learning platform that enables you to focus on your
  models.
\newblock {\em arXiv preprint arXiv:1712.05902}, 2017.

\bibitem{suzuki2018neuralcollage}
Ryohei Suzuki, Masanori Koyama, Takeru Miyato, Taizan Yonetsuji, and Huachun
  Zhu.
\newblock Spatially controllable image synthesis with internal representation
  collaging.
\newblock {\em arXiv preprint arXiv:1811.10153}, 2018.

\bibitem{szegedy2016inceptionv3}
Christian Szegedy, Vincent Vanhoucke, Sergey Ioffe, Jon Shlens, and Zbigniew
  Wojna.
\newblock Rethinking the inception architecture for computer vision.
\newblock In {\em CVPR}, 2016.

\bibitem{ulyanov2017generatorencoder}
Dmitry Ulyanov, Andrea Vedaldi, and Victor Lempitsky.
\newblock It takes (only) two: Adversarial generator-encoder networks.
\newblock {\em arXiv preprint arXiv:1704.02304}, 2017.

\bibitem{ulyanov2017takes}
Dmitry Ulyanov, Andrea Vedaldi, and Victor Lempitsky.
\newblock It takes (only) two: Adversarial generator-encoder networks.
\newblock In {\em AAAI}, 2017.

\bibitem{voynov2020unsupervised}
Andrey Voynov and Artem Babenko.
\newblock Unsupervised discovery of interpretable directions in the gan latent
  space.
\newblock In {\em ICML}, 2020.

\bibitem{wang2020cnndetector}
Sheng-Yu Wang, Oliver Wang, Richard Zhang, Andrew Owens, and Alexei~A Efros.
\newblock Cnn-generated images are surprisingly easy to spot... for now.
\newblock In {\em CVPR}, 2020.

\bibitem{yu2015lsun}
Fisher Yu, Ari Seff, Yinda Zhang, Shuran Song, Thomas Funkhouser, and Jianxiong
  Xiao.
\newblock Lsun: Construction of a large-scale image dataset using deep learning
  with humans in the loop.
\newblock {\em arXiv preprint arXiv:1506.03365}, 2015.

\bibitem{zhan2019spatial}
Fangneng Zhan, Hongyuan Zhu, and Shijian Lu.
\newblock Spatial fusion gan for image synthesis.
\newblock In {\em CVPR}, 2019.

\bibitem{zhang2018lpips}
Richard Zhang, Phillip Isola, Alexei~A Efros, Eli Shechtman, and Oliver Wang.
\newblock The unreasonable effectiveness of deep features as a perceptual
  metric.
\newblock In {\em CVPR}, 2018.

\bibitem{zhu2020indomaingan}
Jiapeng Zhu, Yujun Shen, Deli Zhao, and Bolei Zhou.
\newblock In-domain gan inversion for real image editing.
\newblock In {\em ECCV}, 2020.

\bibitem{zhu2016igan}
Jun-Yan Zhu, Philipp Kr{\"a}henb{\"u}hl, Eli Shechtman, and Alexei~A Efros.
\newblock Generative visual manipulation on the natural image manifold.
\newblock In {\em ECCV}, 2016.

\bibitem{zhu2017cyclegan}
Jun-Yan Zhu, Taesung Park, Phillip Isola, and Alexei~A Efros.
\newblock Unpaired image-to-image translation using cycle-consistent
  adversarial networks.
\newblock In {\em ICCV}, 2017.

\bibitem{zhu2020sean}
Peihao Zhu, Rameen Abdal, Yipeng Qin, and Peter Wonka.
\newblock Sean: Image synthesis with semantic region-adaptive normalization.
\newblock In {\em CVPR}, 2020.

\end{thebibliography}
}

\end{document}